\documentclass[a4paper,fleqn,oneside]{cas-dc}

\UseRawInputEncoding

\usepackage{amsmath,amssymb,amsfonts}

\usepackage{graphicx}
\usepackage[caption=false,font=normalsize,labelfont=sf,textfont=sf]{subfig}
\usepackage{svg}
\usepackage{ragged2e}
\usepackage{caption}

\usepackage{booktabs}
\usepackage{longtable}
\usepackage{array}
\usepackage{tabularx}
\usepackage{multirow}

\usepackage{algorithmic}

\usepackage{natbib}
\usepackage{hyperref}
\usepackage{textcomp}

\usepackage{stfloats}
\usepackage{cuted}
\usepackage{balance}

\usepackage{pifont}

\usepackage[dvipsnames]{xcolor}

\newcommand{\SmallTableFont}{\fontsize{6.8pt}{8pt}\selectfont}
\newcommand{\LargeTableFont}{\fontsize{7.1pt}{8.8pt}\selectfont}
\begin{document}


\title{\Large Vision Language Action Models in Robotic Manipulation: A Systematic Review}
\author[1]{Muhayy {Ud Din}}
\author[1]{Waseem Akram}
\author[1]{Lyes {Saad Saoud}}
\author[2]{Jan Rosell}
\author[1]{Irfan Hussain*}

\affiliation[1]{Khalifa University Center for Autonomous Robotic Systems (KUCARS), Khalifa University, United Arab Emirates.}
\affiliation[2]{Institute of Industrial and Control Engineering (IOC), Universitat Politecnica de Catalunya, Spain.}

\cortext[cor1]{corresponding author: irfan.hussain@ku.ac.ae}
\shorttitle {Vision Language Action Models a Review}
\shortauthors{Muhayy Ud Din et~al.}

\begin{keywords}
Vision language models \sep Robotic and embodied control \sep Foundation models \sep Language-conditioned manipulation and control
\end{keywords}

\maketitle

\begin{abstract}
Vision Language Action (VLA) models represent a transformative shift in robotics, with the aim of unifying visual perception, natural language understanding, and embodied control within a single learning framework. This review presents a comprehensive and forward-looking synthesis of the VLA paradigm, with a particular emphasis on robotic manipulation and instruction-driven autonomy. We comprehensively analyze 102 VLA models, 26 foundational datasets, and 12 simulation platforms that collectively shape the development and evaluation of VLAs models. These models are categorized into key architectural paradigms,
each reflecting distinct strategies for integrating vision, language, and control in robotic systems.
Foundational datasets are evaluated using a novel criterion based on task complexity, variety of modalities, and dataset scale, allowing a comparative analysis of their suitability for generalist policy learning. We introduce a two-dimensional characterization framework that organizes these datasets based on semantic richness and multimodal alignment, showing underexplored regions in the current data landscape. Simulation environments are evaluated for their effectiveness in generating large-scale data, as well as their ability to facilitate transfer from simulation to real-world settings and the variety of supported tasks.
Using both academic and industrial contributions, we recognize ongoing challenges and outline strategic directions such as scalable pretraining protocols, modular architectural design, and robust multimodal alignment strategies. This review serves as both a technical reference and a conceptual roadmap for advancing embodiment and robotic control, providing insights that span from dataset generation to real world deployment of generalist robotic agents. For reference, a public repository summarizing VLA models, datasets, and simulators can be found at: \hbox{\textcolor{blue}{\url{https://github.com/Muhayyuddin/VLAs}}}
\end{abstract}
\tableofcontents
\section{Introduction}
\begin{figure*}[t]
    \centering
    \includegraphics[width=0.9\linewidth]{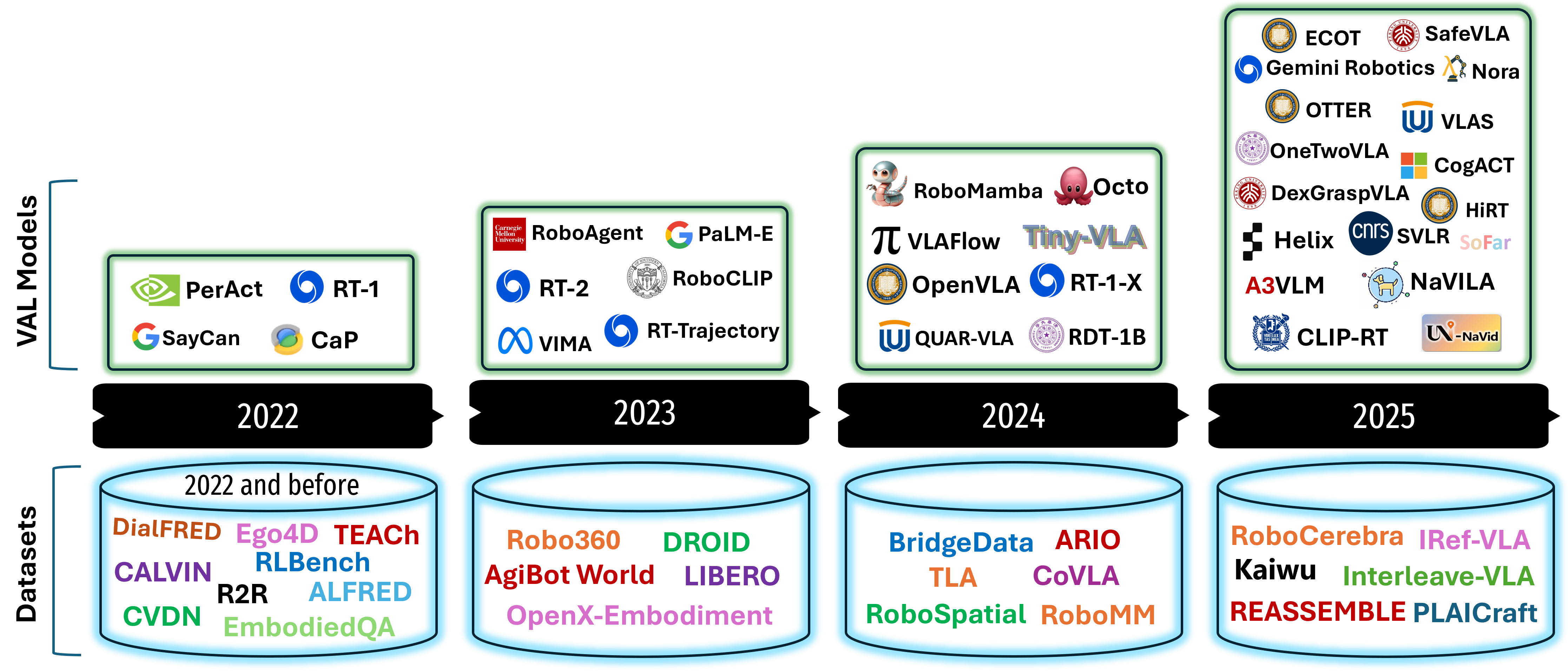} 
    \caption{VLA models, datasets, and contributing institutions from 2022 to 2025. The top row presents major VLA models introduced each year, alongside their associated institutions (logos within red boxes). The bottom row displays key datasets used to train and evaluate these models, grouped by release year.  The figure highlights the increasing scale and diversity of datasets and institutional involvement, with contributions from academic (e.g., CMU, CNRS, UC, Peking Uni) and industrial labs (e.g., Google, NVIDIA, Microsoft). This timeline highlights the rapid advancements in VLA research.}
    \label{fig:trend}
\end{figure*}

The integration of vision, language, and action into a unified framework has emerged as a paradigm-shifting approach in robotics and embodied artificial intelligence. Traditionally, robotic systems that depend on task-specific programming struggle in dynamic and unstructured environments. In contrast, VLA models aim to utilize the generalization capabilities of large-scale foundation models to enable robotic systems that can understand instructions in natural language, perceive their surroundings, and perform complex tasks autonomously \citep{cite:18,ahn2022saycan,jiang2022vima,team2024octo}.
The fundamental concept of this paradigm is transformer architecture, which has transformed natural language processing and vision with self-attention mechanisms and large-scale pretraining. Models like GPT \citep{brown2020language}, BERT \citep{devlin2018bert}, ViT \citep{dosovitskiy2020image}, and CLIP \citep{radford2021learning} demonstrate that massive datasets and parameter scales can produce remarkable generalizability and robustness. These insights have led to new architectures that fuse vision and language into robotics control policies, which step forward towards generalist agents such as \hbox{RT-1} \citep{cite:18}, SayCan \citep{ahn2022saycan}, VIMA \citep{jiang2022vima}, and Octo \citep{team2024octo}. The trend of developing such models is continuously growing as depicted in Fig~\ref{fig:trend}. 

The development of effective VLA models is fundamentally dependent on the availability of large-scale, diverse, and multi-model datasets, together with realistic simulation platforms. These elements are essential for training models that can robustly understand language instructions, perceive visual environments, and generate meaningful action sequences.
For instance, the \textit{Open X-Embodiment}~\citep{openx} dataset unifies data from 22 robot embodiments and more than 500 tasks using a shared action space. It enables the pre-training of foundation models like RT-1-X, significantly enhancing cross-robot generalization. Similarly, the \textit{DROID} dataset~\citep{droid} uses internet-scale data, combining human-annotated language with robotic video demonstrations for scenes with complex manipulations. These datasets significantly advance the data ecosystem for training and benchmarking VLAs.
Comprehensive datasets enable the learning of diverse tasks in both household and industrial contexts. These datasets offer rich annotations, including demonstration trajectories, object state transitions, and diverse natural language prompts. However, real-world data collection is labor-intensive, expensive, and limited in diversity, which highlights the importance of simulation.

Simulation environments allow data generation to be scaled across a wide range of settings, object types, lighting conditions, and agent embodiments. Platforms such as \textit{Habitat} \citep{savva2019habitat}, \textit{Isaac Gym} \citep{makoviychuk2021isaac}, and \textit{RoboSuite} \citep{zhu2020robosuite} offer programmable and photorealistic environments with physics-based interactions, facilitating both imitation and reinforcement learning paradigms.
More recently, tools such as \textit{ iGibson} \citep{xia2020interactive} and \textit{AI2-THOR} \citep{kolve2017ai2} have added support for human-centric indoor environments with naturalistic object arrangements, enhancing semantic realism. Simulation also enables automatic generation of multimodal annotations, such as action trajectories, object states, and natural language instructions that are crucial to align visual, linguistic, and motor modalities.
Recent efforts also emphasize the importance of synthetic language generation aligned with task semantics (e.g., VLN-CE \citep{krantz2020beyond}, ALFRED \citep{shridhar2020alfred}) to ensure linguistic diversity and instruction complexity. The integration of simulation and large-scale synthetic datasets is therefore crucial
for building VLA systems that are robust, scalable, and applicable to real-world deployment. 

Since the field matures at a rapid pace, numerous architectures, datasets, and frameworks are being proposed in a fast order. Despite the growing body of research, there remains a gap in the literature for a comprehensive and systematic synthesis that organizes and categorizes the architectural foundations, benchmark datasets, simulation platforms, and evaluation protocols that collectively shape the current VLA landscape. It is critically needed to contextualize the state-of-the-art and identify emerging patterns, limitations, and opportunities.  
This study will serve as both
a technical reference and a conceptual roadmap to accelerate research in embodied foundation models and generalist robotic intelligence.
\\
\textit{\textbf{Contributions:}}\\
This systematic review makes the following key contributions to the field of VLA models in robotics:

\begin{itemize}
\item \textit{Structured taxonomy of VLA architectures:} We present a detailed classification of VLA model architectures, organizing them into key categories according to their distinct methods for integrating perception, natural language understanding, and embodied control.
\item \textit{A novel quantitative benchmarking for VLA datasets:} We introduce a novel quantitative framework designed to systematically benchmark VLA datasets, using metrics that are empirically adjusted for task complexity ($\mathcal{C}_{\text{task}}$) and modality richness ($\mathcal{C}_{\text{mod}}$). This benchmarking approach generates a two-dimensional graph (Fig.~\ref{fig:vla_landscape}) that efficiently represents current datasets, allowing us to pinpoint significant gaps, particularly the lack of datasets that combine very high levels of task complexity with comprehensive multi-modelities.
\item \textit{In-depth review of simulation platforms:} We review the most critical simulation platforms used in VLA research, focusing on their distinct sensory modalities, applications and their essential role in facilitating large-scale and reproducible VLA experiments and data generation.
\item \textit{Identification of challenges and future roadmap:} We identify persistent challenges facing VLA model development and articulate a clear roadmap for future research directions, emphasizing key areas such as architectural modularity, scalable data generation strategies, advancing dynamic simulation through differentiable contact modeling, and unified language-grounding APIs to enable robust and scalable real-world deployment.
\end{itemize}
\begin{figure}[t]
    \centering
    \includegraphics[width=1\linewidth]{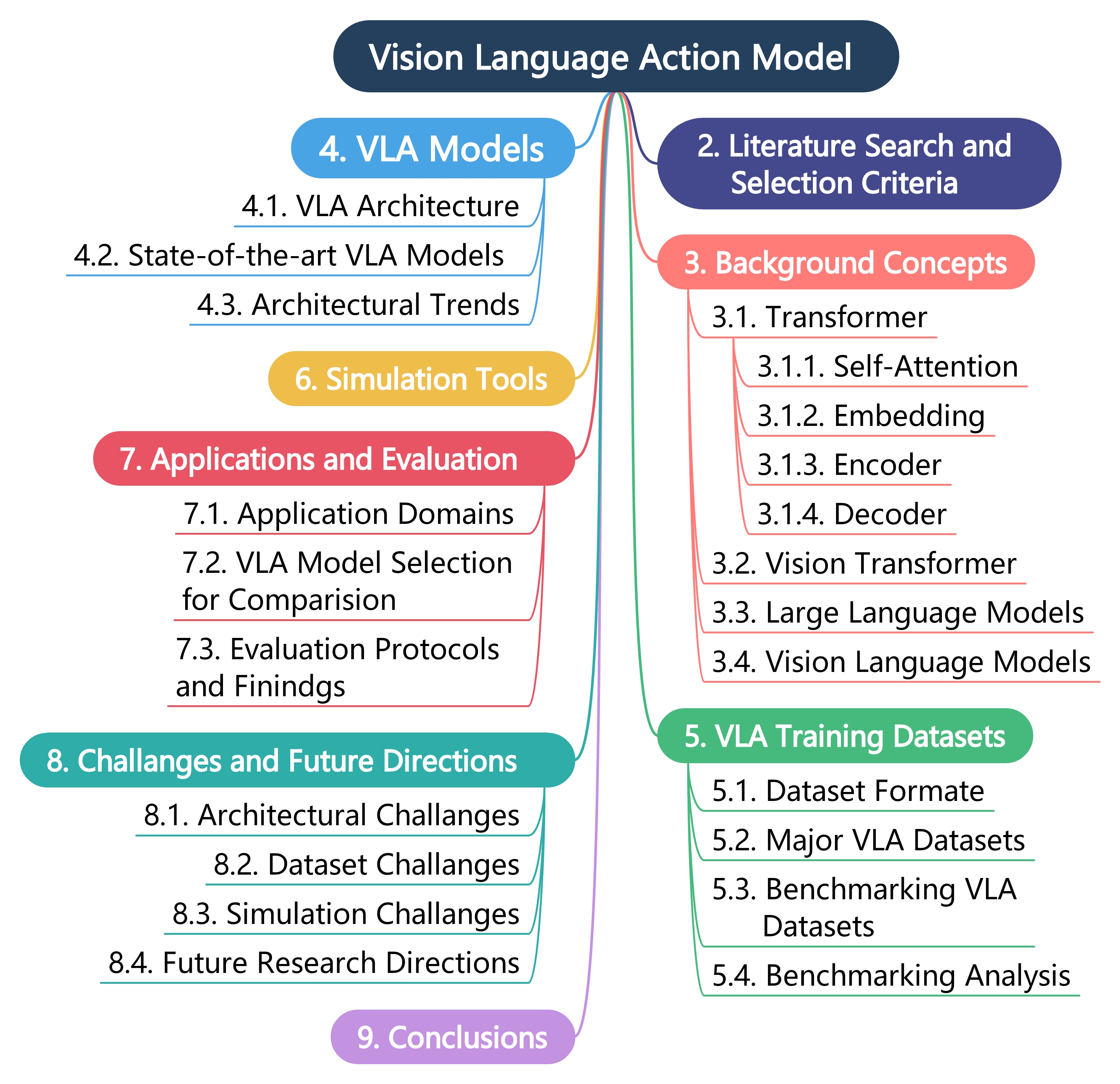} 
    \caption{Overview of the skeleton of the paper, highlighting the main sections and their interrelated subtopics.}
    \label{fig:structure}
\end{figure}
The rest of the paper is organized as illustrated in Fig.~\ref{fig:structure}. Sec.~\ref{sec:search} outlines the literature search and selection criteria used to collect the most relevant and representative VLA systems. Sec.~\ref{sec:background} introduces core background concepts including transformers, vision transformers, large-language models, and vision-language models, which collectively form the foundation of modern VLA architectures. Sec.~\ref{sec:vla} presents an in-depth review of VLA model architectures, highlights state-of-the-art systems, and identifies common architectural trends. Sec.~\ref{sec:dataset} lists and analyzes the main training datasets, discussing dataset formats, benchmarking protocols, and evaluation metrics. Sec.~\ref{sec:simtools} reviews the simulation platforms and tools used for training and testing VLA agents. Sec.~\ref{sec:vlabenchmark} outlines practical application domains, describes the criteria used for selecting representative VLA models for comparison, and details evaluation protocols and key findings. Sec.~\ref{sec:chalanges} discusses key challenges and future directions, categorizing them into architectural, dataset, and simulation challenges, and outlines open problems and opportunities for future research. Finally, Sec.~\ref{sec:conclusion} summarizes the conclusions and offers a roadmap for advancing
VLA-driven robotic autonomy.

\section{Literature Search and Selection Criteria}\label{sec:search}
We conducted an extensive search through IEEE Xplore, Elsevier, Springer Nature, MDPI, Wiley, and arXiv to identify work on VLA models, VLA datasets, and simulation tools for robotic simulation and data generation. To capture each aspect, we developed sets of keywords; for \textit{VLA models}: "vision language action" OR VLA OR (vision AND language AND action) OR Vision Language Models for Robotic manipulation" "Multimodal Robotic Control", "Vision-Language Grounding",  "Visuomotor Transformer", "End-to-End Robot Learning". The keywords for \textit{Training datasets} are "VLA dataset" OR "manipulation dataset" OR "embodied AI dataset for manipulation". Finally for \textit{Simulation tools}: "simulator for robotic manipulation" OR "embodied AI data-generation simulator" OR "robotic manipulation simulator" OR "realistic dynamic simulation for robotic manipulation" 
We applied the following inclusion criteria to identify original research:
1) Proposes or evaluates a VLA model, a VLA dataset, or a simulator for robotic manipulation or manipulation data generation. 2) Presents a novel model, dataset, or a new simulator.

\begin{figure}[t]
    \centering
    \includegraphics[width=1\linewidth]{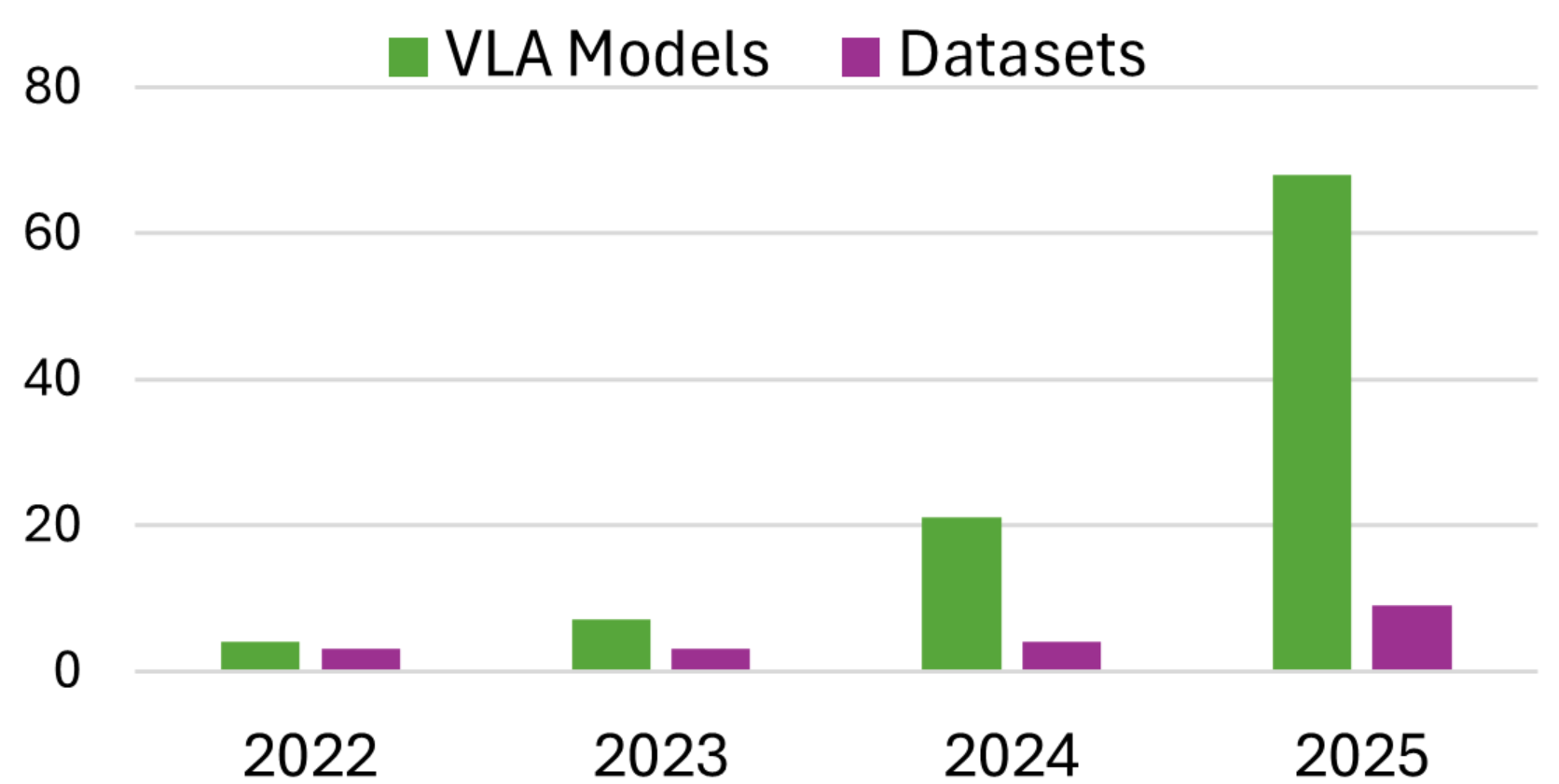} 
    \caption{Annual VLA models and foundational VLA datasets count from 2022 to 2025. Green bars indicate the number of new VLA model introduced each year, while purple bars represent the number of novel dataset releases. The data illustrate a rapid acceleration in model development, particularly in 2025, alongside steady growth in dataset creation to support training and evaluation of these models.
}\label{fig:hist} 
\end{figure}

To ensure comprehensive coverage, we complement traditional database searches with conversational queries in a large-language model (e.g. GPT), using targeted prompts per thematic area. For \textit{VLA models}, we asked GPT to "List vision-language-action models published between 2022 and 2025," "List end-to-end transformer-based VLA architectures", "List VLA methods that use diffusion-based action decoders", "List modular fusion framework VLA models". For datasets, prompts included "List of VLA datasets released between 2022 and 2025", "List manipulation datasets used for VLM training" and "List embodied AI datasets for robotics" and "List of well-known manipulation datasets". For simulation tools, we queried "List simulators for robotic manipulation data generation," "List simulation environments for embodied AI dataset creation,"  "List robotic manipulation dataset generation simulators" and "List robotic dynamic simulators". We then merged these lists with our database results, removed duplicates, and performed manual validation to arrive at the final set of models, datasets, and simulators included in this review.

We also included arXiv e-prints (\url{https://arxiv.org/}) in our search because the field of VLAs has recently begun to mature, and most breakthroughs and novel architectures have appeared as preprints in recent times. Fig.~\ref{fig:hist} shows the VLA (in green) and dataset (in purple) counts per year used in this work. We thoroughly examined the preprints and included only those that make substantial contributions to the field. The integration of preprints ensures that we capture the very latest models, datasets, and simulation tools as soon as they emerge, giving a more accurate and up-to-date picture of this rapidly evolving field.

\section{Background Concepts}\label{sec:background}

This section outlines the core architectures behind VLA models. We start with the transformer, a model that has transformed both language and vision tasks. We then cover Vision Transformers (ViTs), which apply self-attention to image patches for visual feature extraction.
The subsequent category comprises Large Language Models (LLMs), which are transformer-based models trained on large text datasets that perform reasoning, instruction following, and zero-shot tasks. Finally, we will provide an overview of vision language models (VLMs), which fuse visual and textual data through cross-modal attention to ground instructions in robotic actions.
These components form the backbone of modern VLA architectures, detailed in the following subsections.

\subsection{Transformer}

Transformers are a class of deep learning architectures introduced by Vaswani et al. \citep{vaswani2017attention}, these models have revolutionized the fields of natural language processing and computer vision by enabling greater parallelization and scalability in sequence modeling, i.e. in the processing and learning of patterns from sequences of data. The core of Transformers lie in the \textit{self-attention mechanism}, detailed below, which let each token in a sequence weigh and combine information from all other tokens to build a context-aware representation.

\begin{figure}[t]
\centering
\includegraphics[width=0.9\linewidth]{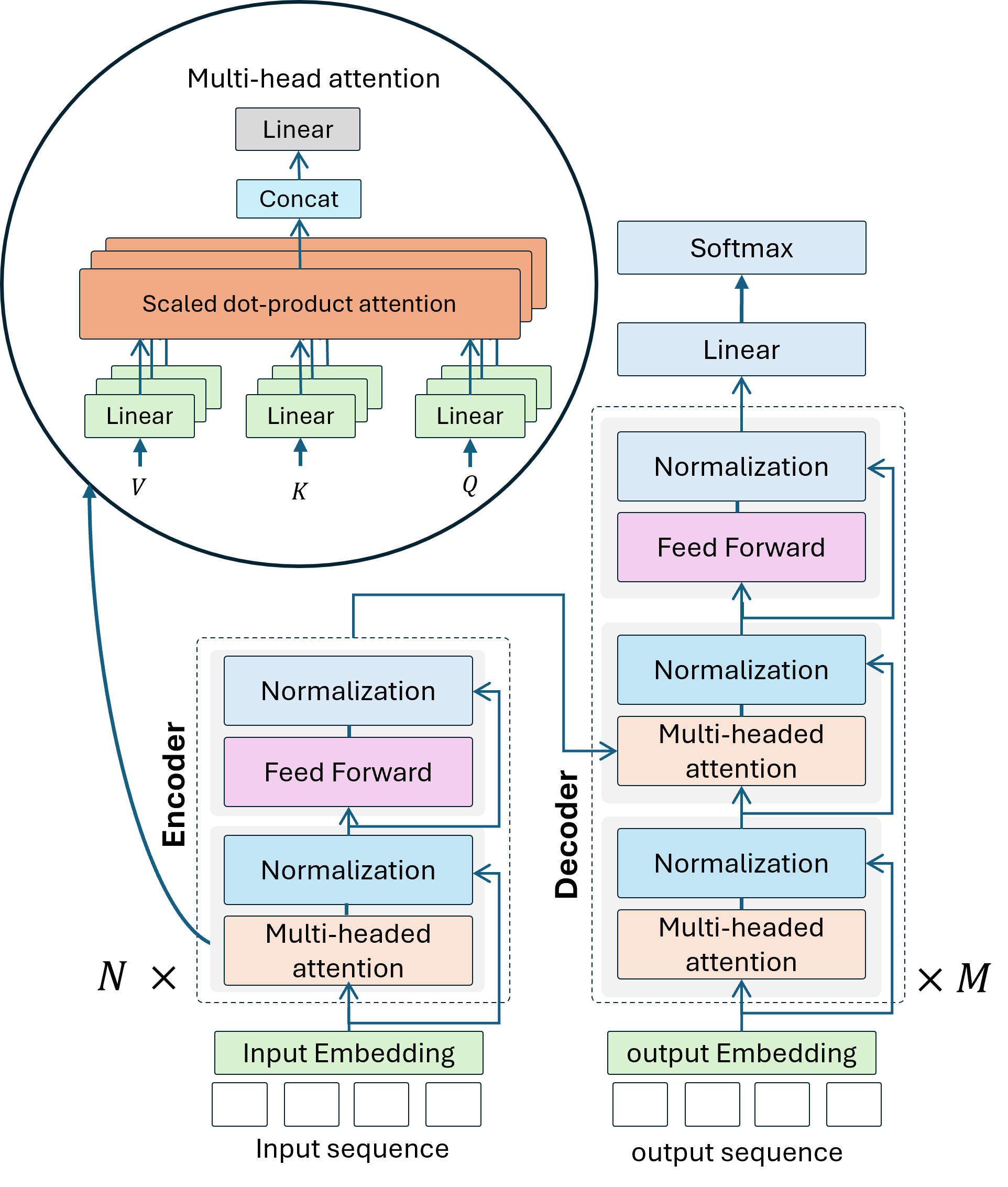}
\caption{An overview of the Transformer architecture highlighting the encoder-decoder structure and the internal mechanism of multi-head attention. The encoder processes input embeddings through layers of multi-head attention, normalization, and feedforward networks. The decoder mirrors this with additional masked attention layers and incorporates encoder outputs for contextual decoding. The magnified view illustrates the scaled dot-product attention and how multiple attention heads are concatenated and linearly transformed to form the final multi-head attention output. The image is adapted from~\citep{vaswani2017attention}}
\label{fig:transformer_architecture}
\end{figure}

The Transformer architecture (depicted in Fig~\ref{fig:transformer_architecture}) comprises three differentiated parts: the embedding layer, the encoder stack, and the decoder stack, which includes the final output projection and softmax layer. They are detailed in the following subsections, after presenting the \textit{self-attention mechanism}.

\subsubsection{Self-Attention}
The self-attention mechanism allows each token in a sequence to attend to all other tokens when computing its representation. To achieve this, the model first transforms each input token into three distinct vector representations: queries ($Q$), keys ($K$), and values ($V$). These vectors are obtained through learned linear projections of the input embeddings and serve different roles in the attention computation: the \textit{query} represents what the current token is looking for, the \textit{key} indicates what information each token provides, and the \textit{value} contains the actual content to be shared. By comparing queries to keys, the model determines attention weights, which are then used to aggregate the values into a new context-aware representation for each token. 

This process is formalized by the \textit{scaled dot-product attention} mechanism, which computes the dot product between queries and keys, scales the result by $\sqrt{d_k}$ (where $d_k$ is the dimension of the keys), applies a softmax function to obtain attention weights, and finally combines these weights with the values to produce the output:
\begin{equation}
\text{Attention}(Q, K, V) = \text{softmax}\left( \frac{QK^\top}{\sqrt{d_k}} \right)V
\end{equation}
Building upon this mechanism, \textit{multi-head attention} extends the model's capacity by enabling it to attend to information from multiple representation subspaces simultaneously. Instead of computing attention once with a single set of projection matrices, the model uses $h$ parallel attention layers, or \textit{heads}, each with its own learned weight matrices $W^Q_i$, $W^K_i$, and $W^V_i$. These matrices independently project the input into $h$ different subspaces, allowing each head to focus on different aspects of the input relationships:
{\small
\begin{eqnarray}
\text{MultiHead}(Q, K, V) \!\!\!\!\!\!&=&\!\!\!\!\!\! \text{Concat}(\text{head}_1, \dots, \text{head}_h)W^O\\
\text{where head}_i \!\!\!\!\!\!&=&\!\!\!\!\!\! \text{Attention}(QW^Q_i, KW^K_i, VW^V_i)
\end{eqnarray}
}
The outputs from all heads are concatenated and passed through a final linear projection using $W^O$ to produce the result of the multi-head attention module.

\subsubsection{Embeddings}
Embedding refers to continuous vector representations that map discrete inputs such as words, image patches, or action tokens into a dense latent space. In Transformer architectures, embeddings serve as the foundational interface between symbolic input (for example, textual instructions) and differentiable computation. Input embeddings enable the model to process structured data in a unified format, while output embeddings decode latent representations into action or token spaces. Since the Transformer architecture lacks recurrence, a set of sinusoidal positional encodings is added to the input embeddings to provide information about the order of elements in a sequence. These learned representations capture semantic, spatial, or temporal relationships that are essential for generalization and reasoning in multimodal tasks~\citep{vaswani2017attention, dosovitskiy2021an, press2017using}.
\subsubsection{Encoder}
The encoder stack consists of $N$ identical layers, where each layer has a multi-head self-attention mechanism followed by a position-wise feedforward network. Both sub-layers are equipped with residual connections (which add the input of a sub-layer to its output to help preserve information and ease optimization) and layer normalization (which normalizes activations to improve training stability). In the first layer, the input embeddings provide the queries, keys, and values. In deeper layers, these are derived from the output of the preceding layer.
\subsubsection{Decoder}
The decoder stack also consists of $M$ identical layers, but with a slightly different architecture. Each layer contains three sub-layers: a masked multi-head self-attention layer, a multi-head encoder-decoder attention layer, and a feedforward network. The masked self-attention prevents a position from attending to subsequent tokens, ensuring that the model generates output in an autoregressive manner. The encoder-decoder attention enables the decoder to query the encoder's outputs, integrating source information into the generation process. Like the encoder, each sub-layer in the decoder is followed by residual connections and layer normalization.

The final output of the decoder is passed through a linear transformation and a softmax function to produce a probability distribution over all possible output tokens, enabling the model to predict the next token in the target sequence one step at a time.

\subsection{Vision Transformers}

\begin{figure}[t]
\centering
\includegraphics[width=\linewidth]{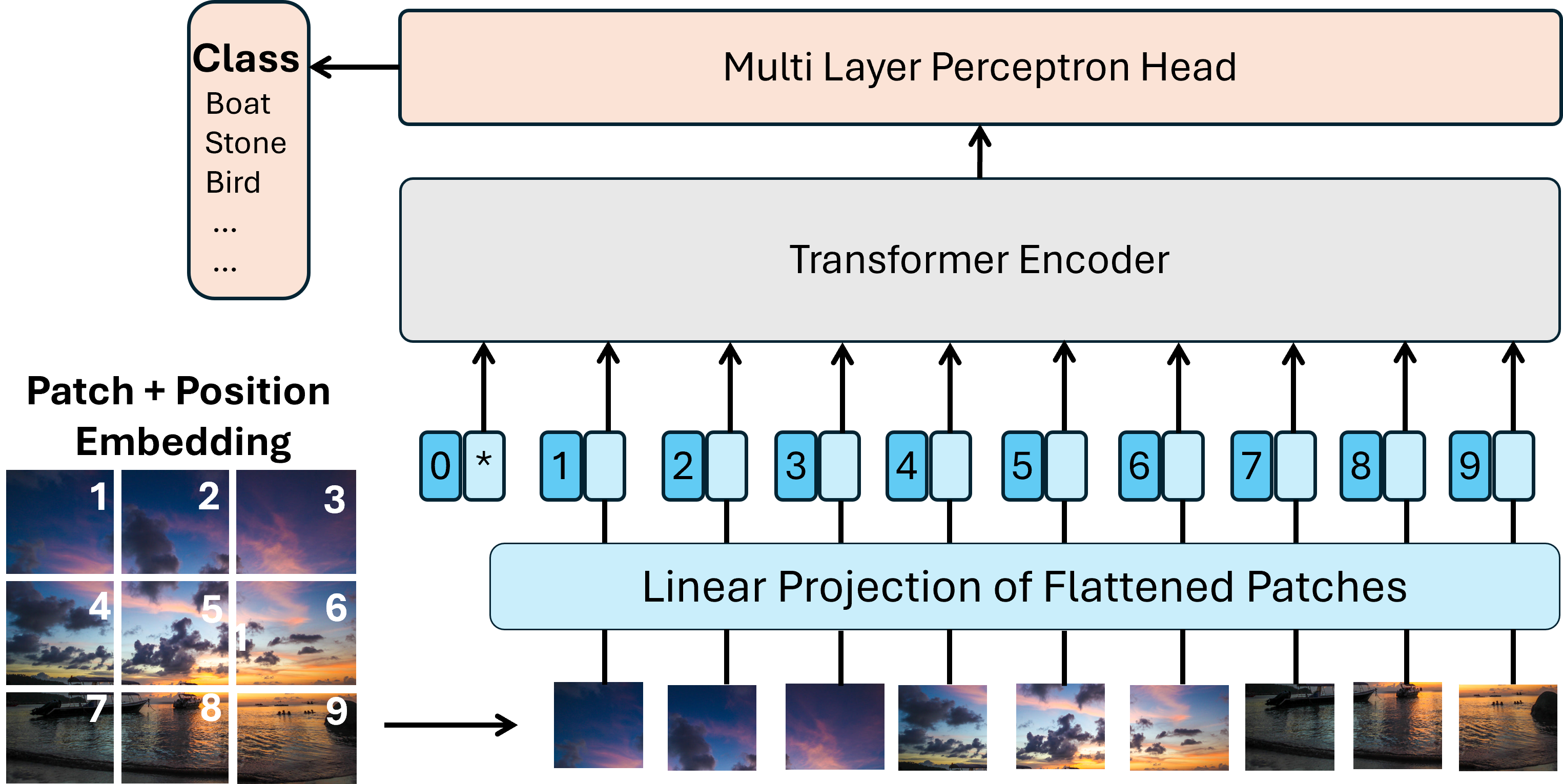}
\caption{Architecture of the ViT. The input image is divided into fixed-size non-overlapping patches which are flattened and linearly projected into embedding vectors. A learnable classification (CLS) token is prepended to the sequence of patch embeddings (shown in darker blue). Positional embeddings are added to retain spatial information before feeding the sequence into a standard Transformer encoder. The output of the CLS token is passed through an MLP head to produce the final class prediction. The image is adpated from~\citep{dosovitskiy2021an}}
\label{fig:vit_architecture}
\end{figure}

The Vision Transformer extends the Transformer architecture to visual domains by treating image patches as input tokens \citep{dosovitskiy2020image}. As shown in Fig~\ref{fig:vit_architecture}, an image is split into a sequence of non-overlapping patches, each of which is linearly projected into a fixed-dimensional embedding. These embeddings are then augmented with learnable positional encodings and passed through a standard Transformer encoder. A classification token is appended, and its final representation is used for prediction through a multi-layer perceptron head (MLP). ViTs achieve competitive performance on benchmarks like ImageNet~\citep{deng2009imagenet}, highlighting the power of attention in learning global visual representations.

\subsection{Large Language Models}
LLMs are transformer architectures (Fig.~\ref{fig:transformer_architecture}) trained on large-scale text data. These models are typically classified into three architectural types: \textit{encoder-only}, \textit{decoder-only}, and \textit{encoder-decoder}. Each structure is optimized for different types of task in natural language processing and robotics applications.

\textit{Encoder-only} models, such as BERT, RoBERTa, and DeBERTa, utilize bidirectional self-attention mechanisms to learn deep contextual relationships from both the left and right of a token~\citep{ghojogh2024attention, zayyanu2024revolutionising}. These models are trained using masked language modeling (MLM) objectives and are well-suited for text classification, semantic similarity, and question answering. Their strength lies in representing sentence-level meaning rather than in generating sequential text.

\textit{Decoder-only} models, such as GPT-3, GPT-4, PaLM, and LLaMA, operate sequentially by predicting the next token in a sequence based only on previous tokens~\citep{ghojogh2024attention}. This unidirectional architecture is optimized for text generation, dialogue, summarization, and other open-ended generative tasks. 

\textit{Encoder-decoder} models, also known as sequence-to-sequence architectures, such as T5, BART, and the original Transformer, include an encoder to represent the input and a decoder to produce the output~\citep{liu2021enct5, brandisauskas2023seq2code, ghojogh2024attention}. These are ideal for tasks where full input processing is required before output generation, including machine translation, summarization, and code generation. Recent applications in robotics also use encoder-decoder models for instruction grounding and language-conditioned action planning.

\begin{figure}[t]
\centering
\includegraphics[width=\linewidth]{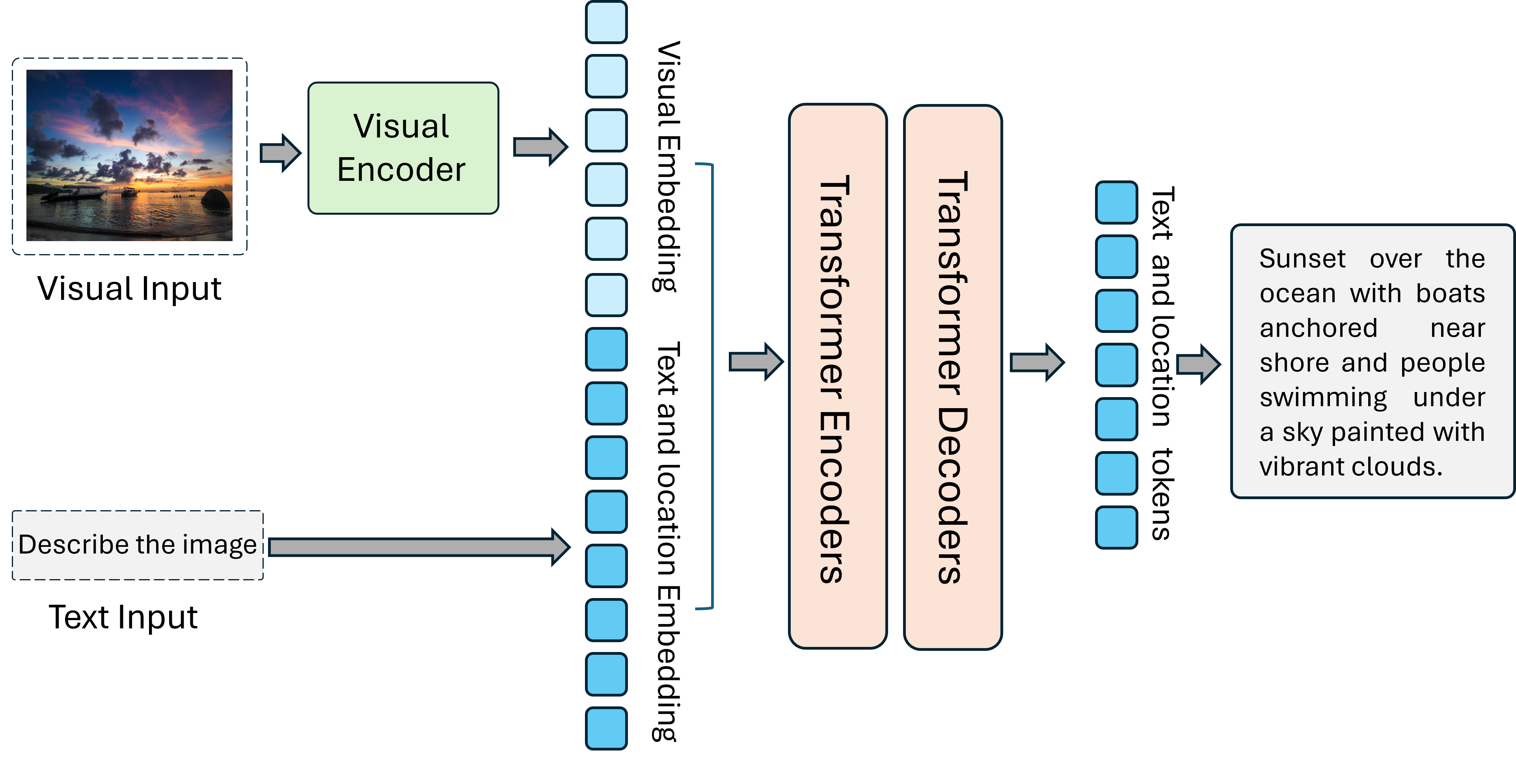}
\caption{Architecture of a VLM for image captioning and semantic understanding. Visual input is processed by a visual encoder to extract patch-based embeddings. In parallel, a text prompt (e.g., "Describe the image") is tokenized and embedded. These embeddings are fused and jointly processed through a Transformer-based encoder-decoder architecture. The model outputs a natural language caption that describes the semantic content of the image, enabling tasks such as captioning, question answering, and visual grounding. the image is adapted from~\citep{Xiao2024CVPR}}

\label{fig:vlm_architecture}
\end{figure}

\subsection{Vision Language Models}
VLMs use the synergy between computer vision and natural language processing to perform tasks such as image captioning, visual question answering, cross-modal retrieval, and instruction grounding. Like LLMs, they are built on the Transformer architecture, using either dual encoders (e.g., CLIP), unified encoder-decoder frameworks (e.g., BLIP, Flamingo), or sequence-to-sequence stacks. During training, they typically align visual and textual inputs using objectives such as contrastive learning (matching image-text pairs), masked modeling (predicting masked tokens or regions), or learning to generate captions from images.

Fig.~\ref{fig:vlm_architecture} illustrates a typical VLM designed for image captioning using a transformer-based encoder-decoder architecture. The process begins with an input image, which is divided into patches and embedded via a \textit{visual encoder}, such as a (ViT)~\citep{mishra2024image, lam2023deep}. A language prompt, e.g., "Describe the image", is also tokenized and embedded, either directly or via lightweight text guidance.
The visual tokens and prompt embeddings are passed through a \textit{Transformer encoder-decoder} stack. The encoder captures spatial and semantic relationships from the image, while the decoder generates output tokens autoregressively. This allows the model to produce a fluent and contextually accurate caption, as shown in Fig.~\ref{fig:vlm_architecture}. This architecture enables semantic-level reasoning over visual input and has shown strong performance on benchmark datasets such as Flickr8k and Flickr30k~\citep{mishra2024image, abdel2024image}.

Recent studies have demonstrated improved caption quality through the use of advanced modules or architectural variants within the encoder-decoder stack, such as Swin Transformers, T5-based decoders, or GPT-based language models, depending on the design~\citep{lam2023deep, mishra2024image, fouad2024image}. These approaches are widely used in robotic perception systems where visual scene understanding must be paired with natural language generation or instruction.

\section{Vision Language Action Models}\label{sec:vla}
VLA models represent a new frontier in robotic intelligence, enabling robots to perceive visual environments, understand natural language instructions, and execute grounded actions accordingly. These models bridge the semantic gap between multimodal inputs, such as images, sensor data, human commands, and low-level robotic control. VLA architectures are particularly relevant for unstructured and dynamic environments, where traditional rule-based programming is infeasible. They empower robots to generalize tasks such as object manipulation, navigation, and interaction by using deep learning, representation alignment, and sequential decision making~\citep{wang2024adversarial, guruprasad2024benchmarking, ladev2024}.

\begin{figure}[!t]
\centering
\includegraphics[width=0.9\linewidth]{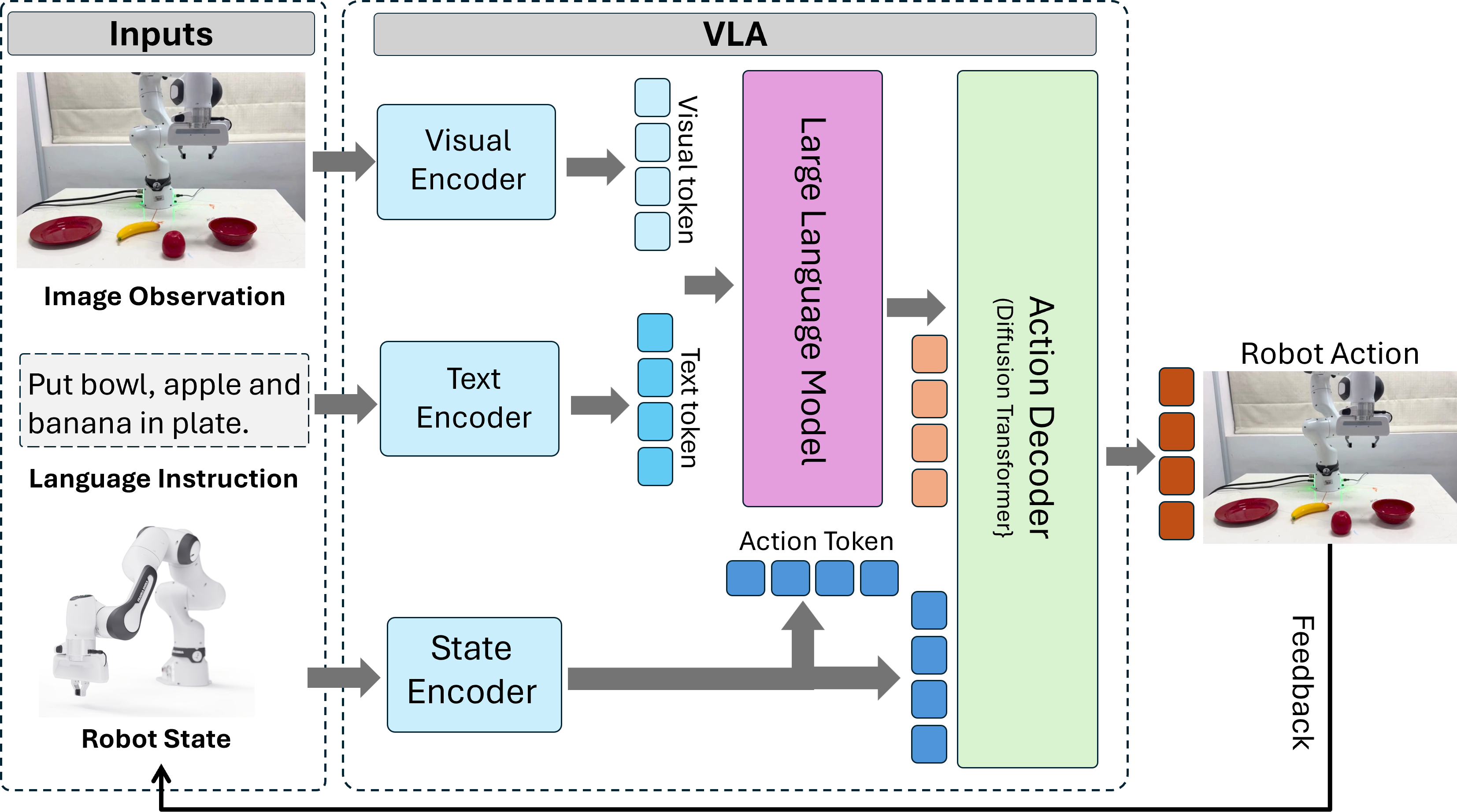}
\caption{Architecture of a VLA system for robotic manipulation. The model processes three inputs: an image of the scene, a natural language instruction, and the robot's internal state. These are encoded respectively through visual, text, and state encoders. The resulting embeddings are passed to an LLM that fuses multimodal information and generates a semantic representation of the intended task. This representation, along with robot state features, is processed by an Action Decoder implemented as a diffusion transformer to generate a trajectory that accomplishes the commanded task.}

\label{fig:vla_robot}
\vspace{-0.5cm}
\end{figure}

\subsection{VLA Architecture}

The VLA architecture illustrated in Fig.~\ref{fig:vla_robot} represents an end-to-end  framework that is a representative of the leading VLA systems such as; RT-2~\citep{cite:224}, OpenVLA~\citep{openvla}, CLIP-RT~\citep{kang2025cliprt}, Octo~\citep{team2024octo}, and RT-1~\citep{cite:18}, all of which employ transformer-based vision and language backbones fused through cross-modal attention. 

The architecture unifies three parallel encoder streams: visual, linguistic, and proprioceptive, into a single transmission diffusion backbone that directly generates control commands. First, a transformer-based \emph{Visual Encoder} (e.g. ViT~\citep{dosovitskiy2020image}, DINOv2~\citep{caron2021dino}) processes raw RGB (depth/semantic) images of the workspace and produces a sequence of fixed-length feature tokens. In parallel, a pre-trained \emph{Language Encoder} (for example, PaLM~\citep{chowdhery2022palm} or LLaMA~\citep{touvron2023llama}) tokenizes and embeds natural language instructions, whether high-level goals (e.g., "Put bowl, apple, and banana on plate.") or detailed stepwise instructions, in the same \(d\)-dimensional space. Similarly, \emph{State Encoder} embeds the proprioceptive and kinematic state of the robot (joint angles, pose of the end effector, gripper status) through an MLP or small transformer into additional tokens, allowing the model to reason about reachability, collision avoidance, and feedback correction. 

All tokens are concatenated and passed into a transformer-based model that produces \emph{action embeddings}. This model may implement either a diffusion policy, using a \emph{Diffusion Transformer} that iteratively denoises a noisy latent trajectory (e.g., as in Diffusion Policy~\citep{cite:34} or VLAFlow~\citep{black2024pi_0}), or a direct policy, which predicts the embeddings in a single pass without diffusion. At inference time, the action embeddings are converted into continuous control signals, such as end-effector velocities or joint torques, either through a lightweight output head or by completing the full diffusion sampling process. In some implementations, the embeddings can also be decoded into imagined next-frame images, enabling an "imagine-and-verify" loop for closed-loop execution.

Models such as OpenVLA and Octo further incorporate proprioceptive tokens, while several systems (e.g., PerAct~\citep{shridhar2022peract}, Helix~\citep{helix2025}) support real-time feedback loops for continual correction. The rapid evolution and plug-and-play modularity of these architectures, where one can swap in a stronger ViT, a larger language model, or a more expressive diffusion sampler are driving to a new direction of instruction-driven autonomy for generalist robotic systems.

\subsection{State-of-the-Art VLA Models}
Table \ref{Table-VLA} is organized to provide a brief yet thorough overview of over a hundred VLA models related to robotic manipulation and instruction-driven autonomy. The first two columns detail the name and year of publication of each model. The next two columns \textit{End-to-End} and \textit{Component Focused} flag whether a model learns a direct mapping from raw visual and language input to control commands or instead concentrates on developing individual building blocks (for example, a better vision backbone or a more effective action sampler). The \textit{Main Contributions} column then summarizes each work's core innovation, whether it introduced a novel fusion architecture, demonstrated a new training paradigm, or achieved state-of-the-art performance on a benchmark.

Each VLA model is based on four essential components: the training dataset, which provides foundational real-world task demonstrations or simulated episodes; the vision encoder, responsible for converting raw images or depth data into detailed feature maps; the language encoder, which maps instructions or annotations into a shared latent space; and the action decoder, which integrates these multimodal embeddings to produce the actual robot instructions, whether they are joint trajectories, discrete tokens, or overarching motion primitives. In Table \ref{Table-VLA}, the final column specifies which dataset each model was trained on, which vision backbone it uses (e.g. CLIP-ViT, ResNet, EfficientNet), which text encoder it employs (e.g. T5, LLaMA, CLIP text), and what kind of action decoder it relies upon (e.g. Transformer head, diffusion policy, CVAE sampler), making it straightforward to compare how different architectures assemble these building blocks.

\subsection{Architectural Trends}
\begin{figure*}[t]
    \centering
    \includegraphics[width=0.99\linewidth]{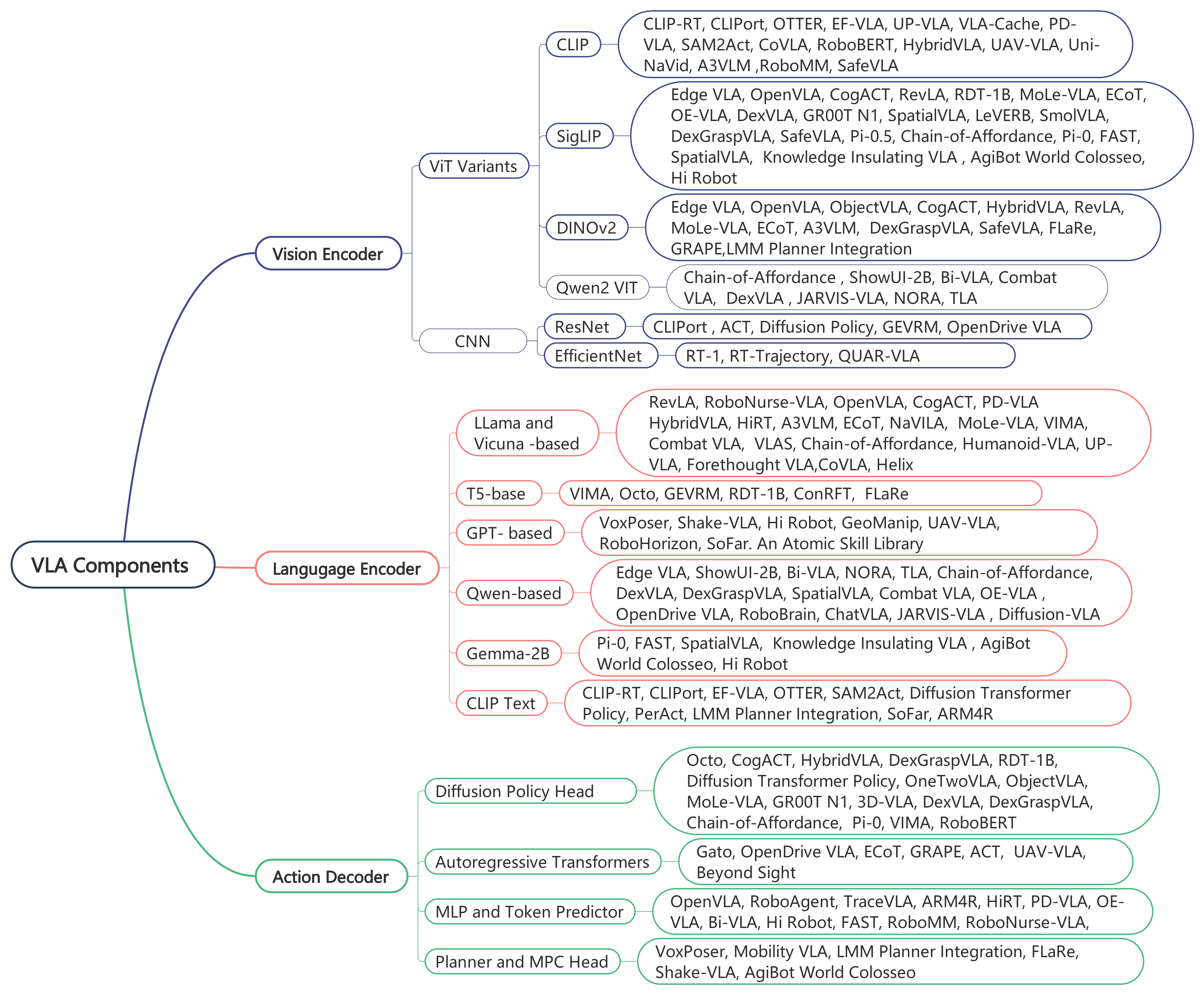}
    \caption{ This mind map presents the principal classes of vision encoders, language encoders, and action decoders employed in state-of-the-art VLA models. Only those encoder and decoder classes that are utilized by at least three different models are visualized, highlighting prevailing architectural trends across the VLA literature. The taxonomy categorizes representative models under each component family based on their dominant backbone; for example, ViT variants (such as; CLIP, SigLIP, DINOv2) and CNNs for vision encoding, LLaMA/Vicuna, T5-base, Qwen, and GPT-based models for language encoding, and diffusion or autoregressive transformers, MLP, and general Token predictors for action decoding. It should be noted that some models are listed under multiple encoder categories due to hybrid or fused architecture designs. For instance, models such as HybridVLA, OpenVLA, and DexGraspVLA appear under both SigLIP and DINOv2, as they integrate features from both backbones to enhance visual grounding and downstream task performance. This fusion-based design supports improved generalization, multi-view robustness, and more flexible multimodal alignment.
}\label{fig:encoder-decoder}
\end{figure*}

Fig.~\ref{fig:encoder-decoder} presents a comprehensive taxonomy of VLA model components, structured around three interconnected modules: vision encoders, language encoders, and action decoders. Within the vision encoder family, several prominent approaches exist. CLIP and SigLIP-based encoders are popular for their strong visual-text alignment through contrastive learning and are utilized in models such as CLIPort, RevLA, and Edge VLA. Other ViT variants such as DINOv2 and Qwen2 VIT, are used in models like Gato, Octo, HybridVLA, and Chain-of-Affordance for their ability to model long-range spatial dependencies and high-level visual semantics. CNN-based encoders such as ResNet and EfficientNet appear in models like CLIPort, ACT, RT-1, and QUAR-VLA.

{\LargeTableFont
\onecolumn

\rowcolors{2}{gray!20}{white}
\begin{longtable}{@{}p{1.7cm} p{1cm} p{1cm} p{4.8cm} p{6.8cm}@{}}\label{Table-VLA}\\
\caption{Comprehensive survey of VLA models developed for robotic manipulation and instruction-driven autonomy. The table lists each model's name, publication year, indicates whether it operates in an end-to-end fashion (mapping raw visual and language inputs directly to actions) or focuses on individual components (vision, language, or action modules), and summarizes its main technical contributions. The final column details the primary training datasets and core model components-including vision encoders, language encoders, and action decoders used by each method.}
\\
\toprule
\textbf{Model Name} & \textbf{End-to-End} & \textbf{Component Focused} & \textbf{Main Contributions } & \textbf{Training Dataset and Model Components} \\
\midrule
\endfirsthead

\toprule
\textbf{Model Name}  & \textbf{End-to-End} & \textbf{Component Focused} & \textbf{Main Contribution} & \textbf{Training Dataset and Model Components} \\
\midrule
\endhead
\\
\midrule \multicolumn{5}{r}{\textit{continue}} 
\\
\bottomrule
\endfoot

\bottomrule
\endlastfoot
CLIPort \citep{cite:157}
  & \ding{52} & \ding{52}
  & Pioneered the semantic grounding of visuomotor policies by integrating CLIP features into dense transport maps for precise pick-and-place.
  & Dataset: Self-collected visuomotor demos; Vision Encoder: CLIP-ResNet50 + Transporter-ResNet; Language Encoder: CLIP text encoder; Action Decoder: LingUNet \\

RT-1 \citep{cite:18}
  & \ding{52} & \ding{52}
  & Introduced a discretized action transformer for scalable multi-task kitchen manipulation.
  & Dataset: Self-collected RT-1-Kitchen; Vision Encoder: EfficientNet CNN; Language Encoder: Universal Sentence Encoder; Action Decoder: Discretized action transformer head \\

Gato \citep{cite:141}
  & \ding{52} & \ding{52}
  & Demonstrated a unified tokenization scheme across vision, language, and control tasks, achieving zero-shot transfer across domains.
  & Dataset: Self-collected multi-domain tasks; Vision Encoder: custom ViT; Language Encoder: Sentence Piece tokenizer; Action Decoder: Autoregressive Transformer \\

VIMA \citep{jiang2022vima}
  & \ding{52} & \ding{52}
  & Handled six distinct vision-language grounding tasks via a prompt-based multimodal policy.
  & Dataset: VIMA self-collected; Vision Encoder: Mask R-CNN; Language Encoder: T5-base; Action Decoder: Transformer policy head \\

PerAct \citep{shridhar2023perceiver}
  & \ding{52} & \ding{56}
  & Uses voxel-based representation with language conditioning for high-precision manipulation; operates directly on point cloud voxels.
  & Dataset: RLBench; Vision Encoder: Perceiver Transformer + voxel grid encoder; Language Encoder: CLIP text encoder; Action Decoder: Transformer voxel policy head \\

SayCan \citep{ahn2022saycan}
  & \ding{56} & \ding{52}
  & Combines language model planning with value function grounding in the real world; interprets high-level goals into executable robot actions.
  & Dataset: Self-collected everyday manipulation demos; Vision Encoder: none; Language Encoder: PaLM; Action Decoder: Value-conditioned execution module \\
RoboAgent \citep{bharadhwaj2023roboagent}
  & \ding{52} & \ding{56}
  & MT-ACT: multi-task transformer policy with semantically augmented CVAE encoding and action-chunking for strong real-world generalization.
  & Dataset: RoboSet teleop demos; Vision Encoder: Multi-view CNN encoder; Language Encoder: Semantic transformer encoder; Action Decoder: CVAE + Chunked trajectory predictor \\

RT-Trajectory \citep{gu2023rttrajectory}
  & \ding{52} & \ding{56}
  & Conditioned policies on user-sketched trajectories to improve generalization to novel layouts and paths.
  & Dataset: RT-1 dataset; Vision Encoder: EfficientNet-B3; Language Encoder: None; Action Decoder: Sketch-conditioned behavioral cloning policy \\

ACT \citep{cite:216}
  & \ding{52} & \ding{52}
  & Applied temporal ensembling to achieve smooth bimanual manipulation with 0.1 mm precision.
  & Dataset: self-collected demos on ALOHA; Vision Encoder: ResNet-18; Language Encoder: none ; Action Decoder: CVAE-Transformer head \\
RT-2 \citep{cite:224}
  & \ding{52} & \ding{52}
  & First large VLA co-finetuned on Internet VQA and robot data, unlocking emergent multi-robot zero-shot capabilities.
  & Dataset: Internet VQA + RT-1-Kitchen; Vision Encoder: PaLI-X/PaLM-E ViT; Language Encoder: PaLI-X/PaLM-E text encoder; Action Decoder: Symbol-tuning transformer \\

VoxPoser \citep{cite:78}
  & \ding{52} & \ding{52}
  & Achieved zero-shot constraint-aware motion planning by composing a frozen VLM and LLM without additional training.
  & Dataset: Self-collected motion demos+RLBench; Vision Encoder: OWL-ViT; Language Encoder: GPT-4; Action Decoder: MPC optimizer \\
CLIP-RT \citep{kang2024clip}
  & \ding{56} & \ding{52}
  & Contrastive policy using CLIP vision and text encoders to select language-based motion primitives, enabling fast learning and robust zero-shot transfer for table-top manipulation.
  & Dataset: OXE; Vision Encoder: CLIP image encoder (ViT-H-14); Language Encoder: CLIP text encoder; Action Decoder: Cosine similarity head over text-embedded motion primitives. \\

Diffusion Policy \citep{cite:34}
  & \ding{52} & \ding{52}
  & Introduced diffusion-based policy modeling for multimodal visuomotor action distributions.
  & Dataset: Self-collected demos; Vision Encoder: ResNet-18; Language Encoder: None; Action Decoder: diffusion policy network \\

Octo \citep{team2024octo}
  & \ding{52} & \ding{52}
  & First generalist diffusion policy trained on 4 M+ trajectories across 22 robot platforms, demonstrating broad transfer.
  & Dataset: Open X-Embodiment; Vision Encoder: CNN encoder ; Language Encoder: T5-base; Action Decoder: Diffusion Transformer head \\

VLATest \citep{cite:182}
  & \ding{56} & \ding{52}
  & Automated framework for large-scale VLA model testing, revealing robustness gaps and guiding improvements in manipulation.
  & Dataset: none ; Vision Encoder: that used in (e.g., OpenVLA,RT-1,Octo); Language Encoder: that used in (e.g., OpenVLA,RT-1,Octo); Action Decoder: that used in (e.g., OpenVLA,RT-1,Octo) \\

NaVILA \citep{cite:32}
  & \ding{52} & \ding{56}
  & Hierarchical planning yields 88\% real-world navigation success for legged robots via language-conditioned topological control.
  & Dataset: Self-collected Real-world legged robot demos; Vision Encoder: VILA Vision Encoder ; Language Encoder: VILA LLM; Action Decoder: Topological graph planner + RL policy for joint commands
 \\

RoboNurse-VLA \citep{cite:103}
  & \ding{52} & \ding{56}
  & Real-time voice-to-action pipeline for surgical instrument handover, robust to unseen tools in dynamic scenes.
  & Dataset: Self-collected Surgical handover videos + voice prompts; Vision Encoder: SAM2 ; Language Encoder: LLaMA-2; Action Decoder: token-based action decoder \\

Mobility VLA \citep{cite:35}
  & \ding{52} & \ding{56}
  & Multimodal instruction navigation with topological mapping for robust long-range mobility.
  & Dataset: MINT instruction tours; Vision Encoder: Gemini 1.5 Pro based (ViT); Language Encoder: Gemini 1.5 Pro based text encoder; Action Decoder: Topological graph-based planner \\

RevLA \citep{cite:39}
  & \ding{52} & \ding{56}
  & Domain adaptation adapters to improve the generalization of robotic foundation models across visual domains.
  & Dataset: Open X-Embodiment (OXE);  Vision Encoder: DINO-v2 + SigLIP; Language Encoder: LLama-7B; Action Decoder: Llama head, outputs 7 discrete action tokens \\

Uni-NaVid \citep{cite:210}
  & \ding{52} & \ding{56}
  & Video-based VLA unifying embodied navigation tasks across multiple benchmarks.
  & Dataset: Room-to-Room (R2R) + REVERIE; Vision Encoder: EVA-CLIP; Language Encoder: Vicuna-7B; Action Decoder: Vicuna-7B head (4 discrete action tokens) \\

RDT-1B \citep{cite:112}
  & \ding{52} & \ding{52}
  & 1.2B-parameter diffusion foundation model excelling at bimanual manipulation and zero-shot generalization.
  & Dataset: self-collected 6K ALOHA episodes; Vision Encoder: SigLIP ; Language Encoder: T5-XXL ; Action Decoder: Diffusion Transformer + MLP decoder \\

RoboMamba \citep{cite:111}
  & \ding{52} & \ding{56}
  & Mamba-based unified VLA with linear-time inference for real-time robotic reasoning.
  & Dataset: SAPIEN sim benchmarks + real-world demos; Vision Encoder: Mamba VLM visual backbone; Language Encoder: Mamba VLM text backbone; Action Decoder: MLP policy head for SE(3) pose predicting \\

Chain-of-Affordance \citep{cite:100}
  & \ding{56} & \ding{52}
  & Sequential affordance reasoning for spatial planning, achieving SOTA on LIBERO dataset.
  & Dataset: LIBERO + real/sim manipulation tasks; Vision Encoder: Qwen2-VL; Language Encoder: Qwen2-VL; Action Decoder: Diffusion policy head \\

Edge VLA \citep{cite:19} 
  & \ding{52} & \ding{52}
  & Lightweight, edge-optimized VLA for low-power real-time inference.
  & Dataset: OXE + Bridge robotics set; Vision Encoder: SigLIP + DINOV2; Language Encoder: Qwen2; Action Decoder: Non-autoregressive control head \\

OpenVLA \citep{cite:94}
  & \ding{52} & \ding{52}
  & LORA-fine-tuned open-source VLA achieving efficient transfer and high success.
  & Dataset: OXE + DROID robot data; Vision Encoder: DINOv2 + SigLIP; Language Encoder: Llama 2; Action Decoder:  Llama 2 output head (predicts discretized action tokens as output)\\

CogACT \citep{cite:102}
  & \ding{52} & \ding{52}
  & Componentized diffusion action transformer, +59.1\% success over OpenVLA with specialized adaptation.
  & Dataset: OXE subset + real trials; Vision Encoder: DINOv2 + SigLIP; Language Encoder: LLaMA-2; Action Decoder: Diffusion Transformer head \\

ShowUI-2B \citep{cite:108}
  & \ding{52} & \ding{52}
  & GUI/web navigation via screenshot grounding and efficient token selection.
  & Dataset: 256 K GUI instruction demos; Vision Encoder: Qwen2-VL-2B ViT; Language Encoder: Qwen2-VL-2B LLM; Action Decoder: Qwen2-VL-2B output head (GUI actions as tokens) \\

Pi-0 \citep{cite:14}
  & \ding{52} & \ding{56}
  & General robot control flow model for high-frequency, open-world tasks.
  & Dataset: Extended OXE called Pi-Cross-Embodiment; Vision Encoder: PaliGemma (SigLIP); Language Encoder: PaliGemma (Gemma-2B); Action Decoder: diffusion-based Flow matching action expert head  \\

HiRT \citep{zhang2024hirt}
  & \ding{52} & \ding{52}
  & Hierarchical planning/control separation, doubling execution speed and improving dynamic task success.
  & Dataset: Self collected Real-world data; Vision Encoder: InstructBLIP; Language Encoder: LLaMA-2; Action Decoder: Latent-conditioned policy head (MLP)\\	

A3VLM \citep{huang2024a3vlm}
  & \ding{56} & \ding{52}
  & Learns articulation-aware affordance grounding purely from RGB video, generalizing to unseen object joints.
  & Dataset: PartNet-Mobility; Vision Encoder: CLIP, DINOv2, Q-Former (fused); Language Encoder: LLaMA2; Action Decoder: Parameterized primitive motion generator \\

SVLR \citep{samson2024svlr}
  & \ding{56} & \ding{52}
  & Modular "segment-to-action" pipeline using visual prompt retrieval for on-device policy execution.
  & Dataset: Self-collected visual prompts; Vision Encoder: Mini InternVL; Language Encoder: Phi-3mini4k; Action Decoder: Script-based action binder \\

Bi-VLA \citep{gbagbe2024bivla}
  & \ding{56} & \ding{52}
  & Dual-arm instruction-to-action planner grounded in recipe demonstrations, achieving 83.4 \% real-world task success.
  & Dataset: Visual-recipe demos; Vision Encoder: Qwen-VL; Language Encoder: Starling-LM; Action Decoder: Python trajectory generator \\

QUAR-VLA \citep{cite:43}
  & \ding{52} & \ding{56}
  & Quadruped-specific VLA with adaptive gait and body command mapping, strong sim-to-real transfer.
  & Dataset: QUART locomotion + manipulation; Vision Encoder:  EfficientNet-B3; Language Encoder: FiLM / VLM tokenizer; Action Decoder: Transformer decoder (discrete tokens) \\

3D-VLA \citep{zhen2024threedvla}
  & \ding{52} & \ding{52}
  & Integrates 3D generative diffusion heads for world reconstruction, enabling planning in RGB+D and point-cloud spaces.
  & Dataset: 3D-language-action pairs; Vision Encoder: 3D-aware transformer; Language Encoder: 3D-LLM; Action Decoder: Multi-head diffusion planner \\ 
  
RoboMM \citep{yan2024robomm}
  & \ding{52} & \ding{52}
  & MIM-based multimodal decoder unifying 3D perception and language for spatially aligned policy fusion.
  & Dataset: RoboData (CALVIN, Meta-World); Vision Encoder: Multi-view CLIP + occupancy network; Language Encoder: Flamingo-style fusion module; Action Decoder: Multimodal MLP/attention decoder \\
FAST \citep{cite:133}
  & \ding{52} & \ding{52}
  & Frequency-space action tokenization for up to 15 times faster inference on general robot control.
  & Dataset: DROID; Vision Encoder: PaliGemma (SigLIP); Language Encoder: PaliGemma (Gemma-2B); Action Decoder: FAST token generator \\
  
OpenVLA-OFT \citep{cite:93}
  & \ding{52} & \ding{52}
  & Optimized fine-tuning of OpenVLA with parallel chunked decoding, achieving 97.1 \% success on LIBERO dataset and 26 time speed-up.
  & Dataset: LIBERO; Vision Encoder: SigLIP + DINOv2; Language Encoder: LLaMA-27B; Action Decoder: Llama 2 Parallel chunking head \\

CoVLA \citep{cite:5}
  & \ding{52} & \ding{56}
  & VLA model for autonomous driving, trained on richly annotated scene data for robust planning.
  & Dataset: Large-scale driving videos + annotations; Vision Encoder: CLIP ViT; Language Encoder: LLaMA-2; Action Decoder: Trajectory prediction module \\

ORION \citep{cite:56}
  & \ding{52} & \ding{56}
  & Holistic end-to-end driving VLA aligning semantic understanding with generative trajectory control.
  & Dataset: E2E driving benchmark; Vision Encoder: EVA-02-L (Transformer); Language Encoder: Vicuna v1.5 (LoRA); Action Decoder: Generative planner head \\

UAV-VLA \citep{cite:150}
  & \ding{52} & \ding{56}
  & Zero-shot aerial mission VLA combining satellite and UAV imagery for scalable instruction-driven flight planning.
  & Dataset: Satellite + UAV flight logs; Vision Encoder: Molmo-7B-D (CLIP); Language Encoder: GPT-3; Action Decoder: Transformer-based path planner \\  

Combat VLA \citep{cite:29}
  & \ding{52} & \ding{56}
  & Ultra-fast tactical reasoning in 3D ARPG environments, achieving 50 times faster inference and human-level success.
  & Dataset: 3D ARPG combat logs; Vision Encoder: Qwen2.5-VL-3B; Language Encoder: Qwen2.5-VL-3B; Action Decoder: LLM-based planner head \\

HybridVLA \citep{cite:110}
  & \ding{52} & \ding{56}
  & Adaptive ensemble decoding that combines diffusion and autoregressive policies for robust multi-task generalization.
  & Dataset: RT-X trajectories + synthetic task fusion; Vision Encoder: CLIP ViT + DINOV2; Language Encoder: LLaMA-2; Action Decoder: Diffusion policy head\\

NORA \citep{cite:79}
  & \ding{52} & \ding{56}
  & Low-overhead VLA with integrated visual reasoning and FAST token decoding for real-time performance.
  & Dataset: OXE; Vision Encoder: Qwen-2.5-VL; Language Encoder: Qwen-2.5-VL; Action Decoder: FAST tokenizer head \\

SpatialVLA \citep{cite:136}
  & \ding{52} & \ding{56}
  & 3D spatial encoding and adaptive action discretization to improve cross-robot manipulation generality.
  & Dataset: OXE; Vision Encoder: SigLIP; Language Encoder: PaliGemma (Gemma-2B); Action Decoder: Adaptive action grid head \\

MoLe-VLA \citep{cite:213}
  & \ding{52} & \ding{56}
  & Selective layer activation in a multi-stage ViT yields 5.6 time faster inference and +8\% task success.
  & Dataset: RLBench + real-world trials; Vision Encoder: DINOv2, SigLIP; Language Encoder: LLaMA-2; Action Decoder: Diffusion head \\

JARVIS-VLA \citep{cite:101}
  & \ding{52} & \ding{56}
  & Open-world instruction following in 3D games via keyboard/mouse action prediction.
  & Dataset: Minecraft gameplay demos; Vision Encoder: ViT (in Llava-Next/Qwen2-VL); Language Encoder: Llava-Next/Qwen2-VL (transformer LLMs); Action Decoder: Key/mouse control head \\

UP-VLA \citep{cite:209}
  & \ding{52} & \ding{56}
  & Precise 3D spatial reasoning, achieving +33 \% success on the CALVIN benchmark.
  & Dataset: CALVIN; Vision Encoder: CLIP-ViT; Language Encoder: Phi-1.5; Action Decoder: MLP policy head \\

Shake-VLA \citep{cite:92}
  & \ding{52} & \ding{52}
  & Modular bimanual VLA achieving 100\% success on cluttered cocktail-mixing tasks.
  & Dataset: Cocktail mixing demos; Vision Encoder: YOLOv8, EasyOCR; Language Encoder: GPT-4o,Whisper-1; Action Decoder: Bimanual arm controller \\

MORE \citep{cite:214}
  & \ding{56} & \ding{56}
  & Scalable Mixture of Expert (MoE) enhanced RL framework for quadruped multi-task learning.
  & Dataset: Quadruped navigation/manipulation demos; Vision Encoder: CLIP-like; Language Encoder: Fuyu 8B ; Action Decoder: Mixture-of-Experts + LoRA adapter \\

DexGraspVLA \citep{cite:219}
  & \ding{52} & \ding{56}
  & Diffusion-based dexterous grasping with $\geq90\%$ zero-shot success across diverse objects.
  & Dataset: Self-collected Dexterous grasp data; Vision Encoder: DINOv2; Language Encoder: Qwen-VL, Qwen2.5-VL; Action Decoder: Diffusion policy head \\
  
DexVLA \citep{cite:185}
  & \ding{52} & \ding{56}
  & Cross-embodiment diffusion expert enabling rapid adaptation without per-task tuning.
  & Dataset: OXE, RLBench; Vision Encoder: Qwen2-VL (ViT), ResNet-50; Language Encoder:Qwen2-VL,DistilBERT; Action Decoder: Diffusion Transformer head \\

Humanoid-VLA \citep{cite:42}
  & \ding{52} & \ding{56}
  & Hierarchical VLA for full-body humanoid control, integrating perception and latent action planning.
  & Dataset: Self-collected humanoid robot episodes; Vision Encoder: Video Visual Encoder,Cross-Attention; Language Encoder: Llama3-70B; Action Decoder: Token-based Motion Decoder + RL Whole-Body Ctrlr \\

ObjectVLA \citep{cite:223}
  & \ding{52} & \ding{56}
  & End-to-end open-world object manipulation without task-specific data.
  & Dataset: RoboSpatial manipulation episodes; Vision Encoder: DinoX, DiVLA VLM; Language Encoder: DiVLA (LLM backbone); Action Decoder: Object-centric diffusion controller head \\

Gemini Robotics \citep{gemini2025robotics}
  & \ding{52} & \ding{52}
  & General-purpose VLA built on the Gemini 2.0 foundation, enabling long-horizon dexterous manipulation across diverse robot embodiments with zero-shot adaptability.
  & Dataset: Self-collected ALOHA2 demos + web-scale VL Dataset; Vision Encoder: Gemini 2.0 vision component; Language Encoder: Gemini 2.0 language component; Action Decoder: Local zero-shot policy head \\

ECoT \citep{zawalski2025ecot}
  & \ding{52} & \ding{52}
  & Embodied chain-of-thought planning for interpretable, stepwise VLA control.
  & Dataset: Bridge v2 ; Vision Encoder: SigLIP, DINOv2; Language Encoder: LLaMA-2 7B; Action Decoder: Autoregressive VLA decoder with CoT module \\

OTTER \citep{huang2025otter}
  & \ding{52} & \ding{52}
  & Zero-shot generalization via a frozen CLIP backbone and causal transformer action decoding.
  & Dataset:  LIBERO; Vision Encoder: Frozen CLIP ViT; Language Encoder: CLIP text encoder; Action Decoder: Causal transformer delta-trajectory head \\

$\pi$-0.5 \citep{black2025pi05}
  & \ding{52} & \ding{52}
  & Hierarchical VLA co-trained on real robot demos and web-scale vision-language data for robust household task generalization.
  & Dataset: self-collected 400h robot teleop + web VL dataset; Vision Encoder: SigLIP; Language Encoder: Gemma (2B/2.6B); Action Decoder: Flow Matching Head \\

OneTwoVLA \citep{lin2025onetwo}
  & \ding{52} & \ding{52}
  & Unified reasoning-acting framework that dynamically toggles between planning and control via decision tokens.
  & Dataset: Self-collected 16K reasoning-augmented robot episodes; Vision Encoder: same as pi-0 vla; Language Encoder:same as pi-0 vla; Action Decoder: Diffusion policy head\\

Helix \citep{helix2025}
  & \ding{52} & \ding{52}
  & First 200 Hz VLA for full humanoid control on embedded systems, enabling zero-shot task transfer.
  & Dataset: self-collected 200Hz teleop + sim logs; Vision Encoder: Pretrained VLM ; Language Encoder: Pretrained VLM ; Action Decoder: Fast transformer policy \\

Gemini Robotics On-Device \citep{parada2025geminiondevice}
  & \ding{52} & \ding{52}
  & On-device optimized variant of Gemini VLA, delivering low-latency dual-arm and humanoid control on embedded hardware.
  & Dataset: Self-collected ALOHA2 + few-shot adaptation demos; Vision Encoder: Gemini SDK vision module; Language Encoder: Gemini SDK language module; Action Decoder: On-device optimized policy head \\

OE-VLA \citep{zhao2025oevla}
  & \ding{52} & \ding{52}
  & Curriculum-tuned LLaVA backbone with interleaved multimodal prompting for improved generalization across vision-language-action tasks.
  & Dataset: CALVIN; Vision Encoder: SigLIP-400M ViT; Language Encoder: Qwen-1.5 language module; Action Decoder: MLP token generator \\
SmolVLA \citep{shukor2025smolvla}
  & \ding{52} & \ding{52}
  & Ultra-lightweight VLA trained on community-contributed robot demonstrations, capable of real-time inference on CPU.
  & Dataset: 22.9K community episodes; Vision Encoder: SigLIP (VLM-2) visual backbone; Language Encoder: SmolVLM2 text backbone; Action Decoder: Chunked flow-matching head\\ 

EF-VLA \citep{anonymous2025efvla}
  & \ding{52} & \ding{52}
  & Early fusion of fine-grained CLIP visual tokens into the language-action pipeline, boosting zero-shot generalization.
  & Dataset: Self-collected real and simulated tasks; Vision Encoder: Frozen CLIP ViT; Language Encoder: Frozen CLIP text encoder; Action Decoder: causal transformer \\

PD-VLA \citep{song2025accelerating}
  & \ding{52} & \ding{56}
  & First parallel decoding method with action chunking for VLA, achieving a 2.52 times speed-up without sacrificing control fidelity.
  & Dataset: Chunked trajectory demonstrations; Vision Encoder: CLIP-ViT-Large-Patch14-336 (LLaVA); Language Encoder: Vicuna-7B-v1.5 (LLaVA); Action Decoder: Fixed-point token predictor \\

LeVERB \citep{xue2025leverb}
  & \ding{52} & \ding{52}
  & Dual-process latent VLA for whole-body humanoid control, achieving 58.5 \% success on sim-to-real humanoid demos.
  & Dataset: sim-to-real humanoid demos; Vision Encoder: SigLIP ViT; Language Encoder: SigLIP text encoder; Action Decoder: Latent CVAE verb + transformer policy \\

TLA \citep{hao2025tla}
  & \ding{52} & \ding{52}
  & First language-grounded tactile-action model for high-precision contact tasks, with 85 \% success on peg-in-hole task.
  & Dataset: TLA Data; Vision Encoder: ViT (Qwen2-VL); Language Encoder: Qwen2-VL; Action Decoder: Multimodal $\Delta x/\Delta y/\Delta z$ predictor \\

Interleave-VLA \citep{fan2025interleave}
  & \ding{52} & \ding{52}
  & Model-agnostic wrapper enabling interleaved image-text instruction processing..
  & Dataset: Interleave-VLA data; Vision Encoder: Any base VLM (e.g., OpenVLA); Language Encoder: Any LLM (e.g., Pi-0); Action Decoder: Minimal interleaved-processing module \\

iRe-VLA \citep{guo2025irevla}
  & \ding{52} & \ding{52}
  & Iterative RL and supervised fine-tuning pipeline for robust control and rapid generalization across embodiments.
  & Dataset: Franka-Kitchen, real Panda robot demos; Vision Encoder: BLIP-2 (pre-trained VLM); Language Encoder: BLIP-2; Action Decoder: MLP action head after token learner \\

TraceVLA \citep{zheng2025tracevla}
  & \ding{52} & \ding{52}
  & Visual trace prompting to incorporate spatio-temporal cues, boosting task success by 3.5 time over OpenVLA.
  & Dataset: OXE + 150K trace-annotated demos; Vision Encoder: Phi-3-Vision with trace overlay; Language Encoder: Phi-3 LLM; Action Decoder: Quantized delta-motion tokens \\

OpenDrive VLA \citep{zhu2025opendrivevla}
  & \ding{52} & \ding{56}
  & End-to-end driving VLA with semantic scene alignment and temporal abstraction for robust trajectory planning.
  & Dataset: Autonomous driving QA/planning benchmarks; Vision Encoder: ResNet-101 + Query Transformers; Language Encoder: Qwen2.5-Instruct (LLM); Action Decoder: Ego-vehicle action autoregressor \\

V-JEPA 2 \citep{assran2025v}
  & \ding{52} & \ding{56}
  & Dual-stream self-supervised video JEPA enabling predictive planning in vision-language-action tasks.
  & Dataset: Droid video data; Vision Encoder: ViT (self-supervised) ; Language Encoder: LLM for QA/alignment; Action Decoder: Action-conditioned transformer predictor head \\

Knowledge Insulating VLA \citep{driess2025knowledge}
  & \ding{52} & \ding{56}
  & Implements insulation layers between vision-language and action modules, accelerating training and inference while maintaining generalization.
  & Dataset: Multi-domain VL datasets; Vision Encoder: PaliGemma (SigLIP); Language Encoder: PaliGemma (Gemma-2B) encoder; Action Decoder: Diffusion Modular policy head \\

GR00T N1 \citep{bjorck2025gr00t}
  & \ding{52} & \ding{56}
  & Self-collected Diffusion-based foundation model enabling unified humanoid control with policy tokenization.
  & Dataset: Multi-modal humanoid demonstrations; Vision Encoder: SigLIP-2 ViT (Eagle-2 VLM); Language Encoder: SmolLM2 (Eagle-2 VLM); Action Decoder: Generative diffusion transformer based planner \\

AgiBot World Colosseo \citep{bu2025agibot}
  & \ding{52} & \ding{56}
  & Integrates multiple embodied datasets into a unified platform for scalable training and evaluation of VLA models.
  & Dataset: AgiBot World Data; Vision Encoder: PaliGemma (SigLIP); Language Encoder: PaliGemma (Gemma-2B); Action Decoder: Latent action planner + policy head \\

Hi Robot \citep{shi2025hi}
  & \ding{52} & \ding{56}
  & Hierarchical separation of planning and control for open-ended instruction following in complex environments.
  & Dataset: Self-collected Instruction-following data; Vision Encoder: PaliGemma-3B (SigLIP); Language Encoder: PaliGemma-3B (Gemma-2B); Action Decoder: Flow-Matching Action Expert  \\

EnerVerse \citep{huang2025enerverse}
  & \ding{52} & \ding{56}
  & World-model LLM for predictive future-space modeling, enabling long-horizon manipulation planning.
  & Dataset: self-collected Synthetic task fusion data; Vision Encoder:Pretrained VAE + Diffusion Generator; Language Encoder: Tokenized instruction prompt; Action Decoder: Diffusion Policy Head \\

FLaRe \citep{hu2024flare}
  & \ding{52} & \ding{56}
  & Large-scale RL fine-tuning framework generating robust, adaptive robot policies across domains.
  & Dataset: Multi-domain RL demonstrations; Vision Encoder: DinoV2; Language Encoder: Transformer policy (language tokens); Action Decoder: RL policy head \\

Beyond Sight \citep{jones2025beyond}
  & \ding{52} & \ding{56}
  & Fuses heterogeneous sensor modalities via language-grounded attention to improve VLA generalization.
  & Dataset: self-collected Multi-sensor data; Vision Encoder: Multi-modal ViT; Language Encoder: Transformer (shared, task language input); Action Decoder: Transformer action head  \\

GeoManip \citep{tang2025geomanip}
  & \ding{52} & \ding{56}
  & Encodes geometric constraints as model interfaces, enhancing robustness and precision in manipulation.
  & Dataset: Self-collected Simulated geometry tasks; Vision Encoder: VLM (GPT-4o) + Grounding-DINO; Language Encoder: GPT-4o; Action Decoder: Constraint solver head \\

Universal Actions \citep{zheng2025universal}
  & \ding{52} & \ding{56}
  & Defines a universal action dictionary to standardize policy transfer and improve cross-task adaptability.
  & Dataset: Self-collected Cross-domain manipulation demos; Vision Encoder: Shared VLM (LLaVA-OneVion-0.5B); Language Encoder: LLaVA; Action Decoder: Unified action tokenizer head \\
RoboHorizon \citep{chen2025robohorizon}
  & \ding{52} & \ding{56}
  & LLM-enhanced multi-view environment modeling for robust long-horizon task planning.
  & Dataset: Self-collected Multi-view robot trajectories ; Vision Encoder: Multi-view transformer (ViT); Language Encoder: GPT‐based planner; Action Decoder: DreamerV2 Actor-Critic RL Head \\

SAM2Act \citep{fang2025sam2act}
  & \ding{52} & \ding{56}
  & Utilizes SAM-based segmentation prompts with memory-augmented VLA for improved object-centric manipulation.
  & Dataset: SAM-labeled manipulation tasks; Vision Encoder: SAM2 segmentation encoder; Language Encoder: CLIP text encoder; Action Decoder: Memory-augmented policy head \\

LMM Planner Integration \citep{li2025integrating}
  & \ding{52} & \ding{56}
  & Merges LMM-based strategic planning with 3D skill policies for generalizable manipulation.
  & Dataset: skill library demos; Vision Encoder: DINO (2D semantics) + PointNext (3D); Language Encoder: CLIP Language Encoder; Action Decoder:3D Transformer head \\

VLA-Cache \citep{xu2025vla}
  & \ding{52} & \ding{56}
  & Introduces token-caching to reuse computation across time steps, boosting inference efficiency.
  & Dataset: LIBERO; Vision Encoder: CLIP ViT; Language Encoder: LLaMA-2; Action Decoder: Cached inference head \\

Forethought VLA \citep{wu2025foresight}
  & \ding{52} & \ding{56}
  & Aligns latent vision and action spaces for foresight-driven policy steering.
  & Dataset: Self-collected Latent alignment demonstrations; Vision Encoder: Phi-3 Vision; Language Encoder: LLama; Language; Action Decoder: Diffusion policy  \\

GRAPE \citep{zhang2024grape}
  & \ding{52} & \ding{56}
  & Preference-guided policy adaptation via personalized feedback alignment.
  & Dataset: Self-collected Preference-labeled demos; Vision Encoder: Dinov2; Language Encoder: LLaMA-2; Action Decoder: Autoregressive transformer head \\

HAMSTER \citep{li2025hamster}
  & \ding{52} & \ding{56}
  & Hierarchical skill decomposition to sequence multi-step manipulation actions.
  & Dataset: Self-collected Decomposed manipulation tasks; Vision Encoder: VILA-1.5-13B; Language Encoder: VILA-1.5-13B; Action Decoder: Robotic View Transformer Skill execution head \\
TempoRep VLA \citep{myers2025temporal}
  & \ding{52} & \ding{56}
  & Use successor representation temporal encoding for compositional action planning.
  & Dataset: Self-collected Temporal demonstration sequences; Vision Encoder: ResNet-34 CNN; Language Encoder: retrained transformer (CLIP-style); Action Decoder: MLP (3x256) head on ResNet feature \\

ConRFT \citep{chen2025conrft}
  & \ding{52} & \ding{56}
  & Applies consistency regularized fine-tuning with reinforcement for stable policy learning.
  & Dataset: Self-collected data for fine-tuning; Vision Encoder: same as in octo; Language Encoder:same as in octo; Action Decoder: Reinforced policy head \\
RoboBERT \citep{wang2025robobert}
  & \ding{52} & \ding{56}
  & Unified multimodal Transformer for end-to-end vision-language-action manipulation, pre-trained on diverse robot and language data.
  & Dataset: Self-collected Multi-domain robot demos; Vision Encoder: CLIP ViT; Language Encoder: BERT-base; Action Decoder: CNN-based Diffusion Policy Head \\

Diffusion Transformer Policy \citep{hou2024diffusion}
  & \ding{52} & \ding{56}
  & Adapts diffusion-based transformer architectures to VLA policy learning, enabling robust multimodal action sampling.
  & Dataset: LIBERO + CALVIN; Vision Encoder: DINOv2; Language Encoder: CLIP Text Encoder; Action Decoder: Diffusion generator head \\

GEVRM \citep{zhang2025gevrm}
  & \ding{52} & \ding{56}
  & Generative video modeling of goal-oriented tasks to enhance planning for visual manipulation.
  & Dataset: CALVIN; Vision Encoder: ResNet-34; Language Encoder: T5 Encoder; Action Decoder: Diffusion Policy \\

SoFar \citep{qi2025sofar}
  & \ding{52} & \ding{56}
  & Introduces successor-feature orientation representations bridging spatial reasoning and robotic manipulation.
  & Dataset: Self-collected Orientation task demonstrations; Vision Encoder: Florence-2 (ViT-style), SAM; Language Encoder: CLIP Text Encode; Action Decoder: VLM (e.g., LLaVA or GPT-4o) for 6D goal pose, then motion planner \\

ARM4R \citep{niu2025pre}
  & \ding{52} & \ding{56}
  & Auto-regressive 4D transition model for predicting and planning manipulator trajectories.
  & Dataset: 76K
videos from the Epic-Kitchens100 dataset ; Vision Encoder: ViT-Base; Language Encoder: CLIP text encoder; Action Decoder:  2-layer MLP \\

Magma \citep{yang2025magma}
  & \ding{52} & \ding{56}
  & Foundation multimodal agent model unifying vision, language, and action domains for end-to-end control.
  & Dataset: Self-collected Multimodal interaction dataset; Vision Encoder: ConvNeXt-XXlarge; Language Encoder: LLaMA-3-8B (decoder-only LLM); Action Decoder: Decoder-Only LLM Head (LLaMA-3-8B) \\

An Atomic Skill Library \citep{li2025atomic}
  & \ding{52} & \ding{56}
  & Constructs an atomic skill library for modular, data-efficient composition of robotic actions.
  & Dataset: Self-collected Skill primitive demonstrations; Vision Encoder: Prismatic VLM (scene description.), DINO-X (obj detection), SAM-2 (segmentation); Language Encoder: Prismatic, GPT-4 (for planning); Action Decoder: Skill executor module \\

VLAS \citep{zhao2025vlas}
  & \ding{52} & \ding{56}
  & Integrates speech-based LLM guidance for customizable voice-driven vision-language-action control.
  & Dataset: Speech-guided robot demos; Vision Encoder: CLIP ViT; Language Encoder: Vicuna (LLaMA-7B/13B); Action Decoder: Vicuna as an Autoregressive Action Decoder \\

ChatVLA \citep{zhou2025chatvla}
  & \ding{52} & \ding{52}
  & Unified conversational VLA enabling natural-language and vision-driven interactive robot control with real-time multimodal feedback.
  & Dataset: Interactive human-robot demos; Vision Encoder: ViT + LoRA; Language Encoder: Qwen2-VL-2B (LLM); Action Decoder: mixture-of-expert action head, as in DiVLA) \\

RoboBrain \citep{ji2025robobrain}
  & \ding{52} & \ding{56}
  & Knowledge-grounded policy brain that maps abstract high-level plans to concrete multimodal actions across diverse tasks.
  & Dataset: Multi-domain robot and plan data; Vision Encoder: SigLIPr; Language Encoder: Qwen2.5-7B-Instruct (decoder-only LLM); Action Decoder: LoRA adapters for skill \\

SafeVLA \citep{zhang2025safevla}
  & \ding{52} & \ding{56}
  & Safety-aware VLA integrating constraint feedback through safe RL to ensure collision-free, reliable manipulation.
  & Dataset: Safety-scenario demonstrations; Vision Encoder: Modular (DINOv2, SigLIP, CLIP); Language Encoder: LLM (model-agnostic, e.g., T5, LLaMA, Qwen); Action Decoder: Safety-constraint policy head \\

CognitiveDrone \citep{lykov2025cognitivedrone}
  & \ding{52} & \ding{56}
  & Embodied cognitive reasoning VLA for UAVs, combining vision-language understanding with autonomous flight planning.
  & Dataset: UAV mission logs; Vision Encoder: OpenVLA visual encoder	(ViT-style) ; Language Encoder: OpenVLA’s language encoder; Action Decoder: Transformer policy head \\
  
Diffusion-VLA \citep{wen2024diffusionvla}
& \ding{52} & \ding{52}
& Multimodal VLA framework unifying vision-language reasoning with diffusion-based policy for robust, generalizable manipulation across diverse robot embodiments.
& Dataset: Multi-embodiment manipulation suites; Vision Encoder: SigLIP; Language Encoder: Qwen2-VL (2B/7B/72B); Action Decoder: Latent diffusion policy head + MLP \\
\hline
\end{longtable}
 \twocolumn
}

Language encoders show similar architectural diversity. LLaMA and Vicuna-based language encoders are widely used in models such as RevLA, OpenVLA, and HybridVLA for instruction understanding and zero-shot reasoning. T5-style models appear in VIMA, Octo, and FLARE, offering flexible encoder-decoder structures for sequence generation. GPT-based and Qwen-based encoders, such as those used in VoxPoser, Edge VLA, and DexVLA, balance generalization and compact deployment. Gemma-2B language encoders are found in Pi-0 and FAST, while specialized solutions like CLIP Text encoders are used in CLIPort and PerAct for minimal alignment tasks.

In action decoders, diffusion-based transformers are a leading choice for models like Octo, HybridVLA, and DexGraspVLA, as they offer fine-grained, temporally smooth control via iterative denoising. Autoregressive Transformer heads, such as those in Gato, OpenDrive VLA, and GRAPE, generate action sequences step-by-step, optimizing real-time responsiveness. Several models including VoxPoser, LMM Planner Integration, and FLARE embed Model Predictive Control or specialized planning heads to support decision-making in dynamic tasks. MLP or token predictor heads are used in OpenVLA, TraceVLA, and RoboMamba for efficient low-level control.

Our evaluation of VLA architectures shows a wide variety of algorithms in each core component. For the vision encoder, the majority of models used CLIP and SigLIP-based ViT backbones to extract rich, semantically aligned visual features. In the language domain, the LLaMA family (including LLaMA-2 and its larger variants) is most commonly used, appearing in several models as the primary text encoder. When it comes to action decoding, diffusion-based Transformer heads are the most popular choice with their ability to model complex, multimodal action distributions. On the data side, while many teams rely on privately collected manipulation demonstrations, the publicly available Open X-Embodiment dataset is the most commonly used.

\begin{figure}[t]
    \centering
    \includegraphics[width=0.95\linewidth]{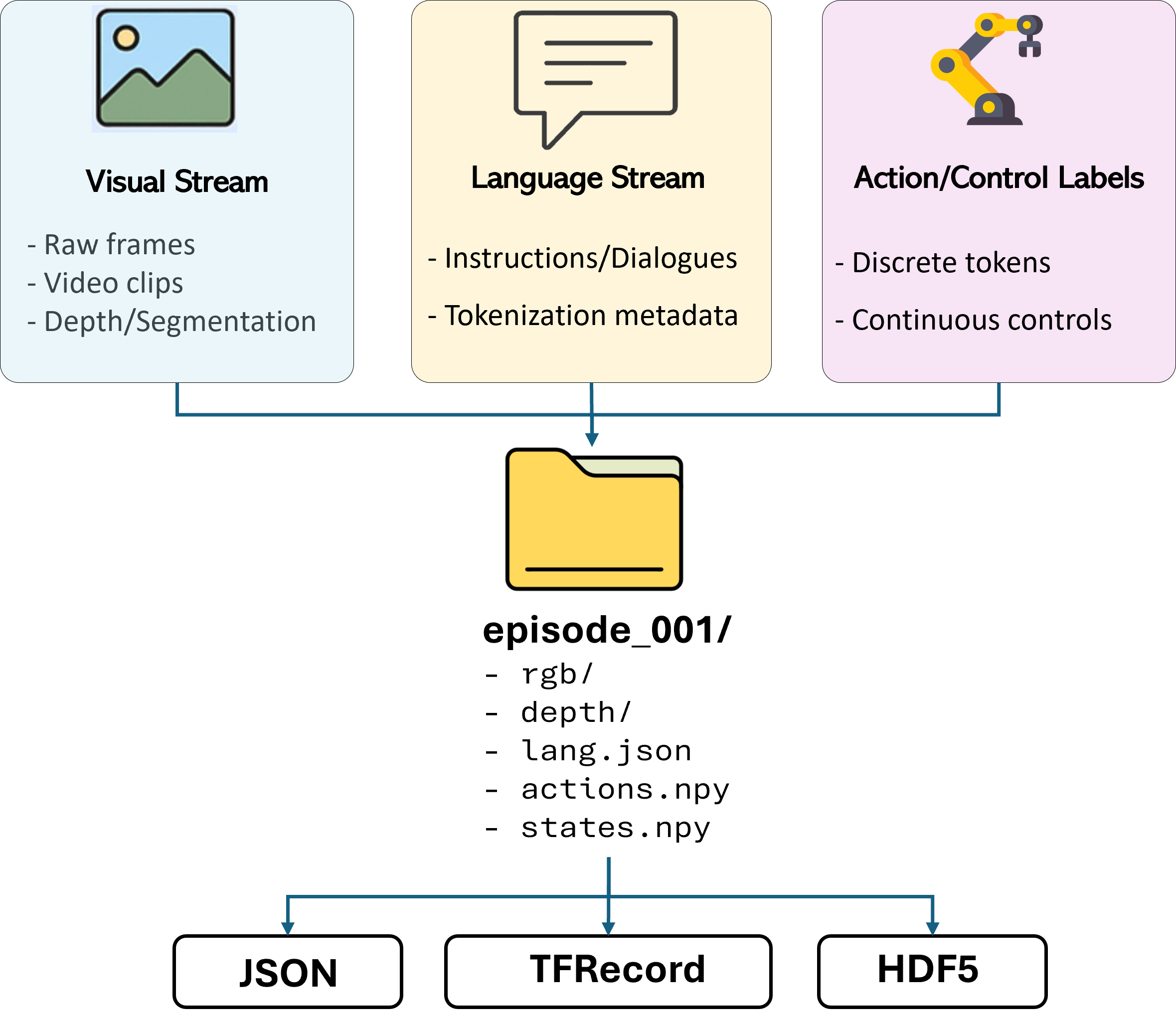}
  \caption{Schematic of the unified VLA training data format. Visual observations, language instructions, and action/control signals are grouped into episode directories and serialized into standardized storage formats (JSON, TFRecord, and HDF5), enabling efficient and scalable data loading for end-to-end model training.}
    \label{fig:proj}
    \vspace{-0.5cm}
\end{figure}
\section{VLA Training Datasets}\label{sec:dataset}
The foundation of VLA models lies in high quality, diverse training datasets. These datasets are crucial as they expose models in the complete range of real and simulated environments, ensuring a tight alignment of visual elements, natural language instructions, and control~\citep{james2020rlbench}. These datasets allow VLAs to learn complex cross-modal correlations, such as how language complexities (e.g., 'gently place') affect motion smoothness without relying on manually prepared heuristics. We first introduce the unified dataset schema that underlies the VLA training pipelines, then survey the most influential public datasets, and finally apply a comprehensive benchmarking strategy to assess the scale, modality coverage, and complexity of each dataset.

\subsection{ Dataset Format}

 Structured overview of the general dataset format commonly used in VLA training pipelines. is illustrated in Fig.~\ref{fig:proj}. It highlights the systematic organization of multimodal data into three primary streams: visual, language, and action/control, which collectively facilitate the training and evaluation of VLA models.

The \textit{Visual Streams} comprise raw RGB frames, video snippets, and optionally, depth maps and segmentation masks. These visual inputs provide essential spatial and contextual data to perception modules in VLA architectures. Typically, the data in this stream is stored in standard image or video formats like JPEG or PNG for individual frames and MP4 or similar formats for video snippets.

The \textit{Language Streams}  incorporate natural-language instructions or dialogues alongside tokenization metadata (such as token offset). These textual annotations are essential for instructing and conditioning robotic actions and are commonly stored in lightweight, structured formats such as JSON or plain text files. The presence of tokenization metadata facilitates efficient textual processing, enabling direct integration with transformer-based language models.

The \textit{Action/Control Labels} include discrete action tokens (e.g. categorical commands like "move forward" or "turn left") and continuous control vectors representing joint positions or end-effector trajectories. These action labels provide explicit supervision signals for model output and are typically stored as NumPy arrays or encoded within structured data containers.

All three modality streams are systematically integrated into standardized episode-level directories (e.g., \texttt{episode/}), where visual data reside in subdirectories such as \texttt{rgb/} and \texttt{depth/}, accompanied by \texttt{lang.json}, \texttt{actions.npy}, and \texttt{states.npy}. Each episode folder can then be serialized in formats like: \textit{JSON} for lightweight, human-readable metadata; \textit{TFRecord}/\textit{TF-Example} for high-throughput, sharded training; or \textit{HDF5} for efficient random access to synchronized frames, actions, and state arrays, enabling balance readability, I/O performance, and scalability in their training pipelines.

\subsection{Major VLA Datasets}

Table~\ref{tab:vla_datasets} summarizes the progress of VLA datasets, highlighting how each dataset advances autonomy by varying in scale, modality, and task complexity. Early collections such as EmbodiedQA and R2R focus on discrete decision making in constrained environments, offering simple state-action mappings suitable for evaluating baseline policy architectures (e.g., PACMAN, Speaker-Follower). As we move into 2020-2022, datasets like ALFRED, RLBench, and CALVIN introduce longer-horizon tasks and richer sensory streams combining RGB, depth, proprioception, and natural language instructions to stress test hierarchical planning and subgoal decomposition methods (e.g., C2F-ARM, VIMA, RT-2). These mid-generation datasets bridge the gap between symbolic planners and end-to-end learning, enabling comparative analyses of model-based control versus learned policies under simulated dynamics.

{\SmallTableFont
\onecolumn
\rowcolors{2}{gray!20}{white}

\begin{longtable}{>{\raggedright\arraybackslash}m{3cm}
                  p{3cm}
                  p{4cm}
                  p{5.5cm}}
\caption{Overview of main VLA datasets used in robotic manipulation and embodied AI research. For each dataset, we list the release year, dataset size, distinctive characteristics, and the data storage format.} \label{tab:vla_datasets} \\
\hline
\textbf{Dataset} 
  & \textbf{Size} 
  & \textbf{Distinctive Characteristics} 
  & \textbf{Data Format} \\
\hline
\endfirsthead

\hline
\textbf{Dataset} 
  & \textbf{Size} 
  & \textbf{Distinctive Characteristics} 
  & \textbf{Data Format} \\
\hline
\endhead

  EmbodiedQA~\citep{Das_2018_CVPR}
& 5,000+ QA episodes across 750+ 3D scenes
& Goal-directed visual question answering in House3D with object and room diversity
& JSON-formatted question-answer pairs, egocentric RGB frame sequences, and agent trajectories\\

R2R~\citep{anderson2018vision}
& 7,189 unique paths with 21,567 natural language instructions
& Real-world vision-language navigation using Matterport3D with path diversity and crowd-sourced instructions
& JSON files with instructions and navigation paths; panoramic JPEG frames and viewpoint graph metadata\\

ALFRED~\citep{shridhar2020alfred}
    & 8,055 expert demonstrations with 25,743 language directives
    & Language-conditioned household manipulation tasks in AI2-THOR 2.0 across 120 indoor scenes
    & Per-demonstration folders with egocentric RGB frames, ground-truth interaction masks, and language annotations; metadata and action/state sequences in JSON format\\
RLBench~\citep{james2020rlbench}
& Expert demonstrations available for 100 vision-based manipulation tasks
& Large-scale few-shot and imitation learning benchmark in PyRep simulation
& Pickled demos include \texttt{joint\_positions}, \texttt{camera\_images}, \texttt{task\_description}, and proprioceptive states.\\

CVDN (NDH)~\citep{cvdn}
& 7,415 navigation-from-dialog-history instances from 2,050 dialogs
& Vision-and-Dialog Navigation benchmark requiring agents to act based on dialog history
& JSON annotations with dialog turns, navigation paths, image features, and speaker roles\\

TEACh~\citep{teach}
& 3,047 successful two-agent gameplay sessions
& Multiturn dialog-driven household task completion in AI2-THOR
& JSON transcripts aligned to visual frames, with egocentric RGB, object masks, action logs, and benchmark CSV splits \\

DialFRED~\citep{gao2022dialfred}
& 53,000+ human-annotated QA pairs across 34,000+ tasks
& Dialogue-enabled embodied instruction following on augmented ALFRED subgoals
& Per-task \texttt{dialogue.json} with human and oracle QAs, action traces, subgoal templates, and frame alignment \\

 Ego4D~\citep{ego4d} 
&  3,670 h of first-person video 
&  Large-scale, real-world egocentric dataset with diverse scenarios and modalities 
&  MP4 video clips; JSON-based narrations and annotations; HDF5/LMDB indices; multilingual narration files. \\

  CALVIN~\citep{calvin} 
&   5,000+ demonstrations 
&   Long-horizon, language-conditioned robotic manipulation tasks 
&   HDF5 archives with synchronized RGB-D frames, proprioception, action sequences, and natural language instructions.\\

  DROID~\citep{droid}
&   76k demonstrations; 564 scenes, 86 tasks
&   High-diversity language-conditioned robot manipulation 
&   RLDS-formatted RGB-D data, stereo video, camera calibrations, language annotations, and robot state/action logs. \\

 Open X-Embodiment~\citep{openx}
&  1M+ trajectories, 500+ skills, 22 robot types
&  Large-scale, multi-embodiment, multi-skill manipulation dataset
&  Sharded TFRecords with RGB/depth frames, language instructions, action vectors; YAML metadata and RLDS format. \\
 
RoboSpatial~\citep{robospatial}
& 1M images, 5K 3D scans, 3M spatial QA pairs
& 2D-3D paired spatial reasoning dataset
& RGB images, 3D scans, and relational graph annotations in support of spatial understanding benchmarks. \\

CoVLA~\citep{covla}
& 83.3 h real-world driving video, 6M frames
& Time-aligned vision-language-action dataset
& Synchronized RGB video, GPS/IMU-based trajectories, and auto-generated captions using rule-based and VLM methods.\\

TLA~\citep{hao2025tla}
& 30K contact-rich peg-in-hole demonstrations
& Tactile-language-action alignment for precise insertion and assembly
& ROS bag files with synchronized \texttt{camera/}, \texttt{tactile/}, \texttt{lang.json}, and \texttt{trajectory.csv} recordings. \\

BridgeData V2~\citep{walke2023bridgedata}& 60,096 trajectories (50,365 teleoperated; 9,731 scripted)
& Multi-skill goal- and language-conditioned manipulation dataset
& TFRecords with RGB images, natural language instructions, and continuous 7-DoF action vectors; includes both human and scripted demonstrations. \\

LIBERO~\citep{liu2023libero}
& 130 tasks: 10 spatial, 10 object, 10 goal, 100 lifelong
& Lifelong VLA benchmark for procedural and declarative knowledge transfer
& JSON and Parquet files with RGB images, language instructions, action trajectories, and structured metadata. \\

 Kaiwu~\citep{kaiwu}
& 1M multimodal robotic episodes
& Real-world, multi-embodiment dataset for dexterous manipulation with natural language commands
& Per-episode HDF5 files with synchronized RGB, depth, 3D skeletons, tactile, EMG, gaze, IMU, audio, language, and motion capture data. \\

PLAICraft~\citep{he2025plaicraftlargescaletimealignedvisionspeechaction}
& 10,000 + hours of multiplayer Minecraft gameplay across 5 modalities
& Open-ended multiplayer interaction with emergent task structures and voice-aligned social play
& JSON-encoded multimodal streams (RGB, audio, keyboard, mouse) with millisecond alignment\\

  AgiBot World~\citep{bu2025agibot} 
& 1M+ multimodal dual-arm trajectories 
& Open-source platform for long-horizon generalist policy learning
& ROS-based: RGB-D, fisheye, tactile, proprioception, language, error annotations, and dexterous control logs. \\

Robo360~\citep{liang2023robo360} 
& 2K+ real trajectories, 86 calibrated views, 100+ diverse objects
& Multimodal dataset for dynamic NeRF, imitation learning, and control
& Synchronized RGB videos, depth maps, audio, robot proprioception, and control signals per frame. \\

REASSEMBLE~\citep{sliwowski2025reassemble}
& 4,551 contact-rich demonstrations, 17 objects, 781 minutes
& Multimodal (RGB, audio, event, force/torque, proprioception)
& Synchronized multistream recordings from RGB cameras, event camera, microphones, force/torque sensors, and proprioceptive signals, collected during haptically teleoperated assembly and disassembly tasks based on NIST benchmark boards. \\

RoboCerebra~\citep{han2025robocerebra}
& 100K long-horizon trajectories across 1K+ tasks
& System-2-level reasoning and generalization in real-world-scale settings
& Structured plan logs, visual transitions, failure annotations, and dense subtask labels from human-verified demonstrations and multi-stage task generation. \\

IRef-VLA~\citep{zhang2025irefvla}
& 11.5K rooms, 7.6M spatial relations, 4.7M instructions
& Imperfect referential grounding in 3D indoor scenes
& Per-room scene graphs, free-space maps, and affordance annotations with synthetic and imperfect language queries. \\

Interleave-VLA~\citep{fan2025interleave}
& 210K episodes (13M frames)
& Interleaved vision-language instruction execution
& Mixed-format episodes with images, sketch overlays, and text prompts aligned with action sequences. \\

RoboMM~\citep{yan2024robomm}
& 30K simulated episodes + 5K real-world trials
& Multimodal fusion of vision, language, proprioception, and touch
& HDF5 per episode with \texttt{rgb/}, \texttt{depth/}, \texttt{tactile.csv}, \texttt{instructions.json}, and \texttt{action\_sequence.json}.\\

ARIO~\citep{zhou2024ario}
& 50K simulated episodes + 5K real-world trials
& Contact-rich manipulation with tactile, audio, and proprioceptive feedback
& Per-episode HDF5 archives with \texttt{rgb/}, \texttt{depth/}, \texttt{tactile.csv}, \texttt{instructions.json}, and \texttt{action\_sequence.json}. \\

\hline
\end{longtable}
\twocolumn
}  

From 2023 onward, the field shifts to truly multimodal control challenges. Datasets such as DROID and Open X-Embodiment embed synchronized RGBD, language, and multi-skill trajectories, facilitating evaluation of sensor fusion strategies and real-time feedback controllers. Large-scale egocentric corpora like Ego4D and CoVLA offer real-world visual streams that drive research on robust perception-action loops under unpredictable dynamics. Recent contact-rich datasets such as ARIO, TLA, RoboMM, and REASSEMBLE integrate high-frequency haptic and force/torque feedback alongside vision and language, enabling fine-grained impedance control and hybrid model-predictive schemes for deformable-object manipulation. Highly multimodal and large-scale datasets such as Kaiwu, PLAICraft, AgiBot World, and Robo360 support open-ended, long-horizon, and real-world tasks with diverse sensor suites including tactile, audio, proprioceptive, and multi-view data.
By standardizing annotation formats (HDF5 bundles, ROS bags, TFRecords) and pairing each collection with representative baselines (e.g., SayCan, HapticBERT, MM-FusionNet), Table~\ref{tab:vla_datasets} provides a detailed overview of these datasets across a continuum of task complexity, modality richness, and real-world fidelity.

\subsection{Benchamrk VLA Datasets}

\begin{figure*}[t]
    \centering
    \includegraphics[width=1\linewidth]{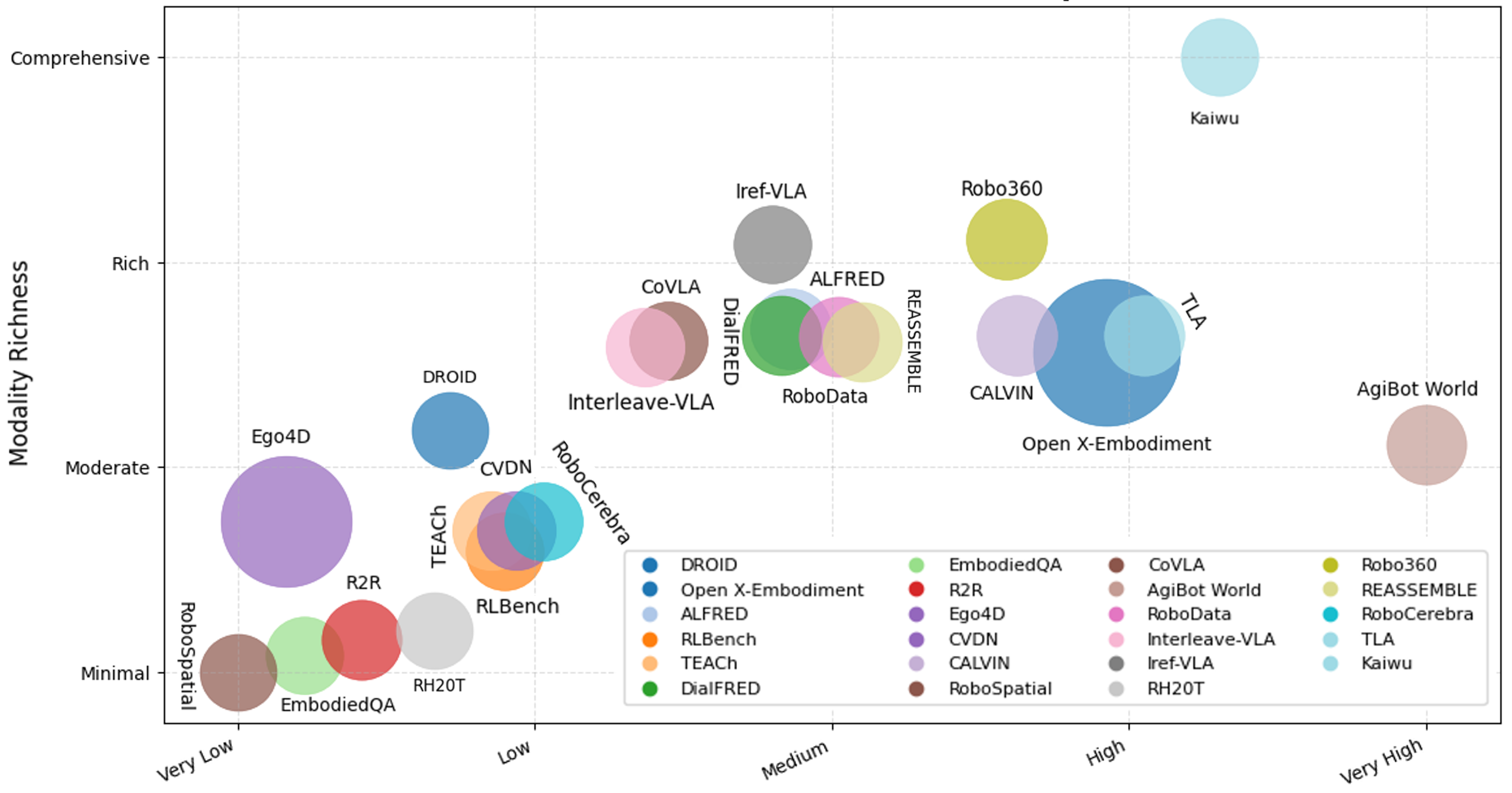}
 
    \caption{Benchmarking VLA Datasets by Task Complexity and Modality Richness. Each bubble represents a VLA dataset, positioned according to its normalized task-complexity score (x-axis) and its modality-richness score (y-axis). The bubble area is proportional to the dataset scale that is number of annotated episodes or interactions.  The script to generate the plot can be found here:\url{https://github.com/Muhayyuddin/VLAs}}

  \label{fig:vla_landscape}
\end{figure*}

In order to benchmark, we map each major VLA dataset onto a two-dimensional plane spanned by task complexity and modality richness, illustrated in Fig.~\ref{fig:vla_landscape}. The \hbox{x-axis} captures how challenging each dataset's manipulation tasks are, ranging from simple single-step actions to long-horizon, multiskill sequences. The y-axis shows the modality richness, from minimal (dual modalities: text and image) to comprehensive (up to seven modalities including audio, video, robot proprioception, control, depth, haptics, and language).

To quantify these dimensions systematically, we assign scalar scores to each dataset reflecting their task complexity and modality richness.
Task complexity, denoted as $\mathcal{C}_{\text{task}}$, incorporates:
\begin{itemize}
    \item Average number of low-level actions per episode ($T$). This captures how many primitive control commands are grouped together in a typical task (e.g., grasp, lift, move).
    \item Number of distinct high-level skills ($S$). This enumerates different semantic subtasks (e.g., open drawer, pick object).
    \item Degree of sequential task dependency ($D$). This denotes the fraction of tasks that require strict ordering of subtasks; $D \in [0,1]$.
    \item Linguistic abstraction level ($L$). Quantifies the average linguistic complexity (e.g., vocabulary size or syntactic depth) of the instruction set; $L \in \mathbb{R}^+$.
\end{itemize}
These attributes are integrated via the following weighted model:
\begin{equation}
\mathcal{C}_{\text{task}}(\mathcal{D}) = \alpha_1 \log(1 + T) + \alpha_2 S + \alpha_3 D + \alpha_4 L,
\end{equation}
where $\alpha_i > 0$ for $i = 1, \ldots, 4$ are weights that normalize each term to commensurate scales and can be tuned to reflect the emphasis on action length, skill diversity, sequential structure, or language complexity. For our benchmark, we set all weights equal to one.

Modality richness, captured by the score $\mathcal{C}_{\text{mod}}$, integrates four factors reflecting the scope and quality of sensory input:
\begin{itemize}
    \item Number of distinct modalities ($M$), Such as vision, depth, haptics, and language.
    \item Mean quality $Q = \tfrac{1}{M} \sum_{i=1}^M Q_{m_i}$, Where $Q_{m_i}$ for $i = 1, \dots, M$ are the modality-specific quality scores. $Q_{m_i}$ can be determined by expert annotation, automated signal-to-noise ratio analysis, or based on dataset documentation and previous benchmark studies. For this work, we use a mixture of empirical review and published specifications to assign scores in the range $[0.6, 0.95]$, reflecting the typical range of public datasets.
    \item Fidelity of temporal alignment across modalities ($A$), Measures how tightly modalities are synchronized (e.g., frame-accurate vision-language pairing), with \hbox{$A \in [0,1]$}.
    \item Presence of reasoning-critical modalities ($R$): Such as object masks or scene graphs that enable higher-level reasoning, $R \in \{0,1\}$.
\end{itemize}
This scoring mechanism is formalized as:
\begin{equation}
\mathcal{C}_{\text{mod}} = \beta_1 M + \beta_2 Q + \beta_3 A + \beta_4 R,
\end{equation}
where modality sensitivity weights $\beta_i > 0$ for $i = 1, \ldots, 4$ tune the relative importance of modality count, signal quality, temporal alignment, and reasoning-enabled annotations. For our benchmark, we set all weights equal to one.

Finally, to allow direct comparison across heterogeneous benchmarks, both raw scores are normalized. Task complexity to a standardized $[1,5]$ scale and modality richness to a $[2,5]$ scale. This mapping ensures interpretability: datasets with the lowest complexity or modality richness receive a score of 1 or 2 ("Very Low"/"Minimal") and the highest receive 5 ("Very High"/"Comprehensive"), with intermediate values reflecting proportional positioning. The bubble size then encodes the relative dataset scale (e.g., number of episodes or hours), providing an at-a-glance summary of both range and comprehensiveness across the leading VLA benchmarks.

The resulting visualization effectively categorizes datasets, while it also highlights critical gaps in current benchmark, offering notably the underrepresented region combining highly complex tasks with extensive multimodal integration. This gap underscores a promising direction for future dataset development aimed at advancing truly generalist robotic agents capable of complex, real-world perception and planning.

\subsection{Benchmarking Analysis}

Fig.~\ref{fig:vla_landscape} shows that most current VLA benchmarks are concentrated within a task complexity range from very low to high on the x-axis and span from minimal to rich modality along the y-axis. Early navigation and QA datasets like EmbodiedQA, R2R, and RoboSpatial are characterized by their very low complexity and minimal modality, reflecting simple, discrete decision-making in constrained environments. In contrast, mid-generation collections such as RLBench, TEACh, Ego4D, CVDN, and RoboCerebra, tend to feature low to moderate complexity with moderate modality richness, often focused on navigation, imitation, or basic manipulation tasks involving a limited number of modalities.

As the field evolves, datasets such as ALFRED, DialFRED, CoVLA, Interleave-VLA, RoboData, and REASSEMBLE have moved into the medium-complexity, rich-modality region by incorporating additional sensory streams like depth, language, and proprioceptive signals, enabling more sophisticated evaluation of policy learning and multi-step planning. In particular, a small subset of datasets, including Iref-VLA, Robo360, TLA, CALVIN, and Open X-Embodiment, simultaneously achieve high task complexity and rich modality, each with a particular focus. Robo360 on multiview real-robot visual fidelity, Iref-VLA on referential grounding in 3D scenes, TLA on tactile-language-action alignment for contact-rich assembly, CALVIN on long-horizon language-conditioned robotic manipulation, and Open X-Embodiment on multirobot, multiskill demonstrations at scale.

The only dataset positioned at the extreme of both axes is Kaiwu, which achieves very high task complexity alongside the most comprehensive modality richness, integrating vision, depth, language, proprioception, haptics, and additional streams. Meanwhile, AgiBot World stands out in the very high complexity quadrant while exhibiting just moderate modality diversity, emphasizing large-scale, long-horizon dual-arm tasks rather than maximal sensor integration. This disparity highlights a critical gap: current VLA benchmarks do not yet fully integrate the challenges of long-horizon, multi-skill control with exhaustive multimodal input (vision, depth, language, proprioception, haptics, audio, and scene graphs). Without such datasets, the development of robust and generalist robotic agents remains limited. Future efforts should therefore focus on the upper right quadrant of the landscape, creating new VLA benchmarks that maximize both task difficulty and multimodal diversity to accelerate progress toward general-purpose embodied intelligence.

{
\rowcolors{2}{gray!20}{white}      
\begin{table*}[t]
  \caption{Overview of simulation platforms commonly used for generating and evaluating VLA datasets. The table summarizes each simulator's supported sensory modalities, primary use cases, core capabilities, and the datasets that rely on them. These tools span diverse domains such as photorealistic indoor navigation, dexterous manipulation, and large-scale reinforcement learning, with varying degrees of physics realism.}
  \label{tab:sim_tools}
  \scriptsize
  \begin{tabular}{
      >{\raggedright\arraybackslash}m{2.5cm}
      p{3cm}
      p{3.3cm}
      p{3.3cm}
      p{3.3cm}
    }
    \hline
    \textbf{Simulator}
      & \textbf{Modalities}
      & \textbf{Use Cases}
      & \textbf{Capabilities}
      & \textbf{Datasets}
    \\ 
    \hline

AI2-THOR~\citep{kolve2017ai2} \textbf{[\href{https://ai2thor.allenai.org}{link}]}
      & RGB, depth, semantic/instance segmentation, object states
      & Embodied navigation, object manipulation
      & Photorealistic indoor scenes; procedural scene generation; physics-based object/agent interaction; built-in interaction APIs; language and task integration
      & ALFRED~\citep{shridhar2020alfred}, TEACh~\citep{teach}, DialFRED~\citep{dialfred}
\\

Habitat~\citep{savva2019habitat} \textbf{[\href{https://aihabitat.org}{link}]}
  & RGB, depth, semantic segmentation, agent pose
  & Vision-language navigation, embodied QA, point-goal navigation
  & Photorealistic, high-performance rendering; large-scale 3D scene support; modular sensor and agent APIs
  & R2R~\citep{anderson2018vision}, CVDN~\citep{cvdn}, EmbodiedQA~\citep{Das_2018_CVPR}
\\

NVIDIA Isaac Sim~\citep{nvidiaisac} \textbf{[\href{https://developer.nvidia.com/isaac-sim}{link}]}
  & RGB, depth, LiDAR, semantic/instance segmentation, bounding boxes, point clouds, physics states, force/torque, event camera
  & RL \& control, sim-to-real transfer, synthetic dataset generation, embodied AI, multi-robot simulation, digital twin, warehouse \& industrial robotics
  & Physically accurate PhysX dynamics; RTX-accelerated photorealistic rendering; procedural scene generation; domain randomization; noise models; ROS/ROS2 and Python API; cloud deployment 
  & Open X-Embodiment~\citep{openx}, Isaac Gym, RLBench, custom synthetic datasets
\\

Gazebo~\citep{koenig2004design} \textbf{[\href{http://gazebosim.org}{link}]}
  & RGB, depth, LiDAR, IMU, joint states, force/torque, contact, GPS
  & Control algorithm development, multirobot coordination, sim-to-real transfer, embodied navigation and manipulation
  & Open-source; plugin-based extensibility; realistic multi-physics engines; ROS1/ROS2 integration; multi-robot, multi-sensor support
  & RoboSpatial~\citep{robospatial}, 
\\

PyBullet~\citep{coumans2016pybullet} \textbf{[\href{https://pybullet.org}{link}]}
  & RGB, depth, contact forces, joint states
  & RL, robotic manipulation prototyping, physics-based simulation
  & Real-time physics; Python API; cross-platform; easy scripting; supports robotics and VR
  & QUAR-VLA~\citep{quarvla}, various custom RL/manipulation datasets
\\

CoppeliaSim~\citep{rohmer2013v} \textbf{[\href{https://www.coppeliarobotics.com}{link}]}
  & RGB, depth, joint states, force/torque sensors, proximity sensors
  & Multi-robot coordination, task scripting, manipulation, education
  & Multiple built-in physics engines; remote APIs (Python, ROS, C++); graphical scene editor; flexible scripting
  & RLBench~\citep{james2020rlbench},  CALVIN~\citep{calvin}
\\

Webots~\citep{michel2004webots} \textbf{[\href{https://cyberbotics.com}{link}]}
  & RGB, depth, sound, GPS, proximity, IMU, lidar, joint states
  & Mobile navigation, multi-robot and swarm robotics, manipulation, education
  & Cross-platform; extensive sensor and actuator models; GUI scenario/world design; ROS integration; physics-based simulation
  & AgiBot World~\citep{bu2025agibot}
\\
Unity ML-Agents~\citep{juliani2018unity} \textbf{[\href{https://unity.com/ml-agents}{link}]}
  & RGB, depth, raycasts, physics states
  & Reinforcement \& imitation learning, interactive tasks
  & Unity engine visual fidelity; Python/C\# APIs; curriculum learning
  & Used in custom RL and navigation datasets (e.g., Obstacle Tower, RoomNav, MiniWorld); 
\\

MuJoCo~\citep{todorov2012mujoco} \textbf{[\href{https://mujoco.org}{link}]}
  & Joint positions, contact forces, kinematics, RGB 
  & Continuous control, dynamics learning, RL research
  & High-speed simulation; accurate contact and soft body modeling; analytic gradients
  & Meta-World~\citep{yu2020meta}, RoboSuite~\citep{zhu2020robosuite}, custom RL benchmarks
\\

iGibson~\citep{xia2020interactive} \textbf{[\href{https://svl.stanford.edu/igibson}{link}]}
  & RGB, depth, semantic \& instance masks, object poses, contact forces
  & Interactive navigation, manipulation, semantic reasoning
  & Photorealistic dynamic scenes; real-world scene reconstructions; interactive objects and agents
  & iGibson v1/v2~\citep{xia2020interactive}
\\

UniSim~\citep{yang2023learning} \textbf{[\href{https://universal-simulator.github.io/unisim/}{link}]}
      & RGB, depth, proprioception, haptics, audio
      & Multi-modal dataset generation, multi-agent coordination, manipulation, navigation
      & Unified multi-sensor API; scalable cloud-native simulation; plugin-based extensibility; support for real and simulated sensor data
      & UniSim-VLA~\citep{yang2023learning}
    \\

SAPIEN~\citep{xiang2020sapien} \textbf{[\href{https://sapien.ucsd.edu}{link}]}
      & RGB, depth, segmentation masks, contact forces, articulated object states
      & Deformable and articulated object manipulation, semantic reasoning, dexterous grasping
      & High-fidelity GPU-based physics; real-time dynamic simulation; Python API; large-scale articulated object library
      & DexGraspNet~\citep{wang2023dexgraspnet}, TLA~\citep{hao2025tla}
    \\

    \hline
  \end{tabular}
\end{table*}
}

\section{Simulation Tools}\label{sec:simtools}

Simulation environments have become essential for VLA research, offering scalable, repeatable, and extensively annotated data at orders of magnitude greater than what is feasible in the physical world. Modern platforms such as AI2-THOR, Habitat, and NVIDIA Isaac Sim provide high-precision physics, realistic rendering, and customizable multimodal sensors ranging from RGBD cameras, force/torque and tactile probes, to proprioceptive encoders and language interfaces all configurable at fine temporal resolutions. Using procedural scene generation, randomized object properties, and scripted agent behaviors, simulators enable the automated synthesis of hundreds of thousands of trajectories, complete with ground truth annotations for object poses, semantic maps, action sequences, and natural language instructions. Crucially, built-in toolkits for scenario scripting and domain randomization facilitate systematic studies of generalization under varied lighting, object geometries, and task orders, while lightweight GPU accelerated backends support rapid iteration of new dataset designs. Together, this ecosystem of VLA simulators accelerates the co-development of control algorithms and benchmark datasets, ensuring that advances in multimodal perception, language grounding, and closed-loop planning can be evaluated and refined in a controlled, reproducible framework before deployment on real robotic platforms.

Table~\ref{tab:sim_tools} summarizes the current state-of-the-art simulation platforms used for the generation of VLA datasets. The table lists four essential aspects for each simulator: the \textit{Modalities} of sensors it offers, the main \textit{Use Cases} it supports, its fundamental technical \textit{Capabilities}, and the representative \textit{Datasets} that are based on it. This unified view allows researchers to directly compare engines based on their multi-modal sensor suite, physics accuracy, scalability, and integration with language and control APIs.

In the first column, platforms such as AI2-THOR and Habitat provide photorealistic RGB, depth, and semantic streams, making them ideal for embodied navigation and visual question answering benchmarks (e.g., ALFRED, R2R, EmbodiedQA, CVDN). Middle entries like NVIDIA Isaac Sim and Gazebo deliver advanced LiDAR, IMU, force/torque, and multi-robot support crucial for large-scale reinforcement learning, sim-to-real transfer, and multi-agent coordination, as in Open X-Embodiment and RoboSpatial. Contact-rich simulators including PyBullet, CoppeliaSim, MuJoCo, and SAPIEN enable precise force, torque, and haptic feedback, powering dexterous manipulation datasets such as DexGraspNet, CALVIN, and TLA. Emerging platforms (Unity ML-Agents, RoboSuite, IsaacGym, UniSim) highlight capabilities such as GPU-parallel rollout, cloud-native simulation, and unified multi-sensor APIs, enabling the creation of next-generation VLA datasets with millions of diverse trajectories spanning vision, language, touch, and audio.
The table provides an essential reference by mapping these four aspects across fifteen simulators: it assists in choosing the optimal backend for dataset generation, clarifies the trade-offs between rendering quality and processing speed, and identifies gaps where enhancements in simulator features could facilitate more detailed VLA benchmarks.
\rowcolors{2}{gray!20}{white}      
\begin{table*}[h]
\centering
\caption{Categorization of VLA Models by Application.}
\label{tab:vla_by_application}
\scriptsize
\begin{tabular}{p{4.5cm}|p{11.5cm}}
\toprule
\textbf{Application} & \textbf{VLA Models} \\
\midrule
\textbf{Manipulation and Task Generalization} &
CLIPort; RT-1; Gato; VIMA; RoboAgent; RT-Trajectory; RT-2; Pi-0; OpenVLA; OpenVLA-OFT; Tiny VLA; DexVLA; OneTwoVLA; RDT-1B; Shake-VLA; DexGraspVLA; Bi-VLA; MoLe-VLA; TLA; ACT; GeoManip; LMM Planner Integration; Online RL VLA; SAM2Act; VoxPoser; Diffusion Policy; Octo; Chain-of-Affordance; CogACT; ECoT; OTTER; Gemini Robotics; V-JEPA 2; Knowledge Insulating VLA Models; AgiBot World Colosseo; Fine-Tuning VLA Models; Hi Robot; EnerVerse; FLaRe; Beyond Sight; Universal Actions; RoboHorizon; VLA-Cache; Forethought VLA; GRAPE; HAMSTER; TempoRep VLA; ConRFT; RoboBERT; Diffusion Transformer Policy; GEVRM; SoFar; ARM4R; Magma; An Atomic Skill Library; VLAS; SafeVLA; iRe-VLA; TraceVLA; RevLA; RoboMamba; A3VLM; SVLR; 3D-VLA; HybridVLA; SpatialVLA; OE-VLA; EF-VLA; RoboBrain; PerAct; SayCan; CLIP-RT; VLATest; HiRT; A3VLM; SVLR; Bi-VLA; QUAR-VLA; 3D-VLA; RoboMM; Shake-VLA; MORE; DexGraspVLA; DexVLA; Humanoid-VLA; ObjectVLA; Gemini Robotics; ECoT; OTTER; $\pi$-0.5; OneTwoVLA; Helix; OE-VLA; SmolVLA; EF-VLA; PD-VLA; LeVERB; TLA; Interleave-VLA; iRe-VLA; TraceVLA; VLATest; OpenVLA-OFT; Edge VLA; CoVLA; RoboMM; RoboBERT; Diffusion Transformer Policy; GEVRM; SoFar; ARM4R; Magma; An Atomic Skill Library; VLAS; ChatVLA; RoboBrain; SafeVLA; Diffusion-VLA
\\
\addlinespace
\textbf{Autonomous Mobility} &
NaVILA; Mobility VLA; Uni-NaVid; COVLA; OpenDrive VLA; ORION; UAV-VLA; CognitiveDrone
\\
\addlinespace
\textbf{Human Assistance and Interaction} &
RoboNurse-VLA; ObjectVLA; RoboMM; Interleave-VLA; ChatVLA
\\
\addlinespace
\textbf{Robotic Platforms} &
QUAR-VLA; MORE; LeVERB; Humanoid-VLA; GR00T N1; Helix
\\
\addlinespace
\textbf{Virtual Environments} &
ShowUI-2B; Combat VLA; JARVIS-VLA; VLATest
\\
\addlinespace
\textbf{Edge and Low-Power Deployment} &
Edge VLA; SmolVLA; Gemini Robotics On-Device; FAST (Pi-0 Fast); NORA; PD-VLA
\\
\bottomrule
\end{tabular}
\end{table*}

\section{Applications and Evaluation of VLAs}\label{sec:vlabenchmark}
\subsection{Application Domains}
The Table~\ref{tab:vla_by_application} organizes VLA models into six broad application domains. These application domains are explained below.

The \textit{Manipulation and Task Generalization} domain covers models that unify visual perception and language instructions into a single control policy for diverse object-level tasks from simple grasping and placement to complex assembly with a focus on maintaining adaptability to novel objects, configurations, and robot models with limited retraining. In \textit{Autonomous Mobility} domain,  models convert high-level language goals into safe, efficient navigation plans for wheeled, legged, or aerial platforms. They combine scene understanding (identifying landmarks, obstacles, and waypoints) with motion planning to follow verbal or written navigation instructions.

In \textit{Human Assistance and Interaction} application area, agents interpret human commands and contexts to perform collaborative tasks, handling tools, manipulating domestic objects on request, or automating GUI workflows via multi-turn dialogues, prioritizing responsiveness and safety when working alongside people.
The \textit{Robotic Platforms } category focuses on models for controllers specialized to specific hardware (quadrupeds, humanoids, custom arms), integrating platform-aware perception and action modules that respect each robot's kinematics, dynamics, and sensor capabilities.
\textit{Virtual Environments} includes purely software-based agents that automate GUIs, play video games, or act as benchmarking frameworks. This domain highlights how VLA techniques are generalized beyond physical robots in simulated or desktop settings.
\textit{Edge and Low-Power Deployment} focuses on lightweight architectures optimized for immediate inference on CPUs or embedded processors, demonstrating that successful VLA integration can function within limited computational and energy constraints.

\rowcolors{2}{gray!20}{white}
\begin{table*}[h]
\centering
\scriptsize
\caption{
Table summarizes the evaluation of ten leading VLA models selected based on manipulation capabilities and task generalization. }
\label{tab:top_vla_benchmark_consistent}
\begin{tabular}{p{1.9cm}p{3.8cm}p{1.3cm}p{1.3cm}p{1.3cm}p{5cm}}
\hline
\textbf{Model}
  & \textbf{Benchmark Datasets}
  & \textbf{Success Rate}
  & \textbf{Zero-Shot}
  & \textbf{Real-Robot}
  & \textbf{Notable Results} \\
\hline

RT-2
  & Open X-Embodiment, BridgeData V2
  & High
  & High
  & \ding{52}
  & Outperforms RT-1 and SayCan on generalization and zero-shot; strong multi-robot, internet-pretrained transfer. \\

Octo
  & RLBench, Open X-Embodiment
  & Medium
  & Medium
  & \ding{52}
  & First diffusion-based generalist trained on 4M+ trajectories across 22 robots; robust sim-to-real. \\

OpenVLA
  & Open X-Embodiment, DROID
  & Medium
  & Medium
  & \ding{52}
  & LoRA fine-tuned, open-source VLA; competitive success with minimal tuning. \\

Gato
  & Internal multi-task dataset
  & Medium
  & Medium
  & \ding{52}
  & Unified policy for vision, language, and robotics; real-robot zero-shot transfer. \\

Pi-0
  & Pi-Cross-Embodiment
  & Medium
  & Medium
  & \ding{52}
  & 200Hz+ low-latency control; generalizes to new embodiments and setups. \\

DexVLA
  & RT-X, RLBench
  & Medium
  & Medium
  & \ding{52}
  & Plug-in diffusion experts; cross-embodiment adaptation without fine-tuning. \\

CLIPort
  & Ravens pick-and-place suite
  & Medium
  & Low
  & \ding{52}
  & CLIP-based dense transport achieves state-of-the-art on tabletop tasks. \\

RoboAgent 
  & RoboSet
  & High
  & High
  & \ding{52}
  & CVAE action-chunking, semantic augmentation; strong real-world generalization. \\

VIMA
  & VIMA dataset
  & Medium
  & Medium
  & \ding{52}
  & Prompt-based multimodal transformer; compositional generalization, real-robot demos. \\

TLA
  & TLA benchmark
  & Medium
  & High
  & \ding{52}
  & First language-tactile VLA; achieves 85\%+ success on contact-rich tasks. \\

\hline
\end{tabular}
\end{table*}
\subsection{VLA Model Selection and Evaluation}\label{sec:selection}

Since manipulation and task generalization remain the dominant challenges in VLA research, we selected ten models that best exemplify high manipulation skill and broad task generalization. These models were chosen according to: (1) the breadth of task coverage, including the ability to handle unseen problems, (2) zero-shot or few-shot generalization to new embodiments and environments, (3) robust real-robot validation, (4) architectural novelty in fusion or decoding mechanisms, and (5) computational practicality regarding inference speed and resource usage. 

Our evaluation framework employs three standardized metrics that are success rate, zero-shot generalization, and real-robot validation to enable direct comparisons across VLA architectures. Table \ref{tab:top_vla_benchmark_consistent} summarizes ten representative models, listing each model's name and its primary benchmark dataset. The \textit{Success Rate} column categorizes average task completion as High ($\geq$ 90 \%), Medium (70-90 \%), or Low (< 70 \%). \textit{Zero-Shot Capability} is estimated High for $\geq$ 80 \% success on unseen tasks, Medium for 50-80 \%, and Low for < 50 \%. Finally, \textit{Real-Robot Deployment} indicates whether a model has been validated on physical hardware (Yes) or remains simulation only (No). This uniform metric framework allows for direct comparison between architectures and datasets.

RT-2 demonstrates this balance by co-finetuning on
internet-scale VQA data and robot trajectories to achieve zero-shot transfer across dozens of tasks on multiple robots. Pi-0 shows that a lightweight 3B-parameter model can operate at over 200 Hz while generalizing to new tasks and robots. CLIPort introduced dense semantic grounding using CLIP-enhanced transport maps, setting state-of-the-art results on diverse tabletop manipulation tasks. VIMA demonstrated that a single, prompt-based, multimodal policy could perform six grounding tasks, including pushing, grasping, and stacking, all within a unified model and with real-robot demos. RoboAgent advances action decomposition and semantic augmentation for real-world manipulation, achieving high success in real kitchen settings. OpenVLA offers a LoRA-tuned, open-source alternative to RT-2, matching competitive performance with lower tuning overhead and real-robot support. Octo introduced the use of diffusion-based policies at scale, training on over 4 million trajectories across 22 robot platforms and achieving robust sim-to-real transfer. DexVLA uses plug-in diffusion experts to rapidly adapt across different embodiments without task-specific fine-tuning. TLA introduces the first language-tactile model, reaching over 85\% success in challenging tasks that involve a lot of contact. Gato established a unified token-based policy that spans vision, language, and robotic control, with strong zero-shot transfer on real robots.

Together, these models illustrate two main trajectories in the VLA field. On the one hand are large generalist architectures such as RT-2, Octo, Gato, and OpenVLA which use massive transformer-based backbones and diffusion decoders trained on millions of diverse trajectories. These excel in broad zero-shot generalization and are effective in many tasks and robots. On the other hand, modular and task-specialized systems, such as DexVLA, CLIPort, TLA, and RoboAgent, use targeted modules (e.g., object-centric ViTs, tactile encoders, LoRA adapters, and semantic augmentation) to enhance robustness and data efficiency for specific manipulation skills. This division demonstrates that while scaling and pretraining bring broad generalization, specialized pipelines remain critical for bridging sim-to-real gaps and achieving high-precision real-world performance.

\section{Challenges and Future Directions}\label{sec:chalanges}
This section presents open challenges and future directions essential to advancing VLA models. We categorize these into three interconnected domains: Architectural Challenges, Dataset Challenges, and Simulation Challenges. Addressing these challenges will be essential to develop robust and generalizable robotic autonomy.
\subsection{Architectural Challenges}
VLA models rely on a unified Transformer backbone to process high-resolution images or video frames alongside natural-language instructions and output platform-specific action commands. This end-to-end approach exposes several core architectural challenges that arise from the heterogeneity, scale, and physical diversity inherent to robotic control.

\paragraph{1. Tokenization and Vocabulary Alignment:}  
VLA models must process heterogeneous inputs including natural language, image patches, and continuous robot states, however standard techniques such as byte-pair encoding (BPE) for text and fixed patch embeddings for vision often fail to capture the complexities of visual and proprioceptive signals. This misalignment results in inconsistent token distributions and degraded cross-modal attention. To address this, recent approaches have introduced unified tokenization schemes. Perceiver IO uses shared latent arrays for multimodal fusion \citep{jaegle2021perceiver}, BLIP-2 introduces a Q-former to dynamically select vision tokens compatible with language models \citep{li2023blip2}, and adapter-based quantization layers allow flexible discretization within each modality stream \citep{pfeiffer2020adapterfusion}. Despite these advances, several key challenges remain, such as efficiently encoding high-dimensional sensor streams without information loss, dynamically adapting vocabularies in the presence of noise or novel configurations, achieving low-latency token generation on resource-constrained platforms, and designing interpretable token spaces to support transparent and reliable cross-modal reasoning.

\paragraph{2. Modality Fusion:}
Simply concatenating visual and linguistic features or applying basic cross-attention often fails to align the distinct statistical properties of pixel-level and word-level representations, resulting in weak visual grounding. Recent advances adopt an "align-then-fuse" paradigm to strengthen cross-modal representations. For instance,  align-before-fusing employs momentum-based contrastive learning to pre-align vision and language modalities \citep{li2022albef}, and VLMo introduces multimodal expert layers within Transformer blocks to adaptively balance contributions from each stream \citep{wang2022vlmo}. Despite these gains, key challenges remain: effectively fusing asynchronous sensory streams like haptics or audio; incorporating additional modalities such as force/torque signals; dynamically reweighting modality importance under domain shift (e.g., lighting changes or ambiguous language); improving interpretability of cross-attention layers for debugging; and enabling low-latency, resource efficient fusion for deployment on embedded robotic platforms.

\paragraph{3. Generalization Across Embodiments:}  
Fixed action vocabularies and rigid kinematic bindings severely limit the ability of VLA models to transfer across different robot models. Recent approaches address this by conditioning action generation on robot-specific descriptors or learned affordance models. For example, PaLM-E encodes explicit hardware embeddings to adapt vision-language reasoning to new platforms \citep{driess2023palm}, while RT-2 freezes its vision-language planning module and delegates embodiment/model-specific control to a lightweight action adapter. More recent efforts, such as DexVLA, go further by enabling plug-and-play cross-embodiment adaptation using diffusion-based expert modules trained across diverse kinematic structures. Despite these advances, zero-shot generalization to entirely novel robot models, payload distributions, or joint limits continues to degrade without fine-tuning. Moreover, sim-to-real transfer remains unstable under noisy sensor readings and unexpected dynamics, and generating smooth, compliant trajectories that adapt to varying torque, speed, and stiffness profiles across platforms remains an open and critical challenge.

\paragraph{4. Manipulator Motion Smoothness:}
Although many VLA models emphasize the prediction of discrete action tokens, they often neglect the quality of continuous motion trajectories, which are essential for smooth, safe and precise manipulation. Recent approaches such as Diffusion Policy \citep{chi2023diffusionpolicy} reformulate visuomotor control as a conditional denoising process, enabling the generation of temporally coherent action sequences. Based on this, the diffusion transformer policy \citep{hou2024diffusiontra} integrates large transformer architectures directly into the diffusion framework, achieving improved stability and generalization across diverse robotic platforms. However, several challenges remain unresolved: achieving real-time inference with latency-sensitive diffusion models, ensuring robust collision avoidance under sensor noise and dynamic uncertainty, maintaining a balance between trajectory smoothness and fast reactivity to changing goals, coupling diffusion-based controllers with high-level language planners.

\subsection{Dataset Challenges}

Comprehensive, varied, and well-organized datasets form the basis for developing VLA models. However, current data sets exhibit several significant limitations that obstruct the path toward robust, general-purpose VLA models.

\paragraph{1. Task Diversity:}
Current datasets are highly specialized, focusing on narrow, short-horizon tasks. For instance, ALFRED and CALVIN emphasize pick-and-place operations, while R2R focuses on navigation and finding pathways guided by language. However, few datasets integrate long-horizon task planning that combines spatial reasoning, navigation, and fine-grained object manipulation in open-ended, multi-scene environments. This fragmentation restrains the training of agents capable of seamlessly switching between locomotion and manipulation tasks in realistic household or industrial scenarios.

\paragraph{2. Modality Imbalance:}
Most VLA datasets primarily offer RGB images and textual annotations, often excluding critical sensor modalities such as depth maps, force/torque signals, tactile feedback, or proprioceptive data. When these streams are present, they are frequently captured at inconsistent sampling rates or resolutions. This lack of high-quality, synchronized multimodal data significantly limits the development of models that can perform robust sensor fusion despite environmental uncertainty.

\paragraph{3. Annotation Quality and Cost:}
Obtaining accurate labels such as 6-DoF object poses, frame-aligned multi-sensor data, or detailed natural language explanations is resource intensive and time-consuming, requiring either detailed manual annotation or unreliable semi-automated pipelines. Although simulated environments can provide perfect annotations at scale, domain gaps in appearance, physics, and interaction fidelity often degrade sim-to-real transfer. Meanwhile, current self-supervised and auto-labeling methods remain unreliable across diverse task domains.

\paragraph{4. Realism and Scale:}
Real-world datasets like Open X-Embodiment offer high fidelity data with authentic sensor noise and physical interactions, but are constrained by the cost and time of robot data collection, typically producing only hundreds of hours of recordings. In contrast, simulation platforms can generate millions of trajectories efficiently but struggle to replicate complex real-world dynamics, such as material deformation, lighting variability, or occlusion effects. This trade-off between realism and scalability remains a fundamental bottleneck in the development of models that generalize beyond laboratory conditions.

Addressing these limitations will require coordinated efforts to build long-horizon, cross-domain benchmarks; gather richly synchronized multimodal datasets; reduce annotation costs through self-supervision and automation; and bridge the realism-scale divide via hybrid simulation-real data pipelines. These advances are essential to equip future VLA models with the robustness and adaptability needed for deployment in real-world environments.

\subsection{Simulation Challenges}
Simulators provide scalable, controllable environments for generating training data for VLA models. However, several critical limitations must be addressed to ensure that simulated performance reliably transfers to real-world deployment.

\paragraph{1. Physics Accuracy and Contact Modeling:}
Popular physics engines such as MuJoCo, PyBullet, and NVIDIA Isaac Sim simplify physical interactions by relying on basic Coulomb friction models and point-contact approximations. Although this enables stable and fast simulation, it fails to capture essential dynamics such as soft-body deformation, variable surface friction, and joint compliance. As a result, policies trained in simulation often perform poorly in the real world, leading to issues like object slip, unexpected torque spikes, or unstable contact behavior.

\paragraph{2. Visual Realism and Throughput Trade-offs:}
High-fidelity simulation platforms such as AI2-THOR, Habitat, and Unity ML-Agents provide photorealistic rendering and diverse assets, making them ideal for vision-heavy tasks. However, this comes at the cost of low frame rates and high GPU demand, limiting their suitability for large-scale reinforcement learning or self-supervised pretraining. In contrast, lightweight renderers support high-throughput simulation but suffer from domain gaps in texture, lighting, and occlusion realism, reducing the effectiveness of domain-randomized policies during real-world deployment.

\paragraph{3. Lack of Built-in Language Grounding APIs:}
Most simulators do not provide native support for grounding natural language commands into agent behaviors. This forces to create custom annotation pipelines such as those used in ALFRED or TEACh that align textual instructions with actions and scene representations. These efforts introduce significant development overhead, restrict reproducibility, and lead to fragmented and non-standardized data formats.

\paragraph{4. Multi-Robot and Agent Support Capabilities:}
Support for multiple robots varies widely across simulators. Some platforms like Isaac Sim and Gazebo offer flexible import of arbitrary robot descriptions via URDF or SDF formats, facilitating multi-robot coordination and benchmarking. Others, like Webots and RoboSuite, are optimized for specific robot families, limiting generalization and reusability. This inconsistency complicates cross-platform pretraining and impairs reproducibility across hardware setups.

Overcoming these challenges requires advancing contact-rich physics modeling, optimizing rendering pipelines for both fidelity and throughput, developing standardized language grounding interfaces, and unifying multi-agent simulation support. These improvements are essential to create simulation platforms that can produce realistic, scalable, and transferable datasets for training generalizable VLA models.
\subsection{Future Directions}
To advance the next generation of VLA models, future systems should incorporate learnable, modality-aware tokenizers-such as vector-quantized VAEs or neural dictionaries to jointly discretize continuous sensor streams like proprioception and force/torque alongside visual and textual inputs. Dynamic fusion blocks (e.g., gating networks, mixture-of-experts, or conditional attention) can reweight each modality based on task demands, improving flexibility and robustness. For scaling long video or text sequences, hierarchical architectures are recommended, where lightweight CNN or RNN frontends downsample high-frame-rate inputs before passing them to sparse Transformer layers for efficient long-range modeling. Additionally, integrating diffusion-based trajectory generators with differentiable safety and collision-avoidance filters can produce smooth, compliant motions that align tightly with high-level task planning.

On the dataset side, procedural task grammars embedded in simulators can automatically generate long-horizon, open-ended scenarios that interleave navigation and fine-grained manipulation. To support sensor fusion, standardized multimodal capture pipelines should be adopted to synchronize RGB-D, tactile, force/torque, audio, and language streams at compatible sampling rates, with missing modalities augmented through cross-modal synthesis (e.g., monocular depth estimation). Annotation burdens can be reduced through self-supervised or weakly supervised techniques, including unsupervised segmentation, vision-language co-training, and active learning, to automatically extract object masks, 6-DoF trajectories, and language explanations. Hybrid synthetic-real pipelines, using neural rendering and physics-aware domain randomization, can bridge the realism-scale gap, ensuring that large-scale simulated data generalize better to physical environments.

In simulation platforms, physics fidelity should be improved through differentiable, multi-scale contact models that blend classical solvers with data-driven calibration to better handle soft-body deformation, friction variability, and compliance. Hybrid rendering pipelines that combine high-throughput rasterization for general frames with neural or ray-traced rendering for key scenes can deliver realism without compromising speed. A simulator-agnostic language grounding API should be established to map natural language instructions directly to scene graphs and agent behaviors. Finally, to enable broad generalization, simulators must support multi-robot and multi-agent scenarios, with autoimport of URDF/SDF models and shared simulation protocols, allowing for consistent policy pretraining across heterogeneous robot platforms.
\vspace{0.9cm}
\section{Conclusion}
\label{sec:conclusion}

VLA models are transforming the operational scope of robotic agents by tightly coupling perception, language, and action through integrated learning strategies. This review has systematically organized the current VLA landscape across four key dimensions: architectural design, dataset ecosystems, simulation platforms, and evaluation methodologies. The analysis reveals a rapid evolution toward multimodal, instruction conditioned agents capable of generalizing across tasks, embodiments, and environments.
While significant progress has been made, the field continues to face critical challenges in achieving scalable pretraining, reliable transfer from simulation to physical environments, and transparent decision making in safety critical contexts. Addressing these limitations will require advancements in composable learning systems, data efficient adaptation strategies, and standardized procedures for benchmarking and deployment. The future of VLA research lies in bridging foundational models with real world applications, enabling agents that can reason, act, and adapt in complex, open ended settings. This review contributes to that trajectory by synthesizing core architectural principles, learning paradigms, and deployment infrastructures, while outlining open problems that can shape the development of the next generation of vision language grounded robotic intelligence.

\section*{Declaration}
Declaration of generative AI and AI-assisted technologies in the writing process.

During the preparation of this work the author(s) used ChatGPT in order to refine the English and to search the names of the VLA models. After using this tool/service, the author(s) reviewed and edited the content as needed and take(s) full responsibility for the content of the publication.

\bibliographystyle{cas-model2-names}
\balance
\bibliography{reference}

\begin{thebibliography}{178}
\expandafter\ifx\csname natexlab\endcsname\relax\def\natexlab#1{#1}\fi
\providecommand{\url}[1]{\texttt{#1}}
\providecommand{\href}[2]{#2}
\providecommand{\path}[1]{#1}
\providecommand{\DOIprefix}{doi:}
\providecommand{\ArXivprefix}{arXiv:}
\providecommand{\URLprefix}{URL: }
\providecommand{\Pubmedprefix}{pmid:}
\providecommand{\doi}[1]{\href{http://dx.doi.org/#1}{\path{#1}}}
\providecommand{\Pubmed}[1]{\href{pmid:#1}{\path{#1}}}
\providecommand{\bibinfo}[2]{#2}
\ifx\xfnm\relax \def\xfnm[#1]{\unskip,\space#1}\fi
\bibitem[{Abdel-Hamid et~al.(2024)Abdel-Hamid, Mahmoud et~al.}]{abdel2024image}
\bibinfo{author}{Abdel-Hamid, A.}, \bibinfo{author}{Mahmoud, K.}, et~al., \bibinfo{year}{2024}.
\newblock \bibinfo{title}{Image captioning transformers: A comprehensive review}.
\newblock \bibinfo{journal}{Artificial Intelligence Review} \DOIprefix\doi{10.1007/s10462-024-10560-w}. \bibinfo{note}{early access}.
\bibitem[{Ahn et~al.(2022)Ahn, Brohan, Brown et~al.}]{ahn2022saycan}
\bibinfo{author}{Ahn, M.}, \bibinfo{author}{Brohan, A.}, \bibinfo{author}{Brown, N.}, et~al., \bibinfo{year}{2022}.
\newblock \bibinfo{title}{Do as i can, not as i say: Grounding language in robotic affordances}.
\newblock \bibinfo{journal}{arXiv preprint arXiv:2204.01691} .
\bibitem[{Anderson et~al.(2018)Anderson, Wu, Teney, Bruce, Johnson, Sünderhauf, Reid, Gould and van~den Hengel}]{anderson2018vision}
\bibinfo{author}{Anderson, P.}, \bibinfo{author}{Wu, Q.}, \bibinfo{author}{Teney, D.}, \bibinfo{author}{Bruce, J.}, \bibinfo{author}{Johnson, M.}, \bibinfo{author}{Sünderhauf, N.}, \bibinfo{author}{Reid, I.}, \bibinfo{author}{Gould, S.}, \bibinfo{author}{van~den Hengel, A.}, \bibinfo{year}{2018}.
\newblock \bibinfo{title}{Vision-and-language navigation: Interpreting visually-grounded navigation instructions in real environments}, in: \bibinfo{booktitle}{Proceedings of the IEEE Conference on Computer Vision and Pattern Recognition (CVPR)}.
\bibitem[{Arai et~al.(2025)Arai, Miwa, Sasaki, Watanabe, Yamaguchi, Aoki and Yamamoto}]{cite:5}
\bibinfo{author}{Arai, H.}, \bibinfo{author}{Miwa, K.}, \bibinfo{author}{Sasaki, K.}, \bibinfo{author}{Watanabe, K.}, \bibinfo{author}{Yamaguchi, Y.}, \bibinfo{author}{Aoki, S.}, \bibinfo{author}{Yamamoto, I.}, \bibinfo{year}{2025}.
\newblock \bibinfo{title}{Covla: Comprehensive vision-language-action dataset for autonomous driving}.
\newblock \bibinfo{journal}{2025 IEEE/CVF Winter Conference on Applications of Computer Vision (WACV), IEEE} , \bibinfo{pages}{1933--1943}.
\bibitem[{Arai et~al.(2024)Arai, Miwa, Sasaki, Yamaguchi, Watanabe, Aoki and Yamamoto}]{covla}
\bibinfo{author}{Arai, H.}, \bibinfo{author}{Miwa, K.}, \bibinfo{author}{Sasaki, K.}, \bibinfo{author}{Yamaguchi, Y.}, \bibinfo{author}{Watanabe, K.}, \bibinfo{author}{Aoki, S.}, \bibinfo{author}{Yamamoto, I.}, \bibinfo{year}{2024}.
\newblock \bibinfo{title}{Covla: Comprehensive vision-language-action dataset for autonomous driving} \URLprefix \url{https://arxiv.org/abs/2408.10845}, \href{http://arxiv.org/abs/2408.10845}{\tt arXiv:2408.10845}.
\bibitem[{Assran et~al.(2025)Assran, Bardes, Fan, Garrido, Howes, Muckley, Rizvi, Roberts, Sinha, Zholus et~al.}]{assran2025v}
\bibinfo{author}{Assran, M.}, \bibinfo{author}{Bardes, A.}, \bibinfo{author}{Fan, D.}, \bibinfo{author}{Garrido, Q.}, \bibinfo{author}{Howes, R.}, \bibinfo{author}{Muckley, M.}, \bibinfo{author}{Rizvi, A.}, \bibinfo{author}{Roberts, C.}, \bibinfo{author}{Sinha, K.}, \bibinfo{author}{Zholus, A.}, et~al., \bibinfo{year}{2025}.
\newblock \bibinfo{title}{V-jepa 2: Self-supervised video models enable understanding, prediction and planning}.
\newblock \bibinfo{journal}{arXiv preprint arXiv:2506.09985} .
\bibitem[{authors(2025)}]{anonymous2025efvla}
\bibinfo{author}{authors, A.}, \bibinfo{year}{2025}.
\newblock \bibinfo{title}{Ef-vla: Vision-language-action models with aligned vision language features for better generalization}.
\newblock \bibinfo{journal}{Under review at ICLR 2025} \URLprefix \url{https://openreview.net/forum?id=8512}. \bibinfo{note}{preprint under double-blind review}.
\bibitem[{Bharadhwaj et~al.(2023)Bharadhwaj, Pore, Liang, Singh, Rao, Zeng and Gopalakrishnan}]{bharadhwaj2023roboagent}
\bibinfo{author}{Bharadhwaj, H.}, \bibinfo{author}{Pore, N.}, \bibinfo{author}{Liang, J.}, \bibinfo{author}{Singh, J.}, \bibinfo{author}{Rao, K.}, \bibinfo{author}{Zeng, A.}, \bibinfo{author}{Gopalakrishnan, K.}, \bibinfo{year}{2023}.
\newblock \bibinfo{title}{Roboagent: Generalist robot agent with semantic and temporal understanding}.
\newblock \bibinfo{journal}{arXiv preprint arXiv:2310.08560} \URLprefix \url{https://robopen.github.io/media/roboagent.pdf}.
\bibitem[{Bjorck et~al.(2025)Bjorck, Casta{\~n}eda, Cherniadev, Da, Ding, Fan, Fang, Fox, Hu, Huang et~al.}]{bjorck2025gr00t}
\bibinfo{author}{Bjorck, J.}, \bibinfo{author}{Casta{\~n}eda, F.}, \bibinfo{author}{Cherniadev, N.}, \bibinfo{author}{Da, X.}, \bibinfo{author}{Ding, R.}, \bibinfo{author}{Fan, L.}, \bibinfo{author}{Fang, Y.}, \bibinfo{author}{Fox, D.}, \bibinfo{author}{Hu, F.}, \bibinfo{author}{Huang, S.}, et~al., \bibinfo{year}{2025}.
\newblock \bibinfo{title}{Gr00t n1: An open foundation model for generalist humanoid robots}.
\newblock \bibinfo{journal}{arXiv preprint arXiv:2503.14734} .
\bibitem[{Black et~al.(2025)Black, Brown, Darpinian, Dhabalia, Driess, Esmail, Equi and et~al.}]{black2025pi05}
\bibinfo{author}{Black, K.}, \bibinfo{author}{Brown, N.}, \bibinfo{author}{Darpinian, J.}, \bibinfo{author}{Dhabalia, K.}, \bibinfo{author}{Driess, D.}, \bibinfo{author}{Esmail, A.}, \bibinfo{author}{Equi, M.}, \bibinfo{author}{et~al.}, \bibinfo{year}{2025}.
\newblock \bibinfo{title}{$\pi$-0.5:: A vision-language-action model with open-world generalization}.
\newblock \bibinfo{journal}{arXiv preprint arXiv:2504.16054} \URLprefix \url{https://arxiv.org/abs/2504.16054}.
\bibitem[{Black et~al.(2024a)Black, Brown, Driess, Esmail, Equi, Finn, Fusai, Groom, Hausman, Ichter et~al.}]{cite:14}
\bibinfo{author}{Black, K.}, \bibinfo{author}{Brown, N.}, \bibinfo{author}{Driess, D.}, \bibinfo{author}{Esmail, A.}, \bibinfo{author}{Equi, M.}, \bibinfo{author}{Finn, C.}, \bibinfo{author}{Fusai, N.}, \bibinfo{author}{Groom, L.}, \bibinfo{author}{Hausman, K.}, \bibinfo{author}{Ichter, B.}, et~al., \bibinfo{year}{2024}a.
\newblock \bibinfo{title}{Pi-0: A vision-language-action flow model for general robot control}.
\newblock \bibinfo{journal}{arXiv preprint arXiv:2410.24164} .
\bibitem[{Black et~al.(2024b)Black, Brown, Driess, Esmail, Equi, Finn, Fusai, Groom, Hausman, Ichter et~al.}]{black2024pi_0}
\bibinfo{author}{Black, K.}, \bibinfo{author}{Brown, N.}, \bibinfo{author}{Driess, D.}, \bibinfo{author}{Esmail, A.}, \bibinfo{author}{Equi, M.}, \bibinfo{author}{Finn, C.}, \bibinfo{author}{Fusai, N.}, \bibinfo{author}{Groom, L.}, \bibinfo{author}{Hausman, K.}, \bibinfo{author}{Ichter, B.}, et~al., \bibinfo{year}{2024}b.
\newblock \bibinfo{title}{$pi\_0$: A vision-language-action flow model for general robot control}.
\newblock \bibinfo{journal}{arXiv preprint arXiv:2410.24164} .
\bibitem[{Brandi{\v{s}}auskas et~al.(2023)Brandi{\v{s}}auskas, Žukauskas and Krizhanovsky}]{brandisauskas2023seq2code}
\bibinfo{author}{Brandi{\v{s}}auskas, M.}, \bibinfo{author}{Žukauskas, M.}, \bibinfo{author}{Krizhanovsky, A.}, \bibinfo{year}{2023}.
\newblock \bibinfo{title}{Seq2code: Encoder-decoder model for program synthesis}.
\newblock \bibinfo{journal}{Procedia Computer Science} \bibinfo{volume}{222}, \bibinfo{pages}{1441--1450}.
\newblock \DOIprefix\doi{10.1016/j.procs.2023.11.306}.
\bibitem[{Brohan et~al.(2022)Brohan, Brown, Carbajal, Chebotar, Dabis, Finn, Gopalakrishnan, Hausman, Herzog, Hsu et~al.}]{cite:18}
\bibinfo{author}{Brohan, A.}, \bibinfo{author}{Brown, N.}, \bibinfo{author}{Carbajal, J.}, \bibinfo{author}{Chebotar, Y.}, \bibinfo{author}{Dabis, J.}, \bibinfo{author}{Finn, C.}, \bibinfo{author}{Gopalakrishnan, K.}, \bibinfo{author}{Hausman, K.}, \bibinfo{author}{Herzog, A.}, \bibinfo{author}{Hsu, J.}, et~al., \bibinfo{year}{2022}.
\newblock \bibinfo{title}{Rt-1: Robotics transformer for real-world control at scale}.
\newblock \bibinfo{journal}{arXiv preprint arXiv:2212.06817} .
\bibitem[{Brown et~al.(2020)Brown, Mann, Ryder et~al.}]{brown2020language}
\bibinfo{author}{Brown, T.B.}, \bibinfo{author}{Mann, B.}, \bibinfo{author}{Ryder, N.}, et~al., \bibinfo{year}{2020}.
\newblock \bibinfo{title}{Language models are few-shot learners}.
\newblock \bibinfo{journal}{Advances in Neural Information Processing Systems} \bibinfo{volume}{33}, \bibinfo{pages}{1877--1901}.
\bibitem[{Bu et~al.(2025)Bu, Cai, Chen, Cui, Ding, Feng, Gao, He, Hu, Huang et~al.}]{bu2025agibot}
\bibinfo{author}{Bu, Q.}, \bibinfo{author}{Cai, J.}, \bibinfo{author}{Chen, L.}, \bibinfo{author}{Cui, X.}, \bibinfo{author}{Ding, Y.}, \bibinfo{author}{Feng, S.}, \bibinfo{author}{Gao, S.}, \bibinfo{author}{He, X.}, \bibinfo{author}{Hu, X.}, \bibinfo{author}{Huang, X.}, et~al., \bibinfo{year}{2025}.
\newblock \bibinfo{title}{Agibot world colosseo: A large-scale manipulation platform for scalable and intelligent embodied systems}.
\newblock \bibinfo{journal}{arXiv preprint arXiv:2503.06669} .
\bibitem[{Budzianowski et~al.(2024)Budzianowski, Maa, Freed, Mo, Xie, Tipnis and Bolte}]{cite:19}
\bibinfo{author}{Budzianowski, P.}, \bibinfo{author}{Maa, W.}, \bibinfo{author}{Freed, M.}, \bibinfo{author}{Mo, J.}, \bibinfo{author}{Xie, A.}, \bibinfo{author}{Tipnis, V.}, \bibinfo{author}{Bolte, B.}, \bibinfo{year}{2024}.
\newblock \bibinfo{title}{Edgevla: Efficient vision-language-action models}.
\newblock \bibinfo{journal}{environments} \bibinfo{volume}{20}, \bibinfo{pages}{3}.
\bibitem[{Caron et~al.(2021)Caron, Touvron, Cord, Douze, Massa, Sablayrolles and J{\'e}gou}]{caron2021dino}
\bibinfo{author}{Caron, M.}, \bibinfo{author}{Touvron, H.}, \bibinfo{author}{Cord, M.}, \bibinfo{author}{Douze, M.}, \bibinfo{author}{Massa, F.}, \bibinfo{author}{Sablayrolles, A.}, \bibinfo{author}{J{\'e}gou, H.}, \bibinfo{year}{2021}.
\newblock \bibinfo{title}{Emerging properties in self-supervised vision transformers}, in: \bibinfo{booktitle}{ICCV}.
\bibitem[{Chen et~al.(2025a)Chen, Bu, Wang, Wang, Wang, Guo, Zhao, Zhu, Song, Yang et~al.}]{cite:29}
\bibinfo{author}{Chen, P.}, \bibinfo{author}{Bu, P.}, \bibinfo{author}{Wang, Y.}, \bibinfo{author}{Wang, X.}, \bibinfo{author}{Wang, Z.}, \bibinfo{author}{Guo, J.}, \bibinfo{author}{Zhao, Y.}, \bibinfo{author}{Zhu, Q.}, \bibinfo{author}{Song, J.}, \bibinfo{author}{Yang, S.}, et~al., \bibinfo{year}{2025}a.
\newblock \bibinfo{title}{Combatvla: An efficient vision-language-action model for combat tasks in 3d action role-playing games}.
\newblock \bibinfo{journal}{arXiv preprint arXiv:2503.09527} .
\bibitem[{Chen et~al.(2024)}]{quarvla}
\bibinfo{author}{Chen, X.}, et~al., \bibinfo{year}{2024}.
\newblock \bibinfo{title}{Quar-vla: A vision-language-action model for quadruped robots}.
\newblock \bibinfo{journal}{arXiv preprint arXiv:2310.08532} .
\bibitem[{Chen et~al.(2025b)Chen, Tian, Liu, Zhou, Li and Zhao}]{chen2025conrft}
\bibinfo{author}{Chen, Y.}, \bibinfo{author}{Tian, S.}, \bibinfo{author}{Liu, S.}, \bibinfo{author}{Zhou, Y.}, \bibinfo{author}{Li, H.}, \bibinfo{author}{Zhao, D.}, \bibinfo{year}{2025}b.
\newblock \bibinfo{title}{Conrft: A reinforced fine-tuning method for vla models via consistency policy}.
\newblock \bibinfo{journal}{arXiv preprint arXiv:2502.05450} .
\bibitem[{Chen et~al.(2025c)Chen, Huo, Chen and Gao}]{chen2025robohorizon}
\bibinfo{author}{Chen, Z.}, \bibinfo{author}{Huo, J.}, \bibinfo{author}{Chen, Y.}, \bibinfo{author}{Gao, Y.}, \bibinfo{year}{2025}c.
\newblock \bibinfo{title}{Robohorizon: An llm-assisted multi-view world model for long-horizon robotic manipulation}.
\newblock \bibinfo{journal}{arXiv preprint arXiv:2501.06605} .
\bibitem[{Cheng et~al.(2024)Cheng, Ji, Yang, Gongye, Zou, Kautz, Bıyık, Yin, Liu and Wang}]{cite:32}
\bibinfo{author}{Cheng, A.}, \bibinfo{author}{Ji, Y.}, \bibinfo{author}{Yang, Z.}, \bibinfo{author}{Gongye, Z.}, \bibinfo{author}{Zou, X.}, \bibinfo{author}{Kautz, J.}, \bibinfo{author}{Bıyık, E.}, \bibinfo{author}{Yin, H.}, \bibinfo{author}{Liu, S.}, \bibinfo{author}{Wang, X.}, \bibinfo{year}{2024}.
\newblock \bibinfo{title}{Navila: Legged robot vision-language-action model for navigation}.
\newblock \bibinfo{journal}{arXiv preprint arXiv:2412.04453} .
\bibitem[{Chi et~al.(2023a)Chi, Xu, Feng, Cousineau, Du, Burchfiel, Tedrake and Song}]{cite:34}
\bibinfo{author}{Chi, C.}, \bibinfo{author}{Xu, Z.}, \bibinfo{author}{Feng, S.}, \bibinfo{author}{Cousineau, E.}, \bibinfo{author}{Du, Y.}, \bibinfo{author}{Burchfiel, B.}, \bibinfo{author}{Tedrake, R.}, \bibinfo{author}{Song, S.}, \bibinfo{year}{2023}a.
\newblock \bibinfo{title}{Diffusion policy: Visuomotor policy learning via action diffusion}.
\newblock \bibinfo{journal}{The International Journal of Robotics Research} \bibinfo{volume}{02783649241273668}.
\bibitem[{Chi et~al.(2023b)Chi, Xu, Feng, Cousineau, Du, Burchfiel, Tedrake and Song}]{chi2023diffusionpolicy}
\bibinfo{author}{Chi, C.}, \bibinfo{author}{Xu, Z.}, \bibinfo{author}{Feng, S.}, \bibinfo{author}{Cousineau, E.}, \bibinfo{author}{Du, Y.}, \bibinfo{author}{Burchfiel, B.}, \bibinfo{author}{Tedrake, R.}, \bibinfo{author}{Song, S.}, \bibinfo{year}{2023}b.
\newblock \bibinfo{title}{Diffusion policy: Visuomotor policy learning via action diffusion}.
\newblock \bibinfo{journal}{arXiv preprint arXiv:2303.04137} .
\bibitem[{Chiang et~al.(2024)Chiang, Xu, Fu, Jacob, Zhang, Lee, Yu, Schenck, Rendleman, Shah et~al.}]{cite:35}
\bibinfo{author}{Chiang, H.}, \bibinfo{author}{Xu, Z.}, \bibinfo{author}{Fu, Z.}, \bibinfo{author}{Jacob, M.}, \bibinfo{author}{Zhang, T.}, \bibinfo{author}{Lee, T.}, \bibinfo{author}{Yu, W.}, \bibinfo{author}{Schenck, C.}, \bibinfo{author}{Rendleman, D.}, \bibinfo{author}{Shah, D.}, et~al., \bibinfo{year}{2024}.
\newblock \bibinfo{title}{Mobility vla: Multimodal instruction navigation with long-context vims and topological graphs}.
\newblock \bibinfo{journal}{arXiv preprint arXiv:2407.07775} .
\bibitem[{Chowdhery et~al.(2022)Chowdhery, Narang, Devlin, Bosma, Mishra, Roberts, Barham, Chung, Sutton, Sepassi, Gehrmann, Elsen, Patrick and Mishkin}]{chowdhery2022palm}
\bibinfo{author}{Chowdhery, A.}, \bibinfo{author}{Narang, S.}, \bibinfo{author}{Devlin, J.}, \bibinfo{author}{Bosma, M.}, \bibinfo{author}{Mishra, G.}, \bibinfo{author}{Roberts, A.}, \bibinfo{author}{Barham, P.}, \bibinfo{author}{Chung, H.W.}, \bibinfo{author}{Sutton, C.}, \bibinfo{author}{Sepassi, R.}, \bibinfo{author}{Gehrmann, S.}, \bibinfo{author}{Elsen, E.}, \bibinfo{author}{Patrick, D.}, \bibinfo{author}{Mishkin, P.}, \bibinfo{year}{2022}.
\newblock \bibinfo{title}{Palm: Scaling language modeling with pathways}.
\newblock \bibinfo{journal}{arXiv preprint arXiv:2204.02311} .
\bibitem[{Collaboration et~al.(2025)Collaboration, O'Neill, Rehman, Gupta, Maddukuri, Gupta, Padalkar, Lee and et~al.}]{openx}
\bibinfo{author}{Collaboration, E.}, \bibinfo{author}{O'Neill, A.}, \bibinfo{author}{Rehman, A.}, \bibinfo{author}{Gupta, A.}, \bibinfo{author}{Maddukuri, A.}, \bibinfo{author}{Gupta, A.}, \bibinfo{author}{Padalkar, A.}, \bibinfo{author}{Lee, A.}, \bibinfo{author}{et~al., A.P.}, \bibinfo{year}{2025}.
\newblock \bibinfo{title}{Open x-embodiment: Robotic learning datasets and rt-x models}.
\newblock \URLprefix \url{https://arxiv.org/abs/2310.08864}, \href{http://arxiv.org/abs/2310.08864}{\tt arXiv:2310.08864}.
\bibitem[{Coumans and Bai(2016)}]{coumans2016pybullet}
\bibinfo{author}{Coumans, E.}, \bibinfo{author}{Bai, Y.}, \bibinfo{year}{2016}.
\newblock \bibinfo{title}{Pybullet, a python module for physics simulation for robotics, games and machine learning}.
\newblock \bibinfo{howpublished}{\url{https://pybullet.org}}.
\bibitem[{Das et~al.(2018)Das, Datta, Gkioxari, Lee, Parikh and Batra}]{Das_2018_CVPR}
\bibinfo{author}{Das, A.}, \bibinfo{author}{Datta, S.}, \bibinfo{author}{Gkioxari, G.}, \bibinfo{author}{Lee, S.}, \bibinfo{author}{Parikh, D.}, \bibinfo{author}{Batra, D.}, \bibinfo{year}{2018}.
\newblock \bibinfo{title}{Embodied question answering}, in: \bibinfo{booktitle}{Proceedings of the IEEE Conference on Computer Vision and Pattern Recognition (CVPR)}.
\bibitem[{Deng et~al.(2009)Deng, Dong, Socher, Li, Li and Fei-Fei}]{deng2009imagenet}
\bibinfo{author}{Deng, J.}, \bibinfo{author}{Dong, W.}, \bibinfo{author}{Socher, R.}, \bibinfo{author}{Li, L.J.}, \bibinfo{author}{Li, K.}, \bibinfo{author}{Fei-Fei, L.}, \bibinfo{year}{2009}.
\newblock \bibinfo{title}{Imagenet: A large-scale hierarchical image database}.
\newblock \bibinfo{journal}{2009 IEEE Conference on Computer Vision and Pattern Recognition (CVPR)} , \bibinfo{pages}{248--255}\URLprefix \url{https://doi.org/10.1109/CVPR.2009.5206848}, \DOIprefix\doi{10.1109/CVPR.2009.5206848}.
\bibitem[{Devlin et~al.(2018)Devlin, Chang, Lee and Toutanova}]{devlin2018bert}
\bibinfo{author}{Devlin, J.}, \bibinfo{author}{Chang, M.W.}, \bibinfo{author}{Lee, K.}, \bibinfo{author}{Toutanova, K.}, \bibinfo{year}{2018}.
\newblock \bibinfo{title}{Bert: Pre-training of deep bidirectional transformers for language understanding}.
\newblock \bibinfo{journal}{arXiv preprint arXiv:1810.04805} .
\bibitem[{Dey et~al.(2024)Dey, Zaech, Nikolov, Van~Gool and Paudel}]{cite:39}
\bibinfo{author}{Dey, S.}, \bibinfo{author}{Zaech, J.}, \bibinfo{author}{Nikolov, N.}, \bibinfo{author}{Van~Gool, L.}, \bibinfo{author}{Paudel, D.}, \bibinfo{year}{2024}.
\newblock \bibinfo{title}{Revla: Reverting visual domain limitation of robotic foundation models}.
\newblock \bibinfo{journal}{arXiv preprint arXiv:2409.15250} .
\bibitem[{Ding et~al.(2025)Ding, Ma, Tong, Zou, Luo, Fan, Wang, Lu, Mo, Liu et~al.}]{cite:42}
\bibinfo{author}{Ding, P.}, \bibinfo{author}{Ma, J.}, \bibinfo{author}{Tong, X.}, \bibinfo{author}{Zou, B.}, \bibinfo{author}{Luo, X.}, \bibinfo{author}{Fan, Y.}, \bibinfo{author}{Wang, T.}, \bibinfo{author}{Lu, H.}, \bibinfo{author}{Mo, P.}, \bibinfo{author}{Liu, J.}, et~al., \bibinfo{year}{2025}.
\newblock \bibinfo{title}{Humanoid-vla: Towards universal humanoid control with visual integration}.
\newblock \bibinfo{journal}{arXiv preprint arXiv:2502.14795} .
\bibitem[{Ding et~al.(2024)Ding, Zhao, Zhang, Song, Zhang, Huang, Yang and Wang}]{cite:43}
\bibinfo{author}{Ding, P.}, \bibinfo{author}{Zhao, H.}, \bibinfo{author}{Zhang, W.}, \bibinfo{author}{Song, W.}, \bibinfo{author}{Zhang, M.}, \bibinfo{author}{Huang, S.}, \bibinfo{author}{Yang, N.}, \bibinfo{author}{Wang, D.}, \bibinfo{year}{2024}.
\newblock \bibinfo{title}{Quar-vla: Vision-language-action model for quadruped robots}.
\newblock \bibinfo{journal}{European Conference on Computer Vision, Springer} , \bibinfo{pages}{352--367}.
\bibitem[{Dosovitskiy et~al.(2021)Dosovitskiy, Beyer, Kolesnikov, Weissenborn, Zhai, Unterthiner, Dehghani, Minderer, Heigold, Gelly, Uszkoreit and Houlsby}]{dosovitskiy2021an}
\bibinfo{author}{Dosovitskiy, A.}, \bibinfo{author}{Beyer, L.}, \bibinfo{author}{Kolesnikov, A.}, \bibinfo{author}{Weissenborn, D.}, \bibinfo{author}{Zhai, X.}, \bibinfo{author}{Unterthiner, T.}, \bibinfo{author}{Dehghani, M.}, \bibinfo{author}{Minderer, M.}, \bibinfo{author}{Heigold, G.}, \bibinfo{author}{Gelly, S.}, \bibinfo{author}{Uszkoreit, J.}, \bibinfo{author}{Houlsby, N.}, \bibinfo{year}{2021}.
\newblock \bibinfo{title}{An image is worth 16x16 words: Transformers for image recognition at scale}, in: \bibinfo{booktitle}{International Conference on Learning Representations (ICLR)}.
\newblock \URLprefix \url{https://arxiv.org/abs/2010.11929}.
\bibitem[{Dosovitskiy et~al.(2020)Dosovitskiy, Beyer, Kolesnikov et~al.}]{dosovitskiy2020image}
\bibinfo{author}{Dosovitskiy, A.}, \bibinfo{author}{Beyer, L.}, \bibinfo{author}{Kolesnikov, A.}, et~al., \bibinfo{year}{2020}.
\newblock \bibinfo{title}{An image is worth 16x16 words: Transformers for image recognition at scale}.
\newblock \bibinfo{journal}{arXiv preprint arXiv:2010.11929} .
\bibitem[{Driess et~al.(2023)Driess, Ruiz, Goyal, Chebotar, Irpan, Ailon, Levine and Finn}]{driess2023palm}
\bibinfo{author}{Driess, D.}, \bibinfo{author}{Ruiz, N.}, \bibinfo{author}{Goyal, K.}, \bibinfo{author}{Chebotar, Y.}, \bibinfo{author}{Irpan, A.}, \bibinfo{author}{Ailon, X.}, \bibinfo{author}{Levine, S.}, \bibinfo{author}{Finn, C.}, \bibinfo{year}{2023}.
\newblock \bibinfo{title}{Palm-e: An embodied multimodal language model}.
\newblock \bibinfo{journal}{arXiv preprint arXiv:2303.03378} .
\bibitem[{Driess et~al.(2025)Driess, Springenberg, Ichter, Yu, Li-Bell, Pertsch, Ren, Walke, Vuong, Shi et~al.}]{driess2025knowledge}
\bibinfo{author}{Driess, D.}, \bibinfo{author}{Springenberg, J.T.}, \bibinfo{author}{Ichter, B.}, \bibinfo{author}{Yu, L.}, \bibinfo{author}{Li-Bell, A.}, \bibinfo{author}{Pertsch, K.}, \bibinfo{author}{Ren, A.Z.}, \bibinfo{author}{Walke, H.}, \bibinfo{author}{Vuong, Q.}, \bibinfo{author}{Shi, L.X.}, et~al., \bibinfo{year}{2025}.
\newblock \bibinfo{title}{Knowledge insulating vision-language-action models: Train fast, run fast, generalize better}.
\newblock \bibinfo{journal}{arXiv preprint arXiv:2505.23705} .
\bibitem[{Driess et~al.(2024)}]{openvla}
\bibinfo{author}{Driess, D.}, et~al., \bibinfo{year}{2024}.
\newblock \bibinfo{title}{Openvla: Open-source vision-language-action models for robotics}.
\newblock \bibinfo{journal}{arXiv preprint arXiv:2406.09246} .
\bibitem[{Fan et~al.(2025)Fan, Jia, Sun, Wang, Wei, Gong, Zhao, Tomizuka, Yang, Yan and Ding}]{fan2025interleave}
\bibinfo{author}{Fan, C.}, \bibinfo{author}{Jia, X.}, \bibinfo{author}{Sun, Y.}, \bibinfo{author}{Wang, Y.}, \bibinfo{author}{Wei, J.}, \bibinfo{author}{Gong, Z.}, \bibinfo{author}{Zhao, X.}, \bibinfo{author}{Tomizuka, M.}, \bibinfo{author}{Yang, X.}, \bibinfo{author}{Yan, J.}, \bibinfo{author}{Ding, M.}, \bibinfo{year}{2025}.
\newblock \bibinfo{title}{Interleave-vla: Enhancing robot manipulation with interleaved image-text instructions}.
\newblock \bibinfo{journal}{arXiv preprint arXiv:2406.07000} \URLprefix \url{https://arxiv.org/abs/2406.07000}.
\bibitem[{Fang et~al.(2025)Fang, Grotz, Pumacay, Wang, Fox, Krishna and Duan}]{fang2025sam2act}
\bibinfo{author}{Fang, H.}, \bibinfo{author}{Grotz, M.}, \bibinfo{author}{Pumacay, W.}, \bibinfo{author}{Wang, Y.R.}, \bibinfo{author}{Fox, D.}, \bibinfo{author}{Krishna, R.}, \bibinfo{author}{Duan, J.}, \bibinfo{year}{2025}.
\newblock \bibinfo{title}{Sam2act: Integrating visual foundation model with a memory architecture for robotic manipulation}.
\newblock \bibinfo{journal}{arXiv preprint arXiv:2501.18564} .
\bibitem[{Fouad et~al.(2024)}]{fouad2024image}
\bibinfo{author}{Fouad, S.}, et~al., \bibinfo{year}{2024}.
\newblock \bibinfo{title}{Image captioning using deep learning: A comprehensive review and future perspectives}.
\newblock \bibinfo{journal}{Multimedia Tools and Applications} \DOIprefix\doi{10.1007/s11042-024-17234-3}.
\bibitem[{Fu et~al.(2025)Fu, Zhang, Zhao, Cui, Liang, Zhang, Zhang, Xie, Wang and Bai}]{cite:56}
\bibinfo{author}{Fu, H.}, \bibinfo{author}{Zhang, D.}, \bibinfo{author}{Zhao, Z.}, \bibinfo{author}{Cui, J.}, \bibinfo{author}{Liang, D.}, \bibinfo{author}{Zhang, C.}, \bibinfo{author}{Zhang, D.}, \bibinfo{author}{Xie, H.}, \bibinfo{author}{Wang, B.}, \bibinfo{author}{Bai, X.}, \bibinfo{year}{2025}.
\newblock \bibinfo{title}{Orion: A holistic end-to-end autonomous driving framework by vision-language instructed action generation}.
\newblock \bibinfo{journal}{arXiv preprint arXiv:2503.19755} .
\bibitem[{Gao et~al.(2022a)Gao, Gao, Gong, Lin, Thattai and Sukhatme}]{gao2022dialfred}
\bibinfo{author}{Gao, X.}, \bibinfo{author}{Gao, Q.}, \bibinfo{author}{Gong, R.}, \bibinfo{author}{Lin, K.}, \bibinfo{author}{Thattai, G.}, \bibinfo{author}{Sukhatme, G.S.}, \bibinfo{year}{2022}a.
\newblock \bibinfo{title}{{DialFRED: Dialogue-Enabled Agents for Embodied Instruction Following}}.
\newblock \bibinfo{journal}{IEEE Robotics and Automation Letters} \bibinfo{volume}{7}, \bibinfo{pages}{10049--10056}.
\newblock \DOIprefix\doi{10.1109/LRA.2022.3193254}.
\bibitem[{Gao et~al.(2022b)Gao, Gao, Gong, Lin, Thattai and Sukhatme}]{dialfred}
\bibinfo{author}{Gao, X.}, \bibinfo{author}{Gao, Q.}, \bibinfo{author}{Gong, R.}, \bibinfo{author}{Lin, K.}, \bibinfo{author}{Thattai, G.}, \bibinfo{author}{Sukhatme, G.S.}, \bibinfo{year}{2022}b.
\newblock \bibinfo{title}{Dialfred: Dialogue-enabled agents for embodied instruction following}.
\newblock \bibinfo{journal}{IEEE Robotics and Automation Letters} \bibinfo{volume}{7}, \bibinfo{pages}{10049–10056}.
\newblock \URLprefix \url{http://dx.doi.org/10.1109/LRA.2022.3193254}, \DOIprefix\doi{10.1109/lra.2022.3193254}.
\bibitem[{Gbagbe et~al.(2024)Gbagbe, Cabrera, Alabbas, Alyunes, Lykov and Tsetserukou}]{gbagbe2024bivla}
\bibinfo{author}{Gbagbe, K.F.}, \bibinfo{author}{Cabrera, M.A.}, \bibinfo{author}{Alabbas, A.}, \bibinfo{author}{Alyunes, O.}, \bibinfo{author}{Lykov, A.}, \bibinfo{author}{Tsetserukou, D.}, \bibinfo{year}{2024}.
\newblock \bibinfo{title}{Bi-vla: Vision-language-action model-based system for bimanual robotic dexterous manipulations}.
\newblock \bibinfo{journal}{arXiv preprint arXiv:2405.06039} \URLprefix \url{https://arxiv.org/abs/2405.06039}.
\bibitem[{Ghojogh et~al.(2024)Ghojogh, Ghodsi, Karray and Crowley}]{ghojogh2024attention}
\bibinfo{author}{Ghojogh, B.}, \bibinfo{author}{Ghodsi, A.}, \bibinfo{author}{Karray, F.}, \bibinfo{author}{Crowley, M.}, \bibinfo{year}{2024}.
\newblock \bibinfo{title}{Attention mechanism in machine learning: A survey}.
\newblock \bibinfo{journal}{arXiv preprint arXiv:2402.05310} \URLprefix \url{https://arxiv.org/abs/2402.05310}.
\bibitem[{Grauman et~al.(2022)Grauman, Westbury, Byrne, Chavis, Furnari and et~al.}]{ego4d}
\bibinfo{author}{Grauman, K.}, \bibinfo{author}{Westbury, A.}, \bibinfo{author}{Byrne, E.}, \bibinfo{author}{Chavis, Z.}, \bibinfo{author}{Furnari, A.}, \bibinfo{author}{et~al., R.G.}, \bibinfo{year}{2022}.
\newblock \bibinfo{title}{Ego4d: Around the world in 3,000 hours of egocentric video} \URLprefix \url{https://arxiv.org/abs/2110.07058}, \href{http://arxiv.org/abs/2110.07058}{\tt arXiv:2110.07058}.
\bibitem[{Gu et~al.(2023)Gu, Kirmani, Wohlhart, Lu, Arenas, Rao, Yu, Fu, Gopalakrishnan, Xu et~al.}]{gu2023rttrajectory}
\bibinfo{author}{Gu, J.}, \bibinfo{author}{Kirmani, S.}, \bibinfo{author}{Wohlhart, P.}, \bibinfo{author}{Lu, Y.}, \bibinfo{author}{Arenas, M.G.}, \bibinfo{author}{Rao, K.}, \bibinfo{author}{Yu, W.}, \bibinfo{author}{Fu, C.}, \bibinfo{author}{Gopalakrishnan, K.}, \bibinfo{author}{Xu, Z.}, et~al., \bibinfo{year}{2023}.
\newblock \bibinfo{title}{Robotic task generalization via hindsight trajectory sketches}.
\newblock \bibinfo{journal}{arXiv preprint arXiv:2311.01977} \URLprefix \url{https://arxiv.org/abs/2311.01977}.
\bibitem[{Guo et~al.(2025)Guo, Zhang, Chen, Ji, Wang, Hu and Chen}]{guo2025irevla}
\bibinfo{author}{Guo, Y.}, \bibinfo{author}{Zhang, J.}, \bibinfo{author}{Chen, X.}, \bibinfo{author}{Ji, X.}, \bibinfo{author}{Wang, Y.J.}, \bibinfo{author}{Hu, Y.}, \bibinfo{author}{Chen, J.}, \bibinfo{year}{2025}.
\newblock \bibinfo{title}{ire-vla: Improving vision-language-action model with online reinforcement learning}.
\newblock \bibinfo{journal}{arXiv preprint arXiv:2501.16664} \URLprefix \url{https://arxiv.org/abs/2501.16664}.
\bibitem[{Guruprasad et~al.(2024)Guruprasad, Sikka, Song, Wang and Liang}]{guruprasad2024benchmarking}
\bibinfo{author}{Guruprasad, P.}, \bibinfo{author}{Sikka, H.}, \bibinfo{author}{Song, J.}, \bibinfo{author}{Wang, Y.}, \bibinfo{author}{Liang, P.P.}, \bibinfo{year}{2024}.
\newblock \bibinfo{title}{Benchmarking vision, language, \& action models on robotic learning tasks}.
\newblock \bibinfo{journal}{arXiv preprint arXiv:2411.05821} .
\bibitem[{Han et~al.(2025)Han, Qiu, Liao, Huang, Gao, Yan and Liu}]{han2025robocerebra}
\bibinfo{author}{Han, S.}, \bibinfo{author}{Qiu, B.}, \bibinfo{author}{Liao, Y.}, \bibinfo{author}{Huang, S.}, \bibinfo{author}{Gao, C.}, \bibinfo{author}{Yan, S.}, \bibinfo{author}{Liu, S.}, \bibinfo{year}{2025}.
\newblock \bibinfo{title}{Robocerebra: A large-scale benchmark for long-horizon robotic manipulation evaluation}.
\newblock \bibinfo{journal}{arXiv preprint arXiv:2506.06677} .
\bibitem[{Hao et~al.(2025)Hao, Zhang, Li, Cao, Hao, Cui and Wang}]{hao2025tla}
\bibinfo{author}{Hao, P.}, \bibinfo{author}{Zhang, C.}, \bibinfo{author}{Li, D.}, \bibinfo{author}{Cao, X.}, \bibinfo{author}{Hao, X.}, \bibinfo{author}{Cui, S.}, \bibinfo{author}{Wang, S.}, \bibinfo{year}{2025}.
\newblock \bibinfo{title}{Tla: Tactile-language-action model for contact-rich manipulation}.
\newblock \bibinfo{journal}{arXiv preprint arXiv:2503.08548} \URLprefix \url{https://arxiv.org/abs/2503.08548}.
\bibitem[{He et~al.(2025)He, Weilbach, Wojciechowska, Zhang and Wood}]{he2025plaicraftlargescaletimealignedvisionspeechaction}
\bibinfo{author}{He, Y.}, \bibinfo{author}{Weilbach, C.D.}, \bibinfo{author}{Wojciechowska, M.E.}, \bibinfo{author}{Zhang, Y.}, \bibinfo{author}{Wood, F.}, \bibinfo{year}{2025}.
\newblock \bibinfo{title}{Plaicraft: Large-scale time-aligned vision-speech-action dataset for embodied ai}.
\newblock \URLprefix \url{https://arxiv.org/abs/2505.12707}, \href{http://arxiv.org/abs/2505.12707}{\tt arXiv:2505.12707}.
\bibitem[{Hou et~al.(2024a)Hou, Zhang, Xiong, Pu, Zhao, Tong, Qiao, Dai and Chen}]{hou2024diffusion}
\bibinfo{author}{Hou, Z.}, \bibinfo{author}{Zhang, T.}, \bibinfo{author}{Xiong, Y.}, \bibinfo{author}{Pu, H.}, \bibinfo{author}{Zhao, C.}, \bibinfo{author}{Tong, R.}, \bibinfo{author}{Qiao, Y.}, \bibinfo{author}{Dai, J.}, \bibinfo{author}{Chen, Y.}, \bibinfo{year}{2024}a.
\newblock \bibinfo{title}{Diffusion transformer policy}.
\newblock \bibinfo{journal}{arXiv preprint arXiv:2410.15959} .
\bibitem[{Hou et~al.(2024b)Hou, Zhang, Xiong, Pu, Zhao, Tong, Qiao, Dai and Chen}]{hou2024diffusiontra}
\bibinfo{author}{Hou, Z.}, \bibinfo{author}{Zhang, T.}, \bibinfo{author}{Xiong, Y.}, \bibinfo{author}{Pu, H.}, \bibinfo{author}{Zhao, C.}, \bibinfo{author}{Tong, R.}, \bibinfo{author}{Qiao, Y.}, \bibinfo{author}{Dai, J.}, \bibinfo{author}{Chen, Y.}, \bibinfo{year}{2024}b.
\newblock \bibinfo{title}{Diffusion transformer policy: Scaling diffusion transformer for generalist vision–language–action learning}.
\newblock \bibinfo{journal}{arXiv preprint arXiv:2410.15959} .
\bibitem[{Hu et~al.(2024)Hu, Hendrix, Farhadi, Kembhavi, Mart{\'\i}n-Mart{\'\i}n, Stone, Zeng and Ehsani}]{hu2024flare}
\bibinfo{author}{Hu, J.}, \bibinfo{author}{Hendrix, R.}, \bibinfo{author}{Farhadi, A.}, \bibinfo{author}{Kembhavi, A.}, \bibinfo{author}{Mart{\'\i}n-Mart{\'\i}n, R.}, \bibinfo{author}{Stone, P.}, \bibinfo{author}{Zeng, K.H.}, \bibinfo{author}{Ehsani, K.}, \bibinfo{year}{2024}.
\newblock \bibinfo{title}{Flare: Achieving masterful and adaptive robot policies with large-scale reinforcement learning fine-tuning}.
\newblock \bibinfo{journal}{arXiv preprint arXiv:2409.16578} .
\bibitem[{Huang et~al.(2025a)Huang, Liu, Fu, Wu, Mukadam, Malik, Goldberg and Abbeel}]{huang2025otter}
\bibinfo{author}{Huang, H.}, \bibinfo{author}{Liu, F.}, \bibinfo{author}{Fu, L.}, \bibinfo{author}{Wu, T.}, \bibinfo{author}{Mukadam, M.}, \bibinfo{author}{Malik, J.}, \bibinfo{author}{Goldberg, K.}, \bibinfo{author}{Abbeel, P.}, \bibinfo{year}{2025}a.
\newblock \bibinfo{title}{Otter: A vision-language-action model with text-aware visual feature extraction}.
\newblock \bibinfo{journal}{arXiv preprint arXiv:2503.03734} \URLprefix \url{https://arxiv.org/abs/2503.03734}.
\bibitem[{Huang et~al.(2024)Huang, Chang, Liu, Zhu, Dong, Boularias, Gao and Li}]{huang2024a3vlm}
\bibinfo{author}{Huang, S.}, \bibinfo{author}{Chang, H.}, \bibinfo{author}{Liu, Y.}, \bibinfo{author}{Zhu, Y.}, \bibinfo{author}{Dong, H.}, \bibinfo{author}{Boularias, A.}, \bibinfo{author}{Gao, P.}, \bibinfo{author}{Li, H.}, \bibinfo{year}{2024}.
\newblock \bibinfo{title}{A3vlm: Actionable articulation-aware vision language model}, in: \bibinfo{booktitle}{Proceedings of the 8th Conference on Robot Learning (CoRL)}.
\newblock \URLprefix \url{https://arxiv.org/abs/2405.06039}.
\bibitem[{Huang et~al.(2025b)Huang, Chen, Zhou, Chen, Jiang, Hu, Liao, Gao, Li, Yao et~al.}]{huang2025enerverse}
\bibinfo{author}{Huang, S.}, \bibinfo{author}{Chen, L.}, \bibinfo{author}{Zhou, P.}, \bibinfo{author}{Chen, S.}, \bibinfo{author}{Jiang, Z.}, \bibinfo{author}{Hu, Y.}, \bibinfo{author}{Liao, Y.}, \bibinfo{author}{Gao, P.}, \bibinfo{author}{Li, H.}, \bibinfo{author}{Yao, M.}, et~al., \bibinfo{year}{2025}b.
\newblock \bibinfo{title}{Enerverse: Envisioning embodied future space for robotics manipulation}.
\newblock \bibinfo{journal}{arXiv preprint arXiv:2501.01895} .
\bibitem[{Huang et~al.(2023)Huang, Wang, Zhang, Li, Wu and Fei-Fei}]{cite:78}
\bibinfo{author}{Huang, W.}, \bibinfo{author}{Wang, C.}, \bibinfo{author}{Zhang, R.}, \bibinfo{author}{Li, Y.}, \bibinfo{author}{Wu, J.}, \bibinfo{author}{Fei-Fei, L.}, \bibinfo{year}{2023}.
\newblock \bibinfo{title}{Voxposer: Composable 3d value maps for robotic manipulation with language models}.
\newblock \bibinfo{journal}{arXiv preprint arXiv:2307.05973} .
\bibitem[{Hung et~al.(2025)Hung, Sun, Hong, Zadeh, Li, Tan, Majumder, Poria et~al.}]{cite:79}
\bibinfo{author}{Hung, C.}, \bibinfo{author}{Sun, Q.}, \bibinfo{author}{Hong, P.}, \bibinfo{author}{Zadeh, A.}, \bibinfo{author}{Li, C.}, \bibinfo{author}{Tan, U.}, \bibinfo{author}{Majumder, N.}, \bibinfo{author}{Poria, S.}, et~al., \bibinfo{year}{2025}.
\newblock \bibinfo{title}{Nora: A small open-sourced generalist vision language action model for embodied tasks}.
\newblock \bibinfo{journal}{arXiv preprint arXiv:2504.19854} .
\bibitem[{Jaegle et~al.(2022)Jaegle, Gimeno, Brock, Zisserman, Carreira, Vinyals, Verdegaal, Pessoa and Nowozin}]{jaegle2021perceiver}
\bibinfo{author}{Jaegle, A.}, \bibinfo{author}{Gimeno, N.}, \bibinfo{author}{Brock, A.}, \bibinfo{author}{Zisserman, A.}, \bibinfo{author}{Carreira, J.}, \bibinfo{author}{Vinyals, O.}, \bibinfo{author}{Verdegaal, R.}, \bibinfo{author}{Pessoa, P.}, \bibinfo{author}{Nowozin, S.}, \bibinfo{year}{2022}.
\newblock \bibinfo{title}{Perceiver {IO}: A general architecture for structured inputs \& outputs}, in: \bibinfo{booktitle}{International Conference on Learning Representations (ICLR)}.
\bibitem[{James et~al.(2020)James, Ma, Arrojo and Davison}]{james2020rlbench}
\bibinfo{author}{James, S.}, \bibinfo{author}{Ma, Z.}, \bibinfo{author}{Arrojo, D.R.}, \bibinfo{author}{Davison, A.J.}, \bibinfo{year}{2020}.
\newblock \bibinfo{title}{Rlbench: The robot learning benchmark \& learning environment}.
\newblock \bibinfo{journal}{IEEE Robotics and Automation Letters} \bibinfo{volume}{5}, \bibinfo{pages}{3019--3026}.
\newblock \DOIprefix\doi{10.1109/LRA.2020.2972831}.
\bibitem[{Ji et~al.(2025)Ji, Tan, Shi, Hao, Zhang, Zhang, Wang, Zhao, Mu, An et~al.}]{ji2025robobrain}
\bibinfo{author}{Ji, Y.}, \bibinfo{author}{Tan, H.}, \bibinfo{author}{Shi, J.}, \bibinfo{author}{Hao, X.}, \bibinfo{author}{Zhang, Y.}, \bibinfo{author}{Zhang, H.}, \bibinfo{author}{Wang, P.}, \bibinfo{author}{Zhao, M.}, \bibinfo{author}{Mu, Y.}, \bibinfo{author}{An, P.}, et~al., \bibinfo{year}{2025}.
\newblock \bibinfo{title}{Robobrain: A unified brain model for robotic manipulation from abstract to concrete}, in: \bibinfo{booktitle}{Proceedings of the Computer Vision and Pattern Recognition Conference}, pp. \bibinfo{pages}{1724--1734}.
\bibitem[{Jiang et~al.(2025)Jiang, Li, Ren, Zhou, Wang and He}]{kaiwu}
\bibinfo{author}{Jiang, S.}, \bibinfo{author}{Li, H.}, \bibinfo{author}{Ren, R.}, \bibinfo{author}{Zhou, Y.}, \bibinfo{author}{Wang, Z.}, \bibinfo{author}{He, B.}, \bibinfo{year}{2025}.
\newblock \bibinfo{title}{Kaiwu: A multimodal manipulation dataset and framework for robot learning and human-robot interaction}.
\newblock \URLprefix \url{https://arxiv.org/abs/2503.05231}, \href{http://arxiv.org/abs/2503.05231}{\tt arXiv:2503.05231}.
\bibitem[{Jiang et~al.(2022)Jiang, Gupta, Zhang et~al.}]{jiang2022vima}
\bibinfo{author}{Jiang, Y.}, \bibinfo{author}{Gupta, A.}, \bibinfo{author}{Zhang, Z.}, et~al., \bibinfo{year}{2022}.
\newblock \bibinfo{title}{Vima: General robot manipulation with multimodal prompts}.
\newblock \bibinfo{journal}{arXiv preprint arXiv:2210.03094} .
\bibitem[{Jones et~al.(2025)Jones, Mees, Sferrazza, Stachowicz, Abbeel and Levine}]{jones2025beyond}
\bibinfo{author}{Jones, J.}, \bibinfo{author}{Mees, O.}, \bibinfo{author}{Sferrazza, C.}, \bibinfo{author}{Stachowicz, K.}, \bibinfo{author}{Abbeel, P.}, \bibinfo{author}{Levine, S.}, \bibinfo{year}{2025}.
\newblock \bibinfo{title}{Beyond sight: Finetuning generalist robot policies with heterogeneous sensors via language grounding}.
\newblock \bibinfo{journal}{arXiv preprint arXiv:2501.04693} .
\bibitem[{Juliani et~al.(2018)Juliani, Berges, Teng, Gao, Henry, Mattar and Lange}]{juliani2018unity}
\bibinfo{author}{Juliani, A.}, \bibinfo{author}{Berges, V.}, \bibinfo{author}{Teng, E.}, \bibinfo{author}{Gao, Y.}, \bibinfo{author}{Henry, H.}, \bibinfo{author}{Mattar, M.}, \bibinfo{author}{Lange, D.}, \bibinfo{year}{2018}.
\newblock \bibinfo{title}{{Unity}: A general platform for intelligent agents}, in: \bibinfo{booktitle}{Proceedings of the 1st Annual Conference on Robot Learning (CoRL)}, pp. \bibinfo{pages}{49--60}.
\bibitem[{Kang et~al.(2024)Kang, Kim, Shim, Lee and Zhang}]{kang2024clip}
\bibinfo{author}{Kang, G.C.}, \bibinfo{author}{Kim, J.}, \bibinfo{author}{Shim, K.}, \bibinfo{author}{Lee, J.K.}, \bibinfo{author}{Zhang, B.T.}, \bibinfo{year}{2024}.
\newblock \bibinfo{title}{Clip-rt: Learning language-conditioned robotic policies from natural language supervision}.
\newblock \bibinfo{journal}{arXiv preprint arXiv:2411.00508} .
\bibitem[{Kang et~al.(2025)Kang, Kim, Shim, Lee and Zhang}]{kang2025cliprt}
\bibinfo{author}{Kang, G.C.}, \bibinfo{author}{Kim, J.}, \bibinfo{author}{Shim, K.}, \bibinfo{author}{Lee, J.K.}, \bibinfo{author}{Zhang, B.T.}, \bibinfo{year}{2025}.
\newblock \bibinfo{title}{Clip-rt: Learning language-conditioned robotic policies from natural language supervision}, in: \bibinfo{booktitle}{Proceedings of Robotics: Science and Systems (RSS)}.
\bibitem[{Khan et~al.(2025)Khan, Asfaw, Iarchuk, Cabrera, Moreno, Tokmurziyev and Tsetserukou}]{cite:92}
\bibinfo{author}{Khan, M.}, \bibinfo{author}{Asfaw, S.}, \bibinfo{author}{Iarchuk, D.}, \bibinfo{author}{Cabrera, M.}, \bibinfo{author}{Moreno, L.}, \bibinfo{author}{Tokmurziyev, I.}, \bibinfo{author}{Tsetserukou, D.}, \bibinfo{year}{2025}.
\newblock \bibinfo{title}{Shake-vla: Vision-language-action model-based system for bimanual robotic manipulations and liquid mixing}.
\newblock \bibinfo{journal}{arXiv preprint arXiv:2501.06919} .
\bibitem[{Khazatsky(2024)}]{droid}
\bibinfo{author}{Khazatsky, A.e.a.}, \bibinfo{year}{2024}.
\newblock \bibinfo{title}{Droid: A large-scale in-the-wild robot manipulation dataset}, in: \bibinfo{booktitle}{Robotics: Science and Systems (RSS)}.
\newblock \URLprefix \url{https://droid-dataset.github.io/}.
\bibitem[{Kim et~al.(2025)Kim, Finn and Liang}]{cite:93}
\bibinfo{author}{Kim, M.}, \bibinfo{author}{Finn, C.}, \bibinfo{author}{Liang, P.}, \bibinfo{year}{2025}.
\newblock \bibinfo{title}{Fine-tuning vision-language-action models: Optimizing speed and success}.
\newblock \bibinfo{journal}{arXiv preprint arXiv:2502.19645} .
\bibitem[{Kim et~al.(2024)Kim, Pertsch, Karamcheti, Xiao, Balakrishna, Nair, Rafailov, Foster, Lam, Sanketi et~al.}]{cite:94}
\bibinfo{author}{Kim, M.}, \bibinfo{author}{Pertsch, K.}, \bibinfo{author}{Karamcheti, S.}, \bibinfo{author}{Xiao, T.}, \bibinfo{author}{Balakrishna, A.}, \bibinfo{author}{Nair, S.}, \bibinfo{author}{Rafailov, R.}, \bibinfo{author}{Foster, E.}, \bibinfo{author}{Lam, G.}, \bibinfo{author}{Sanketi, P.}, et~al., \bibinfo{year}{2024}.
\newblock \bibinfo{title}{Open-vla: An open-source vision-language-action model}.
\newblock \bibinfo{journal}{arXiv preprint arXiv:2406.09246} .
\bibitem[{Koenig and Howard(2004)}]{koenig2004design}
\bibinfo{author}{Koenig, N.}, \bibinfo{author}{Howard, A.}, \bibinfo{year}{2004}.
\newblock \bibinfo{title}{Design and use paradigms for gazebo, an open-source multi-robot simulator}, in: \bibinfo{booktitle}{2004 IEEE/RSJ International Conference on Intelligent Robots and Systems (IROS)}, \bibinfo{publisher}{IEEE}. pp. \bibinfo{pages}{2149--2154}.
\bibitem[{Kolve et~al.(2017)Kolve, Mottaghi, Han, Randhavane, Zheng, Li, Gupta and Farhadi}]{kolve2017ai2}
\bibinfo{author}{Kolve, E.}, \bibinfo{author}{Mottaghi, R.}, \bibinfo{author}{Han, W.}, \bibinfo{author}{Randhavane, T.}, \bibinfo{author}{Zheng, X.}, \bibinfo{author}{Li, Y.}, \bibinfo{author}{Gupta, A.}, \bibinfo{author}{Farhadi, A.}, \bibinfo{year}{2017}.
\newblock \bibinfo{title}{{AI2-THOR}: An interactive 3d environment for visual ai}, in: \bibinfo{booktitle}{Proceedings of the 1st Annual Conference on Robot Learning (CoRL)}.
\newblock \URLprefix \url{https://ai2thor.allenai.org}.
\bibitem[{Krantz et~al.(2020)Krantz, Wijmans, Mukhopadhyay, Lee, Chernova and Batra}]{krantz2020beyond}
\bibinfo{author}{Krantz, J.}, \bibinfo{author}{Wijmans, E.}, \bibinfo{author}{Mukhopadhyay, A.}, \bibinfo{author}{Lee, S.}, \bibinfo{author}{Chernova, S.}, \bibinfo{author}{Batra, D.}, \bibinfo{year}{2020}.
\newblock \bibinfo{title}{Beyond the nav-graph: Vision-and-language navigation in continuous environments}, in: \bibinfo{booktitle}{Proceedings of the European Conference on Computer Vision (ECCV)}, \bibinfo{publisher}{Springer}. pp. \bibinfo{pages}{104--120}.
\newblock \URLprefix \url{https://arxiv.org/abs/2004.07787}, \DOIprefix\doi{10.1007/978-3-030-58568-6_7}.
\bibitem[{Lam et~al.(2023)Lam, Wang, Lu, Yao and Yang}]{lam2023deep}
\bibinfo{author}{Lam, C.}, \bibinfo{author}{Wang, X.}, \bibinfo{author}{Lu, X.}, \bibinfo{author}{Yao, Y.}, \bibinfo{author}{Yang, M.H.}, \bibinfo{year}{2023}.
\newblock \bibinfo{title}{Deep learning with vision transformers: A survey}.
\newblock \bibinfo{journal}{IEEE Transactions on Pattern Analysis and Machine Intelligence} \bibinfo{volume}{45}, \bibinfo{pages}{13101--13124}.
\newblock \DOIprefix\doi{10.1109/TPAMI.2023.3241477}.
\bibitem[{Li et~al.(2025a)Li, Peng, Li, Qiao, Zheng, Sun, Qin, Li, Luan, Wu et~al.}]{li2025atomic}
\bibinfo{author}{Li, D.}, \bibinfo{author}{Peng, B.}, \bibinfo{author}{Li, C.}, \bibinfo{author}{Qiao, N.}, \bibinfo{author}{Zheng, Q.}, \bibinfo{author}{Sun, L.}, \bibinfo{author}{Qin, Y.}, \bibinfo{author}{Li, B.}, \bibinfo{author}{Luan, Y.}, \bibinfo{author}{Wu, B.}, et~al., \bibinfo{year}{2025}a.
\newblock \bibinfo{title}{An atomic skill library construction method for data-efficient embodied manipulation}.
\newblock \bibinfo{journal}{arXiv preprint arXiv:2501.15068} .
\bibitem[{Li et~al.(2022)Li, Li, Li, Huang, Zhang, Wang, Dou and Ling}]{li2022albef}
\bibinfo{author}{Li, J.}, \bibinfo{author}{Li, X.}, \bibinfo{author}{Li, X.}, \bibinfo{author}{Huang, J.}, \bibinfo{author}{Zhang, J.}, \bibinfo{author}{Wang, L.}, \bibinfo{author}{Dou, Q.}, \bibinfo{author}{Ling, H.}, \bibinfo{year}{2022}.
\newblock \bibinfo{title}{Align before fuse: Vision and language representation learning with momentum distillation}, in: \bibinfo{booktitle}{Proceedings of the IEEE/CVF Conference on Computer Vision and Pattern Recognition (CVPR)}, pp. \bibinfo{pages}{18713--18723}.
\bibitem[{Li et~al.(2023)Li, Peng, Wang, Liu and Feichtenhofer}]{li2023blip2}
\bibinfo{author}{Li, J.}, \bibinfo{author}{Peng, E.A.}, \bibinfo{author}{Wang, C.}, \bibinfo{author}{Liu, J.}, \bibinfo{author}{Feichtenhofer, C.}, \bibinfo{year}{2023}.
\newblock \bibinfo{title}{Blip-2: Bootstrapping language-image pre-training with frozen image encoders and large language models}, in: \bibinfo{booktitle}{Proceedings of the IEEE/CVF International Conference on Computer Vision (ICCV)}, pp. \bibinfo{pages}{10965--10975}.
\bibitem[{Li et~al.(2024a)Li, Zhu, Tang, Wen, Zhu, Liu, Li, Cheng, Peng and Feng}]{cite:100}
\bibinfo{author}{Li, J.}, \bibinfo{author}{Zhu, Y.}, \bibinfo{author}{Tang, Z.}, \bibinfo{author}{Wen, J.}, \bibinfo{author}{Zhu, M.}, \bibinfo{author}{Liu, X.}, \bibinfo{author}{Li, C.}, \bibinfo{author}{Cheng, R.}, \bibinfo{author}{Peng, Y.}, \bibinfo{author}{Feng, F.}, \bibinfo{year}{2024}a.
\newblock \bibinfo{title}{Improving vision-language-action models via chain-of-affordance}.
\newblock \bibinfo{journal}{arXiv preprint arXiv:2412.20451} .
\bibitem[{Li et~al.(2025b)Li, Wang, He, Ma and Liang}]{cite:101}
\bibinfo{author}{Li, M.}, \bibinfo{author}{Wang, Z.}, \bibinfo{author}{He, K.}, \bibinfo{author}{Ma, X.}, \bibinfo{author}{Liang, Y.}, \bibinfo{year}{2025}b.
\newblock \bibinfo{title}{Jarvis-vla: Post-training large-scale vision language models to play visual games with keyboards and mouse}.
\newblock \bibinfo{journal}{arXiv preprint arXiv:2503.16365} .
\bibitem[{Li et~al.(2024b)Li, Liang, Wang, Luo, Chen, Liao, Wei, Deng, Xu, Zhang et~al.}]{cite:102}
\bibinfo{author}{Li, Q.}, \bibinfo{author}{Liang, Y.}, \bibinfo{author}{Wang, Z.}, \bibinfo{author}{Luo, L.}, \bibinfo{author}{Chen, X.}, \bibinfo{author}{Liao, M.}, \bibinfo{author}{Wei, F.}, \bibinfo{author}{Deng, Y.}, \bibinfo{author}{Xu, S.}, \bibinfo{author}{Zhang, Y.}, et~al., \bibinfo{year}{2024}b.
\newblock \bibinfo{title}{Cogact: A foundational vision-language-action model for synergizing cognition and action in robotic manipulation}.
\newblock \bibinfo{journal}{arXiv preprint arXiv:2411.19650} .
\bibitem[{Li et~al.(2024c)Li, Wang, Dai, Ma, Ng, Hu and Li}]{cite:103}
\bibinfo{author}{Li, S.}, \bibinfo{author}{Wang, J.}, \bibinfo{author}{Dai, R.}, \bibinfo{author}{Ma, W.}, \bibinfo{author}{Ng, W.}, \bibinfo{author}{Hu, Y.}, \bibinfo{author}{Li, Z.}, \bibinfo{year}{2024}c.
\newblock \bibinfo{title}{Robonurse-vla: Robotic scrub nurse system based on vision-language-action model}.
\newblock \bibinfo{journal}{arXiv preprint arXiv:2409.19590} .
\bibitem[{Li et~al.(2025c)Li, Deng, Zhang, Jang, Memmel, Yu, Garrett, Ramos, Fox, Li et~al.}]{li2025hamster}
\bibinfo{author}{Li, Y.}, \bibinfo{author}{Deng, Y.}, \bibinfo{author}{Zhang, J.}, \bibinfo{author}{Jang, J.}, \bibinfo{author}{Memmel, M.}, \bibinfo{author}{Yu, R.}, \bibinfo{author}{Garrett, C.R.}, \bibinfo{author}{Ramos, F.}, \bibinfo{author}{Fox, D.}, \bibinfo{author}{Li, A.}, et~al., \bibinfo{year}{2025}c.
\newblock \bibinfo{title}{Hamster: Hierarchical action models for open-world robot manipulation}.
\newblock \bibinfo{journal}{arXiv preprint arXiv:2502.05485} .
\bibitem[{Li et~al.(2025d)Li, Yan, Macaluso, Ji, Zou and Wang}]{li2025integrating}
\bibinfo{author}{Li, Y.}, \bibinfo{author}{Yan, G.}, \bibinfo{author}{Macaluso, A.}, \bibinfo{author}{Ji, M.}, \bibinfo{author}{Zou, X.}, \bibinfo{author}{Wang, X.}, \bibinfo{year}{2025}d.
\newblock \bibinfo{title}{Integrating lmm planners and 3d skill policies for generalizable manipulation}.
\newblock \bibinfo{journal}{arXiv preprint arXiv:2501.18733} .
\bibitem[{Liang et~al.(2023)Liang, Bian, Xiao et~al.}]{liang2023robo360}
\bibinfo{author}{Liang, L.}, \bibinfo{author}{Bian, L.}, \bibinfo{author}{Xiao, C.}, et~al., \bibinfo{year}{2023}.
\newblock \bibinfo{title}{Robo360: A 3d omnispective multi-material robotic manipulation dataset}.
\newblock \bibinfo{journal}{arXiv preprint arXiv:2312.06686} .
\bibitem[{Lin et~al.(2025)Lin, Nai, Hu, You, Zhao and Gao}]{lin2025onetwo}
\bibinfo{author}{Lin, F.}, \bibinfo{author}{Nai, R.}, \bibinfo{author}{Hu, Y.}, \bibinfo{author}{You, J.}, \bibinfo{author}{Zhao, J.}, \bibinfo{author}{Gao, Y.}, \bibinfo{year}{2025}.
\newblock \bibinfo{title}{Onetwovla: A unified vision-language-action model with adaptive reasoning}.
\newblock \bibinfo{journal}{arXiv preprint arXiv:2505.11917} \URLprefix \url{https://arxiv.org/abs/2505.11917}.
\bibitem[{Lin et~al.(2024)Lin, Li, Gao, Yang, Wu, Bai, Lei, Wang and Shou}]{cite:108}
\bibinfo{author}{Lin, K.}, \bibinfo{author}{Li, L.}, \bibinfo{author}{Gao, D.}, \bibinfo{author}{Yang, Z.}, \bibinfo{author}{Wu, S.}, \bibinfo{author}{Bai, Z.}, \bibinfo{author}{Lei, W.}, \bibinfo{author}{Wang, L.}, \bibinfo{author}{Shou, M.}, \bibinfo{year}{2024}.
\newblock \bibinfo{title}{Showui: One vision-language-action model for gui visual agent}.
\newblock \bibinfo{journal}{arXiv preprint arXiv:2411.17465} .
\bibitem[{Liu et~al.(2023)Liu, Zhu, Gao, Feng, Liu, Zhu and Stone}]{liu2023libero}
\bibinfo{author}{Liu, B.}, \bibinfo{author}{Zhu, Y.}, \bibinfo{author}{Gao, C.}, \bibinfo{author}{Feng, Y.}, \bibinfo{author}{Liu, Q.}, \bibinfo{author}{Zhu, Y.}, \bibinfo{author}{Stone, P.}, \bibinfo{year}{2023}.
\newblock \bibinfo{title}{Libero: Benchmarking knowledge transfer for lifelong robot learning}.
\newblock \bibinfo{journal}{Advances in Neural Information Processing Systems} \bibinfo{volume}{36}, \bibinfo{pages}{44776--44791}.
\bibitem[{Liu et~al.(2025)Liu, Chen, An, Liu, Zhang, Gu, Li, Guo, Chen, Liu et~al.}]{cite:110}
\bibinfo{author}{Liu, J.}, \bibinfo{author}{Chen, H.}, \bibinfo{author}{An, P.}, \bibinfo{author}{Liu, Z.}, \bibinfo{author}{Zhang, R.}, \bibinfo{author}{Gu, C.}, \bibinfo{author}{Li, X.}, \bibinfo{author}{Guo, Z.}, \bibinfo{author}{Chen, S.}, \bibinfo{author}{Liu, M.}, et~al., \bibinfo{year}{2025}.
\newblock \bibinfo{title}{Hybridvla: Collaborative diffusion and autoregression in a unified vision-language-action model}.
\newblock \bibinfo{journal}{arXiv preprint arXiv:2503.10631} .
\bibitem[{Liu et~al.(2024a)Liu, Liu, Wang, An, Li, Zhou, Yang, Zhang, Guo and Zhang}]{cite:111}
\bibinfo{author}{Liu, J.}, \bibinfo{author}{Liu, M.}, \bibinfo{author}{Wang, Z.}, \bibinfo{author}{An, P.}, \bibinfo{author}{Li, X.}, \bibinfo{author}{Zhou, K.}, \bibinfo{author}{Yang, S.}, \bibinfo{author}{Zhang, R.}, \bibinfo{author}{Guo, Y.}, \bibinfo{author}{Zhang, S.}, \bibinfo{year}{2024}a.
\newblock \bibinfo{title}{Robomamba: Efficient vision-language-action model for robotic reasoning and manipulation}.
\newblock \bibinfo{journal}{Advances in Neural Information Processing Systems} \bibinfo{volume}{37}, \bibinfo{pages}{40085--40110}.
\bibitem[{Liu et~al.(2024b)Liu, Wu, Li, Tan, Chen, Wang, Xu, Su and Zhu}]{cite:112}
\bibinfo{author}{Liu, S.}, \bibinfo{author}{Wu, L.}, \bibinfo{author}{Li, B.}, \bibinfo{author}{Tan, H.}, \bibinfo{author}{Chen, H.}, \bibinfo{author}{Wang, Z.}, \bibinfo{author}{Xu, K.}, \bibinfo{author}{Su, H.}, \bibinfo{author}{Zhu, J.}, \bibinfo{year}{2024}b.
\newblock \bibinfo{title}{Rdt-1b: a diffusion foundation model for bimanual manipulation}.
\newblock \bibinfo{journal}{arXiv preprint arXiv:2410.07864} .
\bibitem[{Liu et~al.(2021)Liu, Chen, Chen, Chen and Wang}]{liu2021enct5}
\bibinfo{author}{Liu, X.}, \bibinfo{author}{Chen, W.}, \bibinfo{author}{Chen, Y.}, \bibinfo{author}{Chen, Y.S.}, \bibinfo{author}{Wang, W.Y.}, \bibinfo{year}{2021}.
\newblock \bibinfo{title}{Pre-train or prompt? exploring the encoder-decoder framework for zero-shot learning}.
\newblock \bibinfo{journal}{arXiv preprint arXiv:2104.08691} \URLprefix \url{https://arxiv.org/abs/2104.08691}.
\bibitem[{Lykov et~al.(2025)Lykov, Serpiva, Khan, Sautenkov, Myshlyaev, Tadevosyan, Yaqoot and Tsetserukou}]{lykov2025cognitivedrone}
\bibinfo{author}{Lykov, A.}, \bibinfo{author}{Serpiva, V.}, \bibinfo{author}{Khan, M.H.}, \bibinfo{author}{Sautenkov, O.}, \bibinfo{author}{Myshlyaev, A.}, \bibinfo{author}{Tadevosyan, G.}, \bibinfo{author}{Yaqoot, Y.}, \bibinfo{author}{Tsetserukou, D.}, \bibinfo{year}{2025}.
\newblock \bibinfo{title}{Cognitivedrone: A vla model and evaluation benchmark for real-time cognitive task solving and reasoning in uavs}.
\newblock \bibinfo{journal}{arXiv preprint arXiv:2503.01378} .
\bibitem[{Makoviychuk et~al.(2021)Makoviychuk, Wawrzyniak, Rathod, Allshire, Handa, M{\"u}ller, Widmaier, Leal-Taix{\'e}, Makadia and Leutenegger}]{makoviychuk2021isaac}
\bibinfo{author}{Makoviychuk, V.}, \bibinfo{author}{Wawrzyniak, L.}, \bibinfo{author}{Rathod, Y.}, \bibinfo{author}{Allshire, A.}, \bibinfo{author}{Handa, A.}, \bibinfo{author}{M{\"u}ller, J.}, \bibinfo{author}{Widmaier, F.}, \bibinfo{author}{Leal-Taix{\'e}, L.}, \bibinfo{author}{Makadia, A.}, \bibinfo{author}{Leutenegger, S.}, \bibinfo{year}{2021}.
\newblock \bibinfo{title}{Isaac gym: High performance gpu based physics simulation for robot learning}.
\newblock \bibinfo{journal}{arXiv preprint arXiv:2108.10470} \URLprefix \url{https://arxiv.org/abs/2108.10470}.
\bibitem[{Mees et~al.(2022)Mees, Hermann, Rosete-Beas and Burgard}]{calvin}
\bibinfo{author}{Mees, O.}, \bibinfo{author}{Hermann, L.}, \bibinfo{author}{Rosete-Beas, E.}, \bibinfo{author}{Burgard, W.}, \bibinfo{year}{2022}.
\newblock \bibinfo{title}{Calvin: A benchmark for language-conditioned policy learning for long-horizon robot manipulation tasks}.
\newblock \URLprefix \url{https://arxiv.org/abs/2112.03227}, \href{http://arxiv.org/abs/2112.03227}{\tt arXiv:2112.03227}.
\bibitem[{Michel(2004)}]{michel2004webots}
\bibinfo{author}{Michel, O.}, \bibinfo{year}{2004}.
\newblock \bibinfo{title}{Webots: Professional mobile robot simulation}, in: \bibinfo{booktitle}{2004 IEEE/RSJ International Conference on Intelligent Robots and Systems (IROS)}, \bibinfo{publisher}{IEEE}. pp. \bibinfo{pages}{4020--4025}.
\bibitem[{Mishra et~al.(2024)Mishra, Mishra, Singh et~al.}]{mishra2024image}
\bibinfo{author}{Mishra, A.}, \bibinfo{author}{Mishra, A.}, \bibinfo{author}{Singh, G.}, et~al., \bibinfo{year}{2024}.
\newblock \bibinfo{title}{Image transformers: A survey}.
\newblock \bibinfo{journal}{ACM Computing Surveys} \bibinfo{note}{In press}.
\bibitem[{Myers et~al.(2025)Myers, Zheng, Dragan, Fang and Levine}]{myers2025temporal}
\bibinfo{author}{Myers, V.}, \bibinfo{author}{Zheng, B.C.}, \bibinfo{author}{Dragan, A.}, \bibinfo{author}{Fang, K.}, \bibinfo{author}{Levine, S.}, \bibinfo{year}{2025}.
\newblock \bibinfo{title}{Temporal representation alignment: Successor features enable emergent compositionality in robot instruction following temporal representation alignment}.
\newblock \bibinfo{journal}{arXiv preprint arXiv:2502.05454} .
\bibitem[{Niu et~al.(2025)Niu, Sharma, Xue, Biamby, Zhang, Ji, Darrell and Herzig}]{niu2025pre}
\bibinfo{author}{Niu, D.}, \bibinfo{author}{Sharma, Y.}, \bibinfo{author}{Xue, H.}, \bibinfo{author}{Biamby, G.}, \bibinfo{author}{Zhang, J.}, \bibinfo{author}{Ji, Z.}, \bibinfo{author}{Darrell, T.}, \bibinfo{author}{Herzig, R.}, \bibinfo{year}{2025}.
\newblock \bibinfo{title}{Pre-training auto-regressive robotic models with 4d representations}.
\newblock \bibinfo{journal}{arXiv preprint arXiv:2502.13142} .
\bibitem[{{NVIDIA Corporation}()}]{nvidiaisac}
\bibinfo{author}{{NVIDIA Corporation}}, .
\newblock \bibinfo{title}{{NVIDIA Isaac Sim}}.
\newblock \bibinfo{howpublished}{\url{https://developer.nvidia.com/isaac-sim}}.
\newblock \bibinfo{note}{Accessed: 2025-05-18}.
\bibitem[{Padmakumar et~al.(2021)Padmakumar, Thomason, Shrivastava, Lange, Narayan-Chen, Gella, Piramuthu, Tur and Hakkani-Tur}]{teach}
\bibinfo{author}{Padmakumar, A.}, \bibinfo{author}{Thomason, J.}, \bibinfo{author}{Shrivastava, A.}, \bibinfo{author}{Lange, P.}, \bibinfo{author}{Narayan-Chen, A.}, \bibinfo{author}{Gella, S.}, \bibinfo{author}{Piramuthu, R.}, \bibinfo{author}{Tur, G.}, \bibinfo{author}{Hakkani-Tur, D.}, \bibinfo{year}{2021}.
\newblock \bibinfo{title}{Teach: Task-driven embodied agents that chat}.
\newblock \URLprefix \url{https://arxiv.org/abs/2110.00534}, \href{http://arxiv.org/abs/2110.00534}{\tt arXiv:2110.00534}.
\bibitem[{Parada and Team(2025)}]{parada2025geminiondevice}
\bibinfo{author}{Parada, C.}, \bibinfo{author}{Team, G.R.}, \bibinfo{year}{2025}.
\newblock \bibinfo{title}{Gemini robotics on-device brings ai to local robotic devices}.
\newblock \bibinfo{howpublished}{DeepMind Blog}.
\newblock \bibinfo{note}{Available at: https://deepmind.google/discover/blog/gemini-robotics-on-device-brings-ai-to-local-robotic-devices/}.
\bibitem[{Pertsch et~al.(2025)Pertsch, Stachowicz, Ichter, Driess, Nair, Vuong, Mees, Finn and Levine}]{cite:133}
\bibinfo{author}{Pertsch, K.}, \bibinfo{author}{Stachowicz, K.}, \bibinfo{author}{Ichter, B.}, \bibinfo{author}{Driess, D.}, \bibinfo{author}{Nair, S.}, \bibinfo{author}{Vuong, Q.}, \bibinfo{author}{Mees, O.}, \bibinfo{author}{Finn, C.}, \bibinfo{author}{Levine, S.}, \bibinfo{year}{2025}.
\newblock \bibinfo{title}{Fast: Efficient action tokenization for vision-language-action models}.
\newblock \bibinfo{journal}{arXiv preprint arXiv:2501.09747} .
\bibitem[{Pfeiffer et~al.(2020)Pfeiffer, Vulic and Gurevych}]{pfeiffer2020adapterfusion}
\bibinfo{author}{Pfeiffer, J.}, \bibinfo{author}{Vulic, I.}, \bibinfo{author}{Gurevych, I.}, \bibinfo{year}{2020}.
\newblock \bibinfo{title}{Adapterfusion: Non-destructive task composition for transfer learning}, in: \bibinfo{booktitle}{Proceedings of the 2020 Conference on Empirical Methods in Natural Language Processing (EMNLP)}, pp. \bibinfo{pages}{4875--4884}.
\bibitem[{Press and Wolf(2017)}]{press2017using}
\bibinfo{author}{Press, O.}, \bibinfo{author}{Wolf, L.}, \bibinfo{year}{2017}.
\newblock \bibinfo{title}{Using the output embedding to improve language models}, in: \bibinfo{booktitle}{Proceedings of the 15th Conference of the European Chapter of the Association for Computational Linguistics (EACL)}, pp. \bibinfo{pages}{157--163}.
\newblock \URLprefix \url{https://arxiv.org/abs/1608.05859}.
\bibitem[{Qi et~al.(2025)Qi, Zhang, Ding, Dong, Yu, Li, Xu, Li, He, Fan et~al.}]{qi2025sofar}
\bibinfo{author}{Qi, Z.}, \bibinfo{author}{Zhang, W.}, \bibinfo{author}{Ding, Y.}, \bibinfo{author}{Dong, R.}, \bibinfo{author}{Yu, X.}, \bibinfo{author}{Li, J.}, \bibinfo{author}{Xu, L.}, \bibinfo{author}{Li, B.}, \bibinfo{author}{He, X.}, \bibinfo{author}{Fan, G.}, et~al., \bibinfo{year}{2025}.
\newblock \bibinfo{title}{Sofar: Language-grounded orientation bridges spatial reasoning and object manipulation}.
\newblock \bibinfo{journal}{arXiv preprint arXiv:2502.13143} .
\bibitem[{Qu et~al.(2025)Qu, Song, Chen, Yao, Ye, Ding, Wang, Gu, Zhao, Wang et~al.}]{cite:136}
\bibinfo{author}{Qu, D.}, \bibinfo{author}{Song, H.}, \bibinfo{author}{Chen, Q.}, \bibinfo{author}{Yao, Y.}, \bibinfo{author}{Ye, X.}, \bibinfo{author}{Ding, Y.}, \bibinfo{author}{Wang, Z.}, \bibinfo{author}{Gu, J.}, \bibinfo{author}{Zhao, B.}, \bibinfo{author}{Wang, D.}, et~al., \bibinfo{year}{2025}.
\newblock \bibinfo{title}{Spatialvla: Exploring spatial representations for visual-language-action model}.
\newblock \bibinfo{journal}{arXiv preprint arXiv:2501.15830} .
\bibitem[{Radford et~al.(2021)Radford, Kim, Hallacy et~al.}]{radford2021learning}
\bibinfo{author}{Radford, A.}, \bibinfo{author}{Kim, J.W.}, \bibinfo{author}{Hallacy, C.}, et~al., \bibinfo{year}{2021}.
\newblock \bibinfo{title}{Learning transferable visual models from natural language supervision}.
\newblock \bibinfo{journal}{arXiv preprint arXiv:2103.00020} .
\bibitem[{Reed et~al.(2022)Reed, Zolna, Parisotto, Colmenarejo, Novikov, Barth-Maron, Gimenez, Sulsky, Kay, Springenberg et~al.}]{cite:141}
\bibinfo{author}{Reed, S.}, \bibinfo{author}{Zolna, K.}, \bibinfo{author}{Parisotto, E.}, \bibinfo{author}{Colmenarejo, S.}, \bibinfo{author}{Novikov, A.}, \bibinfo{author}{Barth-Maron, G.}, \bibinfo{author}{Gimenez, M.}, \bibinfo{author}{Sulsky, Y.}, \bibinfo{author}{Kay, J.}, \bibinfo{author}{Springenberg, J.}, et~al., \bibinfo{year}{2022}.
\newblock \bibinfo{title}{A generalist agent}.
\newblock \bibinfo{journal}{arXiv preprint arXiv:2205.06175} .
\bibitem[{Rohmer et~al.(2013)Rohmer, Singh and Freese}]{rohmer2013v}
\bibinfo{author}{Rohmer, E.}, \bibinfo{author}{Singh, S.P.}, \bibinfo{author}{Freese, M.}, \bibinfo{year}{2013}.
\newblock \bibinfo{title}{V-rep: A versatile and scalable robot simulation framework}, in: \bibinfo{booktitle}{2013 IEEE/RSJ international conference on intelligent robots and systems}, \bibinfo{organization}{IEEE}. pp. \bibinfo{pages}{1321--1326}.
\bibitem[{Samson et~al.(2024)Samson, Muraccioli and Kanehiro}]{samson2024svlr}
\bibinfo{author}{Samson, M.}, \bibinfo{author}{Muraccioli, B.}, \bibinfo{author}{Kanehiro, F.}, \bibinfo{year}{2024}.
\newblock \bibinfo{title}{Scalable, training-free visual language robotics: A modular multi-model framework for consumer-grade gpus}.
\newblock \bibinfo{journal}{arXiv preprint arXiv:2502.01071} \URLprefix \url{https://arxiv.org/abs/2502.01071}.
\bibitem[{Sautenkov et~al.(2025)Sautenkov, Yaqoot, Lykov, Mustafa, Tadevosyan, Akhmetkazy, Cabrera, Martynov, Karaf and Tsetserukou}]{cite:150}
\bibinfo{author}{Sautenkov, O.}, \bibinfo{author}{Yaqoot, Y.}, \bibinfo{author}{Lykov, A.}, \bibinfo{author}{Mustafa, M.}, \bibinfo{author}{Tadevosyan, G.}, \bibinfo{author}{Akhmetkazy, A.}, \bibinfo{author}{Cabrera, M.}, \bibinfo{author}{Martynov, M.}, \bibinfo{author}{Karaf, S.}, \bibinfo{author}{Tsetserukou, D.}, \bibinfo{year}{2025}.
\newblock \bibinfo{title}{Uav-vla: Vision-language-action system for large scale aerial mission generation}.
\newblock \bibinfo{journal}{arXiv preprint arXiv:2501.05014} .
\bibitem[{Savva et~al.(2019)Savva, Chang, Dosovitskiy et~al.}]{savva2019habitat}
\bibinfo{author}{Savva, M.}, \bibinfo{author}{Chang, A.X.}, \bibinfo{author}{Dosovitskiy, A.}, et~al., \bibinfo{year}{2019}.
\newblock \bibinfo{title}{Habitat: A platform for embodied ai research}.
\newblock \bibinfo{journal}{Proceedings of the IEEE/CVF International Conference on Computer Vision} , \bibinfo{pages}{9339--9347}.
\bibitem[{Shi et~al.(2025)Shi, Ichter, Equi, Ke, Pertsch, Vuong, Tanner, Walling, Wang, Fusai et~al.}]{shi2025hi}
\bibinfo{author}{Shi, L.X.}, \bibinfo{author}{Ichter, B.}, \bibinfo{author}{Equi, M.}, \bibinfo{author}{Ke, L.}, \bibinfo{author}{Pertsch, K.}, \bibinfo{author}{Vuong, Q.}, \bibinfo{author}{Tanner, J.}, \bibinfo{author}{Walling, A.}, \bibinfo{author}{Wang, H.}, \bibinfo{author}{Fusai, N.}, et~al., \bibinfo{year}{2025}.
\newblock \bibinfo{title}{Hi robot: Open-ended instruction following with hierarchical vision-language-action models}.
\newblock \bibinfo{journal}{arXiv preprint arXiv:2502.19417} .
\bibitem[{Shridhar et~al.(2022a)Shridhar, Manuelli and Fox}]{cite:157}
\bibinfo{author}{Shridhar, M.}, \bibinfo{author}{Manuelli, L.}, \bibinfo{author}{Fox, D.}, \bibinfo{year}{2022}a.
\newblock \bibinfo{title}{Cliport: What and where pathways for robotic manipulation}.
\newblock \bibinfo{journal}{Conference on robot learning, PMLR} , \bibinfo{pages}{894--906}.
\bibitem[{Shridhar et~al.(2022b)Shridhar, Manuelli and Fox}]{shridhar2022peract}
\bibinfo{author}{Shridhar, M.}, \bibinfo{author}{Manuelli, L.}, \bibinfo{author}{Fox, D.}, \bibinfo{year}{2022}b.
\newblock \bibinfo{title}{Perceiver-actor: A multi-task transformer for robotic manipulation}, in: \bibinfo{booktitle}{Conference on Robot Learning (CoRL)}.
\newblock \URLprefix \url{https://peract.github.io/paper/peract_corl2022.pdf}.
\bibitem[{Shridhar et~al.(2023)Shridhar, Manuelli and Fox}]{shridhar2023perceiver}
\bibinfo{author}{Shridhar, M.}, \bibinfo{author}{Manuelli, L.}, \bibinfo{author}{Fox, D.}, \bibinfo{year}{2023}.
\newblock \bibinfo{title}{Perceiver-actor: A multi-task transformer for robotic manipulation}, in: \bibinfo{booktitle}{Conference on Robot Learning}, \bibinfo{organization}{PMLR}. pp. \bibinfo{pages}{785--799}.
\bibitem[{Shridhar et~al.(2020)Shridhar, Thomason, Gordon, Bisk, Han, Mottaghi, Zettlemoyer and Fox}]{shridhar2020alfred}
\bibinfo{author}{Shridhar, M.}, \bibinfo{author}{Thomason, J.}, \bibinfo{author}{Gordon, D.}, \bibinfo{author}{Bisk, Y.}, \bibinfo{author}{Han, W.}, \bibinfo{author}{Mottaghi, R.}, \bibinfo{author}{Zettlemoyer, L.}, \bibinfo{author}{Fox, D.}, \bibinfo{year}{2020}.
\newblock \bibinfo{title}{{ALFRED:} a benchmark for interpreting grounded instructions for everyday tasks}, in: \bibinfo{booktitle}{Proceedings of the IEEE Conference on Computer Vision and Pattern Recognition (CVPR)}.
\newblock \URLprefix \url{https://arxiv.org/abs/1912.01734}.
\bibitem[{Shukor et~al.(2025)Shukor, Aubakirova, Capuano, Kooijmans, Palma, Zouitine and et~al.}]{shukor2025smolvla}
\bibinfo{author}{Shukor, M.}, \bibinfo{author}{Aubakirova, D.}, \bibinfo{author}{Capuano, F.}, \bibinfo{author}{Kooijmans, P.}, \bibinfo{author}{Palma, S.}, \bibinfo{author}{Zouitine, A.}, \bibinfo{author}{et~al.}, \bibinfo{year}{2025}.
\newblock \bibinfo{title}{Smolvla: A vision-language-action model for affordable and efficient robotics}.
\newblock \bibinfo{journal}{arXiv preprint arXiv:2506.01844} \URLprefix \url{https://arxiv.org/abs/2506.01844}.
\bibitem[{Sliwowski et~al.(2025)Sliwowski, Jadav, Stanovcic, Orbik, Heidersberger and Lee}]{sliwowski2025reassemble}
\bibinfo{author}{Sliwowski, D.}, \bibinfo{author}{Jadav, S.}, \bibinfo{author}{Stanovcic, S.}, \bibinfo{author}{Orbik, J.}, \bibinfo{author}{Heidersberger, J.}, \bibinfo{author}{Lee, D.}, \bibinfo{year}{2025}.
\newblock \bibinfo{title}{Reassemble: A multimodal dataset for contact-rich robotic assembly and disassembly}.
\newblock \bibinfo{journal}{arXiv preprint arXiv:2502.05086} .
\bibitem[{Song et~al.(2025a)Song, Blukis, Tremblay, Tyree, Su and Birchfield}]{robospatial}
\bibinfo{author}{Song, C.H.}, \bibinfo{author}{Blukis, V.}, \bibinfo{author}{Tremblay, J.}, \bibinfo{author}{Tyree, S.}, \bibinfo{author}{Su, Y.}, \bibinfo{author}{Birchfield, S.}, \bibinfo{year}{2025}a.
\newblock \bibinfo{title}{Robospatial: Teaching spatial understanding to 2d and 3d vision-language models for robotics}.
\newblock \URLprefix \url{https://arxiv.org/abs/2411.16537}, \href{http://arxiv.org/abs/2411.16537}{\tt arXiv:2411.16537}.
\bibitem[{Song et~al.(2025b)Song, Chen, Ding, Zhao, Zhao, Zhong, Ge, Ma and Li}]{song2025accelerating}
\bibinfo{author}{Song, W.}, \bibinfo{author}{Chen, J.}, \bibinfo{author}{Ding, P.}, \bibinfo{author}{Zhao, H.}, \bibinfo{author}{Zhao, W.}, \bibinfo{author}{Zhong, Z.}, \bibinfo{author}{Ge, Z.}, \bibinfo{author}{Ma, J.}, \bibinfo{author}{Li, H.}, \bibinfo{year}{2025}b.
\newblock \bibinfo{title}{Accelerating vision-language-action model integrated with action chunking via parallel decoding}.
\newblock \bibinfo{journal}{arXiv preprint arXiv:2503.02310} .
\bibitem[{Tang et~al.(2025)Tang, Pan, Liu, Tomizuka, Li, Fu and Ding}]{tang2025geomanip}
\bibinfo{author}{Tang, W.}, \bibinfo{author}{Pan, J.H.}, \bibinfo{author}{Liu, Y.H.}, \bibinfo{author}{Tomizuka, M.}, \bibinfo{author}{Li, L.E.}, \bibinfo{author}{Fu, C.W.}, \bibinfo{author}{Ding, M.}, \bibinfo{year}{2025}.
\newblock \bibinfo{title}{Geomanip: Geometric constraints as general interfaces for robot manipulation}.
\newblock \bibinfo{journal}{arXiv preprint arXiv:2501.09783} .
\bibitem[{Team(2025a)}]{helix2025}
\bibinfo{author}{Team, F.A.}, \bibinfo{year}{2025}a.
\newblock \bibinfo{title}{Helix: A vision-language-action model for generalist humanoid control}.
\newblock \bibinfo{note}{Available at: https://www.figure.ai/news/helix}.
\bibitem[{Team(2025b)}]{gemini2025robotics}
\bibinfo{author}{Team, G.R.}, \bibinfo{year}{2025}b.
\newblock \bibinfo{title}{Gemini robotics: Bringing ai into the physical world}.
\newblock \bibinfo{journal}{arXiv preprint arXiv:2503.20020} \URLprefix \url{https://arxiv.org/abs/2503.20020}.
\bibitem[{Team et~al.(2024)Team, Ghosh, Walke, Pertsch, Black, Mees, Dasari, Hejna, Kreiman, Xu et~al.}]{team2024octo}
\bibinfo{author}{Team, O.M.}, \bibinfo{author}{Ghosh, D.}, \bibinfo{author}{Walke, H.}, \bibinfo{author}{Pertsch, K.}, \bibinfo{author}{Black, K.}, \bibinfo{author}{Mees, O.}, \bibinfo{author}{Dasari, S.}, \bibinfo{author}{Hejna, J.}, \bibinfo{author}{Kreiman, T.}, \bibinfo{author}{Xu, C.}, et~al., \bibinfo{year}{2024}.
\newblock \bibinfo{title}{Octo: An open-source generalist robot policy}.
\newblock \bibinfo{journal}{arXiv preprint arXiv:2405.12213} .
\bibitem[{Thomason et~al.(2019)Thomason, Murray, Cakmak and Zettlemoyer}]{cvdn}
\bibinfo{author}{Thomason, J.}, \bibinfo{author}{Murray, M.}, \bibinfo{author}{Cakmak, M.}, \bibinfo{author}{Zettlemoyer, L.}, \bibinfo{year}{2019}.
\newblock \bibinfo{title}{Vision-and-dialog navigation}.
\newblock \URLprefix \url{https://arxiv.org/abs/1907.04957}, \href{http://arxiv.org/abs/1907.04957}{\tt arXiv:1907.04957}.
\bibitem[{Todorov et~al.(2012)Todorov, Erez and Tassa}]{todorov2012mujoco}
\bibinfo{author}{Todorov, E.}, \bibinfo{author}{Erez, T.}, \bibinfo{author}{Tassa, Y.}, \bibinfo{year}{2012}.
\newblock \bibinfo{title}{Mujoco: A physics engine for model-based control}, in: \bibinfo{booktitle}{2012 IEEE/RSJ international conference on intelligent robots and systems}, \bibinfo{organization}{IEEE}. pp. \bibinfo{pages}{5026--5033}.
\bibitem[{Touvron et~al.(2023)Touvron, Martin, Stone, Albert, Almahairi, Laradji, Aqaj, Baratin, Lee, Verde, Kaplanyan, Azar, Gelly and Joulin}]{touvron2023llama}
\bibinfo{author}{Touvron, H.}, \bibinfo{author}{Martin, T.}, \bibinfo{author}{Stone, L.}, \bibinfo{author}{Albert, A.}, \bibinfo{author}{Almahairi, A.}, \bibinfo{author}{Laradji, I.}, \bibinfo{author}{Aqaj, Y.}, \bibinfo{author}{Baratin, A.}, \bibinfo{author}{Lee, S.}, \bibinfo{author}{Verde, Z.}, \bibinfo{author}{Kaplanyan, A.}, \bibinfo{author}{Azar, M.}, \bibinfo{author}{Gelly, S.}, \bibinfo{author}{Joulin, A.}, \bibinfo{year}{2023}.
\newblock \bibinfo{title}{Llama: Open and efficient foundation language models}.
\newblock \bibinfo{journal}{arXiv preprint arXiv:2302.13971} .
\bibitem[{Vaswani et~al.(2017)Vaswani, Shazeer, Parmar, Uszkoreit, Jones, Gomez, Kaiser and Polosukhin}]{vaswani2017attention}
\bibinfo{author}{Vaswani, A.}, \bibinfo{author}{Shazeer, N.}, \bibinfo{author}{Parmar, N.}, \bibinfo{author}{Uszkoreit, J.}, \bibinfo{author}{Jones, L.}, \bibinfo{author}{Gomez, A.N.}, \bibinfo{author}{Kaiser, Å.}, \bibinfo{author}{Polosukhin, I.}, \bibinfo{year}{2017}.
\newblock \bibinfo{title}{Attention is all you need}, in: \bibinfo{booktitle}{Advances in Neural Information Processing Systems (NeurIPS)}, pp. \bibinfo{pages}{5998--6008}.
\newblock \URLprefix \url{https://arxiv.org/abs/1706.03762}.
\bibitem[{Walke et~al.(2023)Walke, Black, Zhao, Vuong, Zheng, Hansen-Estruch, He, Myers, Kim, Du et~al.}]{walke2023bridgedata}
\bibinfo{author}{Walke, H.R.}, \bibinfo{author}{Black, K.}, \bibinfo{author}{Zhao, T.Z.}, \bibinfo{author}{Vuong, Q.}, \bibinfo{author}{Zheng, C.}, \bibinfo{author}{Hansen-Estruch, P.}, \bibinfo{author}{He, A.W.}, \bibinfo{author}{Myers, V.}, \bibinfo{author}{Kim, M.J.}, \bibinfo{author}{Du, M.}, et~al., \bibinfo{year}{2023}.
\newblock \bibinfo{title}{Bridgedata v2: A dataset for robot learning at scale}, in: \bibinfo{booktitle}{Conference on Robot Learning}, \bibinfo{organization}{PMLR}. pp. \bibinfo{pages}{1723--1736}.
\bibitem[{Wang et~al.(2022)Wang, Zhang, Zhao, Liu, Bian, Yu, Xu, Lau and Wang}]{wang2022vlmo}
\bibinfo{author}{Wang, L.}, \bibinfo{author}{Zhang, H.}, \bibinfo{author}{Zhao, Y.}, \bibinfo{author}{Liu, Z.}, \bibinfo{author}{Bian, J.}, \bibinfo{author}{Yu, H.}, \bibinfo{author}{Xu, C.}, \bibinfo{author}{Lau, R.}, \bibinfo{author}{Wang, S.}, \bibinfo{year}{2022}.
\newblock \bibinfo{title}{Vlmo: Unified vision-language pre-training with mixture-of-modality-experts}, in: \bibinfo{booktitle}{ACM International Conference on Multimedia (MM)}.
\bibitem[{Wang et~al.(2023)Wang, Zhang, Chen, Xu, Li, Liu and Wang}]{wang2023dexgraspnet}
\bibinfo{author}{Wang, R.}, \bibinfo{author}{Zhang, J.}, \bibinfo{author}{Chen, J.}, \bibinfo{author}{Xu, Y.}, \bibinfo{author}{Li, P.}, \bibinfo{author}{Liu, T.}, \bibinfo{author}{Wang, H.}, \bibinfo{year}{2023}.
\newblock \bibinfo{title}{Dexgraspnet: A large-scale robotic dexterous grasp dataset for general objects based on simulation}, in: \bibinfo{booktitle}{2023 IEEE International Conference on Robotics and Automation (ICRA)}, \bibinfo{organization}{IEEE}. pp. \bibinfo{pages}{11359--11366}.
\bibitem[{Wang et~al.(2025)Wang, Shan, Zhang, Gao, Han, Chen, Wei, Zhang, Wong, Zhao et~al.}]{wang2025robobert}
\bibinfo{author}{Wang, S.}, \bibinfo{author}{Shan, J.}, \bibinfo{author}{Zhang, J.}, \bibinfo{author}{Gao, H.}, \bibinfo{author}{Han, H.}, \bibinfo{author}{Chen, Y.}, \bibinfo{author}{Wei, K.}, \bibinfo{author}{Zhang, C.}, \bibinfo{author}{Wong, K.}, \bibinfo{author}{Zhao, J.}, et~al., \bibinfo{year}{2025}.
\newblock \bibinfo{title}{Robobert: An end-to-end multimodal robotic manipulation model}.
\newblock \bibinfo{journal}{arXiv preprint arXiv:2502.07837} .
\bibitem[{Wang et~al.(2024a)Wang, Han, Liang, Yang, Liu, Zhang, Wang, Luo and Tang}]{wang2024adversarial}
\bibinfo{author}{Wang, T.}, \bibinfo{author}{Han, C.}, \bibinfo{author}{Liang, J.C.}, \bibinfo{author}{Yang, W.}, \bibinfo{author}{Liu, D.}, \bibinfo{author}{Zhang, L.X.}, \bibinfo{author}{Wang, Q.}, \bibinfo{author}{Luo, J.}, \bibinfo{author}{Tang, R.}, \bibinfo{year}{2024}a.
\newblock \bibinfo{title}{Exploring the adversarial vulnerabilities of vision-language-action models in robotics}.
\newblock \bibinfo{journal}{arXiv preprint arXiv:2411.13587} .
\bibitem[{Wang et~al.(2024b)Wang, Zheng, Nie, Xu, Wang, Ye, Li, Zhang, Cheng, Dong, Cai, Lin, Zheng and Liang}]{zhou2024ario}
\bibinfo{author}{Wang, Z.}, \bibinfo{author}{Zheng, H.}, \bibinfo{author}{Nie, Y.}, \bibinfo{author}{Xu, W.}, \bibinfo{author}{Wang, Q.}, \bibinfo{author}{Ye, H.}, \bibinfo{author}{Li, Z.}, \bibinfo{author}{Zhang, K.}, \bibinfo{author}{Cheng, X.}, \bibinfo{author}{Dong, W.}, \bibinfo{author}{Cai, C.}, \bibinfo{author}{Lin, L.}, \bibinfo{author}{Zheng, F.}, \bibinfo{author}{Liang, X.}, \bibinfo{year}{2024}b.
\newblock \bibinfo{title}{All robots in one: A new standard and unified dataset for versatile, general‑purpose embodied agents}.
\newblock \bibinfo{journal}{arXiv preprint arXiv:2408.10899} \URLprefix \url{https://arxiv.org/abs/2408.10899}.
\bibitem[{Wang et~al.(2024c)Wang, Zhou, Song, Huang, Shu and Ma}]{cite:182}
\bibinfo{author}{Wang, Z.}, \bibinfo{author}{Zhou, Z.}, \bibinfo{author}{Song, J.}, \bibinfo{author}{Huang, Y.}, \bibinfo{author}{Shu, Z.}, \bibinfo{author}{Ma, L.}, \bibinfo{year}{2024}c.
\newblock \bibinfo{title}{Towards testing and evaluating vision-language-action models for robotic manipulation: An empirical study}.
\newblock \bibinfo{journal}{arXiv preprint arXiv:2409.12894} .
\bibitem[{Wang et~al.(2024d)Wang, Zhou et~al.}]{ladev2024}
\bibinfo{author}{Wang, Z.}, \bibinfo{author}{Zhou, Z.}, et~al., \bibinfo{year}{2024}d.
\newblock \bibinfo{title}{Ladev: A language-driven testing and evaluation platform for vision-language-action models in robotic manipulation}.
\newblock \bibinfo{journal}{arXiv preprint arXiv:2410.05191} .
\bibitem[{Wen et~al.(2024)Wen, Zhu, Zhu, Tang, Li, Zhou, Li, Liu, Peng, Shen and Feng}]{wen2024diffusionvla}
\bibinfo{author}{Wen, J.}, \bibinfo{author}{Zhu, M.}, \bibinfo{author}{Zhu, Y.}, \bibinfo{author}{Tang, Z.}, \bibinfo{author}{Li, J.}, \bibinfo{author}{Zhou, Z.}, \bibinfo{author}{Li, C.}, \bibinfo{author}{Liu, X.}, \bibinfo{author}{Peng, Y.}, \bibinfo{author}{Shen, C.}, \bibinfo{author}{Feng, F.}, \bibinfo{year}{2024}.
\newblock \bibinfo{title}{Diffusion-vla: Generalizable and interpretable robot foundation model via self-generated reasoning}.
\newblock \bibinfo{journal}{arXiv preprint arXiv:2412.03293} \bibinfo{note}{Accepted by ICML 2025}.
\bibitem[{Wen et~al.(2025)Wen, Zhu, Li, Tang, Shen and Feng}]{cite:185}
\bibinfo{author}{Wen, J.}, \bibinfo{author}{Zhu, Y.}, \bibinfo{author}{Li, J.}, \bibinfo{author}{Tang, Z.}, \bibinfo{author}{Shen, C.}, \bibinfo{author}{Feng, F.}, \bibinfo{year}{2025}.
\newblock \bibinfo{title}{Dexvla: Vision-language model with plug-in diffusion expert for general robot control}.
\newblock \bibinfo{journal}{arXiv preprint arXiv:2502.05855} .
\bibitem[{Wu et~al.(2025)Wu, Tian, Swamy and Bajcsy}]{wu2025foresight}
\bibinfo{author}{Wu, Y.}, \bibinfo{author}{Tian, R.}, \bibinfo{author}{Swamy, G.}, \bibinfo{author}{Bajcsy, A.}, \bibinfo{year}{2025}.
\newblock \bibinfo{title}{From foresight to forethought: Vlm-in-the-loop policy steering via latent alignment}.
\newblock \bibinfo{journal}{arXiv preprint arXiv:2502.01828} .
\bibitem[{Xia et~al.(2020)Xia, Li, Mart{\'i}n-Mart{\'i}n, Litany, Zamir and Savarese}]{xia2020interactive}
\bibinfo{author}{Xia, F.}, \bibinfo{author}{Li, C.}, \bibinfo{author}{Mart{\'i}n-Mart{\'i}n, R.}, \bibinfo{author}{Litany, O.}, \bibinfo{author}{Zamir, A.R.}, \bibinfo{author}{Savarese, S.}, \bibinfo{year}{2020}.
\newblock \bibinfo{title}{Interactive gibson benchmark: A benchmark for interactive navigation in cluttered environments}, in: \bibinfo{booktitle}{Proceedings of the IEEE/RSJ International Conference on Intelligent Robots and Systems (IROS)}, pp. \bibinfo{pages}{5891--5898}.
\newblock \URLprefix \url{https://arxiv.org/abs/2002.10322}, \DOIprefix\doi{10.1109/IROS45743.2020.9341201}.
\bibitem[{Xiang et~al.(2020)Xiang, Qin, Mo, Xia, Zhu, Liu, Liu, Jiang, Yuan, Wang et~al.}]{xiang2020sapien}
\bibinfo{author}{Xiang, F.}, \bibinfo{author}{Qin, Y.}, \bibinfo{author}{Mo, K.}, \bibinfo{author}{Xia, Y.}, \bibinfo{author}{Zhu, H.}, \bibinfo{author}{Liu, F.}, \bibinfo{author}{Liu, M.}, \bibinfo{author}{Jiang, H.}, \bibinfo{author}{Yuan, Y.}, \bibinfo{author}{Wang, H.}, et~al., \bibinfo{year}{2020}.
\newblock \bibinfo{title}{Sapien: A simulated part-based interactive environment}, in: \bibinfo{booktitle}{Proceedings of the IEEE/CVF conference on computer vision and pattern recognition}, pp. \bibinfo{pages}{11097--11107}.
\bibitem[{Xiao et~al.(2024)Xiao, Wu, Xu, Dai, Hu, Lu, Zeng, Liu and Yuan}]{Xiao2024CVPR}
\bibinfo{author}{Xiao, B.}, \bibinfo{author}{Wu, H.}, \bibinfo{author}{Xu, W.}, \bibinfo{author}{Dai, X.}, \bibinfo{author}{Hu, H.}, \bibinfo{author}{Lu, Y.}, \bibinfo{author}{Zeng, M.}, \bibinfo{author}{Liu, C.}, \bibinfo{author}{Yuan, L.}, \bibinfo{year}{2024}.
\newblock \bibinfo{title}{Florence-2: Advancing a unified representation for a variety of vision tasks}, in: \bibinfo{booktitle}{Proceedings of the {IEEE}/{CVF} Conference on Computer Vision and Pattern Recognition (CVPR)}, pp. \bibinfo{pages}{4818--4829}.
\bibitem[{Xu et~al.(2025)Xu, Wang, Xia, Zhu, Huang and Xu}]{xu2025vla}
\bibinfo{author}{Xu, S.}, \bibinfo{author}{Wang, Y.}, \bibinfo{author}{Xia, C.}, \bibinfo{author}{Zhu, D.}, \bibinfo{author}{Huang, T.}, \bibinfo{author}{Xu, C.}, \bibinfo{year}{2025}.
\newblock \bibinfo{title}{Vla-cache: Towards efficient vision-language-action model via adaptive token caching in robotic manipulation}.
\newblock \bibinfo{journal}{arXiv preprint arXiv:2502.02175} .
\bibitem[{Xue et~al.(2025)Xue, Huang, Niu, Liao, Kragerud, Gravdahl, Peng, Shi, Darrell, Sreenath and Sastry}]{xue2025leverb}
\bibinfo{author}{Xue, H.}, \bibinfo{author}{Huang, X.}, \bibinfo{author}{Niu, D.}, \bibinfo{author}{Liao, Q.}, \bibinfo{author}{Kragerud, T.}, \bibinfo{author}{Gravdahl, J.T.}, \bibinfo{author}{Peng, X.B.}, \bibinfo{author}{Shi, G.}, \bibinfo{author}{Darrell, T.}, \bibinfo{author}{Sreenath, K.}, \bibinfo{author}{Sastry, S.}, \bibinfo{year}{2025}.
\newblock \bibinfo{title}{Leverb: Humanoid whole-body control with latent vision-language instruction}.
\newblock \bibinfo{journal}{arXiv preprint arXiv:2506.13751} \URLprefix \url{https://arxiv.org/abs/2506.13751}.
\bibitem[{Yan et~al.(2024)Yan, Liu, Zheng, Zhong, Huang, Guan, Feng and Ma}]{yan2024robomm}
\bibinfo{author}{Yan, F.}, \bibinfo{author}{Liu, F.}, \bibinfo{author}{Zheng, L.}, \bibinfo{author}{Zhong, Y.}, \bibinfo{author}{Huang, Y.}, \bibinfo{author}{Guan, Z.}, \bibinfo{author}{Feng, C.}, \bibinfo{author}{Ma, L.}, \bibinfo{year}{2024}.
\newblock \bibinfo{title}{Robomm: All-in-one multimodal large model for robotic manipulation}.
\newblock \bibinfo{journal}{arXiv preprint arXiv:2412.07215} \URLprefix \url{https://arxiv.org/abs/2412.07215}.
\bibitem[{Yang et~al.(2025)Yang, Tan, Wu, Zheng, Peng, Liang, Gu, Cai, Ye, Jang et~al.}]{yang2025magma}
\bibinfo{author}{Yang, J.}, \bibinfo{author}{Tan, R.}, \bibinfo{author}{Wu, Q.}, \bibinfo{author}{Zheng, R.}, \bibinfo{author}{Peng, B.}, \bibinfo{author}{Liang, Y.}, \bibinfo{author}{Gu, Y.}, \bibinfo{author}{Cai, M.}, \bibinfo{author}{Ye, S.}, \bibinfo{author}{Jang, J.}, et~al., \bibinfo{year}{2025}.
\newblock \bibinfo{title}{Magma: A foundation model for multimodal ai agents}, in: \bibinfo{booktitle}{Proceedings of the Computer Vision and Pattern Recognition Conference}, pp. \bibinfo{pages}{14203--14214}.
\bibitem[{Yang et~al.(2023)Yang, Du, Ghasemipour, Tompson, Schuurmans and Abbeel}]{yang2023learning}
\bibinfo{author}{Yang, M.}, \bibinfo{author}{Du, Y.}, \bibinfo{author}{Ghasemipour, K.}, \bibinfo{author}{Tompson, J.}, \bibinfo{author}{Schuurmans, D.}, \bibinfo{author}{Abbeel, P.}, \bibinfo{year}{2023}.
\newblock \bibinfo{title}{Learning interactive real-world simulators}.
\newblock \bibinfo{journal}{arXiv preprint arXiv:2310.06114} \bibinfo{volume}{1}, \bibinfo{pages}{6}.
\bibitem[{Yu et~al.(2020)Yu, Quillen, He, Julian, Hausman, Finn and Levine}]{yu2020meta}
\bibinfo{author}{Yu, T.}, \bibinfo{author}{Quillen, D.}, \bibinfo{author}{He, Z.}, \bibinfo{author}{Julian, R.}, \bibinfo{author}{Hausman, K.}, \bibinfo{author}{Finn, C.}, \bibinfo{author}{Levine, S.}, \bibinfo{year}{2020}.
\newblock \bibinfo{title}{Meta-world: A benchmark and evaluation for multi-task and meta reinforcement learning}, in: \bibinfo{booktitle}{Conference on robot learning}, \bibinfo{organization}{PMLR}. pp. \bibinfo{pages}{1094--1100}.
\bibitem[{Zawalski et~al.(2025)Zawalski, Chen, Pertsch, Mees, Finn and Levine}]{zawalski2025ecot}
\bibinfo{author}{Zawalski, M.}, \bibinfo{author}{Chen, W.}, \bibinfo{author}{Pertsch, K.}, \bibinfo{author}{Mees, O.}, \bibinfo{author}{Finn, C.}, \bibinfo{author}{Levine, S.}, \bibinfo{year}{2025}.
\newblock \bibinfo{title}{Robotic control via embodied chain-of-thought reasoning}.
\newblock \bibinfo{journal}{arXiv preprint arXiv:2407.08693} \URLprefix \url{https://arxiv.org/abs/2407.08693}.
\bibitem[{Zayyanu et~al.(2024)Zayyanu, Usman, Muda, Salim, Mohamed, Abubakar, Al-Obeidat and Malik}]{zayyanu2024revolutionising}
\bibinfo{author}{Zayyanu, M.}, \bibinfo{author}{Usman, B.T.}, \bibinfo{author}{Muda, Z.}, \bibinfo{author}{Salim, N.A.}, \bibinfo{author}{Mohamed, A.}, \bibinfo{author}{Abubakar, A.Y.}, \bibinfo{author}{Al-Obeidat, F.}, \bibinfo{author}{Malik, M.A.}, \bibinfo{year}{2024}.
\newblock \bibinfo{title}{Revolutionising natural language processing with transformers: A survey}.
\newblock \bibinfo{journal}{Information Processing \& Management} \bibinfo{volume}{61}, \bibinfo{pages}{103528}.
\newblock \DOIprefix\doi{10.1016/j.ipm.2023.103528}.
\bibitem[{Zhang et~al.(2025a)Zhang, Zhang, Ji, Lei, Dai, Chen and Yang}]{zhang2025safevla}
\bibinfo{author}{Zhang, B.}, \bibinfo{author}{Zhang, Y.}, \bibinfo{author}{Ji, J.}, \bibinfo{author}{Lei, Y.}, \bibinfo{author}{Dai, J.}, \bibinfo{author}{Chen, Y.}, \bibinfo{author}{Yang, Y.}, \bibinfo{year}{2025}a.
\newblock \bibinfo{title}{Safevla: Towards safety alignment of vision-language-action model via safe reinforcement learning}.
\newblock \bibinfo{journal}{arXiv e-prints} , \bibinfo{pages}{arXiv--2503}.
\bibitem[{Zhang et~al.(2025b)Zhang, Ding, Lyu, Peng and Wang}]{zhang2025gevrm}
\bibinfo{author}{Zhang, H.}, \bibinfo{author}{Ding, P.}, \bibinfo{author}{Lyu, S.}, \bibinfo{author}{Peng, Y.}, \bibinfo{author}{Wang, D.}, \bibinfo{year}{2025}b.
\newblock \bibinfo{title}{Gevrm: Goal-expressive video generation model for robust visual manipulation}.
\newblock \bibinfo{journal}{arXiv preprint arXiv:2502.09268} .
\bibitem[{Zhang et~al.(2025c)Zhang, Zantout, Kachana, Zhang and Wang}]{zhang2025irefvla}
\bibinfo{author}{Zhang, H.}, \bibinfo{author}{Zantout, N.}, \bibinfo{author}{Kachana, P.}, \bibinfo{author}{Zhang, J.}, \bibinfo{author}{Wang, W.}, \bibinfo{year}{2025}c.
\newblock \bibinfo{title}{Iref-vla: A benchmark for interactive referential grounding with imperfect language in 3d scenes}.
\newblock \bibinfo{journal}{arXiv preprint arXiv:2503.17406} .
\bibitem[{Zhang et~al.(2024a)Zhang, Guo, Chen, Wang, Hu, Shi and Chen}]{zhang2024hirt}
\bibinfo{author}{Zhang, J.}, \bibinfo{author}{Guo, Y.}, \bibinfo{author}{Chen, X.}, \bibinfo{author}{Wang, Y.J.}, \bibinfo{author}{Hu, Y.}, \bibinfo{author}{Shi, C.}, \bibinfo{author}{Chen, J.}, \bibinfo{year}{2024}a.
\newblock \bibinfo{title}{Hirt: Enhancing robotic control with hierarchical robot transformers}, in: \bibinfo{booktitle}{Proceedings of the 8th Conference on Robot Learning (CoRL)}.
\newblock \URLprefix \url{https://arxiv.org/abs/2410.05273}.
\bibitem[{Zhang et~al.(2025d)Zhang, Guo, Hu, Chen, Zhu and Chen}]{cite:209}
\bibinfo{author}{Zhang, J.}, \bibinfo{author}{Guo, Y.}, \bibinfo{author}{Hu, Y.}, \bibinfo{author}{Chen, X.}, \bibinfo{author}{Zhu, X.}, \bibinfo{author}{Chen, J.}, \bibinfo{year}{2025}d.
\newblock \bibinfo{title}{Up-vla: A unified understanding and prediction model for embodied agent}.
\newblock \bibinfo{journal}{arXiv preprint arXiv:2501.18867} .
\bibitem[{Zhang et~al.(2024b)Zhang, Wang, Wang, Li, Liu, Wei, Wang, Zhang and Wang}]{cite:210}
\bibinfo{author}{Zhang, J.}, \bibinfo{author}{Wang, K.}, \bibinfo{author}{Wang, S.}, \bibinfo{author}{Li, M.}, \bibinfo{author}{Liu, H.}, \bibinfo{author}{Wei, S.}, \bibinfo{author}{Wang, Z.}, \bibinfo{author}{Zhang, Z.}, \bibinfo{author}{Wang, H.}, \bibinfo{year}{2024}b.
\newblock \bibinfo{title}{Uni-navid: A video-based vision-language-action model for unifying embodied navigation tasks}.
\newblock \bibinfo{journal}{arXiv preprint arXiv:2412.06224} .
\bibitem[{Zhang et~al.(2025e)Zhang, Dong, Zhang, Heng, Chi, Dai, Du, Wang, Du and Zhang}]{cite:213}
\bibinfo{author}{Zhang, R.}, \bibinfo{author}{Dong, M.}, \bibinfo{author}{Zhang, Y.}, \bibinfo{author}{Heng, L.}, \bibinfo{author}{Chi, X.}, \bibinfo{author}{Dai, G.}, \bibinfo{author}{Du, L.}, \bibinfo{author}{Wang, D.}, \bibinfo{author}{Du, Y.}, \bibinfo{author}{Zhang, S.}, \bibinfo{year}{2025}e.
\newblock \bibinfo{title}{Mole-vla: Dynamic layer-skipping vision language action model via mixture-of-layers for efficient robot manipulation}.
\newblock \bibinfo{journal}{arXiv preprint arXiv:2503.20384} .
\bibitem[{Zhang et~al.(2024c)Zhang, Zheng, Chen, Jang, Li, Han, Wang, Ding, Fox and Yao}]{zhang2024grape}
\bibinfo{author}{Zhang, Z.}, \bibinfo{author}{Zheng, K.}, \bibinfo{author}{Chen, Z.}, \bibinfo{author}{Jang, J.}, \bibinfo{author}{Li, Y.}, \bibinfo{author}{Han, S.}, \bibinfo{author}{Wang, C.}, \bibinfo{author}{Ding, M.}, \bibinfo{author}{Fox, D.}, \bibinfo{author}{Yao, H.}, \bibinfo{year}{2024}c.
\newblock \bibinfo{title}{Grape: Generalizing robot policy via preference alignment}.
\newblock \bibinfo{journal}{arXiv preprint arXiv:2411.19309} .
\bibitem[{Zhao et~al.(2025a)Zhao, Song, Wang, Tong, Ding, Cheng and Ge}]{cite:214}
\bibinfo{author}{Zhao, H.}, \bibinfo{author}{Song, W.}, \bibinfo{author}{Wang, D.}, \bibinfo{author}{Tong, X.}, \bibinfo{author}{Ding, P.}, \bibinfo{author}{Cheng, X.}, \bibinfo{author}{Ge, Z.}, \bibinfo{year}{2025}a.
\newblock \bibinfo{title}{More: Unlocking scalability in reinforcement learning for quadruped vision-language-action models}.
\newblock \bibinfo{journal}{arXiv preprint arXiv:2503.08007} .
\bibitem[{Zhao et~al.(2023)Zhao, Kumar, Levine and Finn}]{cite:216}
\bibinfo{author}{Zhao, T.}, \bibinfo{author}{Kumar, V.}, \bibinfo{author}{Levine, S.}, \bibinfo{author}{Finn, C.}, \bibinfo{year}{2023}.
\newblock \bibinfo{title}{Learning fine-grained bimanual manipulation with low-cost hardware}.
\newblock \bibinfo{journal}{arXiv preprint arXiv:2304.13705} .
\bibitem[{Zhao et~al.(2025b)Zhao, Ding, Zhang, Gong, Bai, Zhao and Wang}]{zhao2025vlas}
\bibinfo{author}{Zhao, W.}, \bibinfo{author}{Ding, P.}, \bibinfo{author}{Zhang, M.}, \bibinfo{author}{Gong, Z.}, \bibinfo{author}{Bai, S.}, \bibinfo{author}{Zhao, H.}, \bibinfo{author}{Wang, D.}, \bibinfo{year}{2025}b.
\newblock \bibinfo{title}{Vlas: Vision-language-action model with speech instructions for customized robot manipulation}.
\newblock \bibinfo{journal}{arXiv preprint arXiv:2502.13508} .
\bibitem[{Zhao et~al.(2025c)Zhao, Li, Gong, Ding, Zhao and Wang}]{zhao2025oevla}
\bibinfo{author}{Zhao, W.}, \bibinfo{author}{Li, G.}, \bibinfo{author}{Gong, Z.}, \bibinfo{author}{Ding, P.}, \bibinfo{author}{Zhao, H.}, \bibinfo{author}{Wang, D.}, \bibinfo{year}{2025}c.
\newblock \bibinfo{title}{Unveiling the potential of vision-language-action models with open-ended multimodal instructions}.
\newblock \bibinfo{journal}{arXiv preprint arXiv:2505.11214} \URLprefix \url{https://arxiv.org/abs/2505.11214}.
\bibitem[{Zhen et~al.(2024)Zhen, Qiu, Chen, Yang, Yan, Du, Hong and Gan}]{zhen2024threedvla}
\bibinfo{author}{Zhen, H.}, \bibinfo{author}{Qiu, X.}, \bibinfo{author}{Chen, P.}, \bibinfo{author}{Yang, J.}, \bibinfo{author}{Yan, X.}, \bibinfo{author}{Du, Y.}, \bibinfo{author}{Hong, Y.}, \bibinfo{author}{Gan, C.}, \bibinfo{year}{2024}.
\newblock \bibinfo{title}{3d-vla: A 3d vision-language-action generative world model}.
\newblock \bibinfo{journal}{arXiv preprint arXiv:2403.09631} \URLprefix \url{https://arxiv.org/abs/2403.09631}.
\bibitem[{Zheng et~al.(2025a)Zheng, Li, Liu, Zheng, Wang, Ou, Liu, Liu, Zhang and Zhan}]{zheng2025universal}
\bibinfo{author}{Zheng, J.}, \bibinfo{author}{Li, J.}, \bibinfo{author}{Liu, D.}, \bibinfo{author}{Zheng, Y.}, \bibinfo{author}{Wang, Z.}, \bibinfo{author}{Ou, Z.}, \bibinfo{author}{Liu, Y.}, \bibinfo{author}{Liu, J.}, \bibinfo{author}{Zhang, Y.Q.}, \bibinfo{author}{Zhan, X.}, \bibinfo{year}{2025}a.
\newblock \bibinfo{title}{Universal actions for enhanced embodied foundation models}.
\newblock \bibinfo{journal}{arXiv preprint arXiv:2501.10105} .
\bibitem[{Zheng et~al.(2025b)Zheng, Liang, Huang, Gao, III, Kolobov, Huang and Yang}]{zheng2025tracevla}
\bibinfo{author}{Zheng, R.}, \bibinfo{author}{Liang, Y.}, \bibinfo{author}{Huang, S.}, \bibinfo{author}{Gao, J.}, \bibinfo{author}{III, H.D.}, \bibinfo{author}{Kolobov, A.}, \bibinfo{author}{Huang, F.}, \bibinfo{author}{Yang, J.}, \bibinfo{year}{2025}b.
\newblock \bibinfo{title}{Tracevla: Visual trace prompting enhances spatial-temporal awareness for generalist robotic policies}, in: \bibinfo{booktitle}{International Conference on Learning Representations (ICLR)}.
\newblock \URLprefix \url{https://arxiv.org/abs/2412.10345}.
\bibitem[{Zhong et~al.(2025)Zhong, Huang, Li, Zhang, Liang, Yang and Chen}]{cite:219}
\bibinfo{author}{Zhong, Y.}, \bibinfo{author}{Huang, X.}, \bibinfo{author}{Li, R.}, \bibinfo{author}{Zhang, C.}, \bibinfo{author}{Liang, Y.}, \bibinfo{author}{Yang, Y.}, \bibinfo{author}{Chen, Y.}, \bibinfo{year}{2025}.
\newblock \bibinfo{title}{Dexgraspvla: A vision-language-action framework towards general dexterous grasping}.
\newblock \bibinfo{journal}{arXiv preprint arXiv:2502.20900} .
\bibitem[{Zhou et~al.(2025)Zhou, Zhu, Zhu, Wen, Liu, Xu, Meng, Cheng, Peng, Shen et~al.}]{zhou2025chatvla}
\bibinfo{author}{Zhou, Z.}, \bibinfo{author}{Zhu, Y.}, \bibinfo{author}{Zhu, M.}, \bibinfo{author}{Wen, J.}, \bibinfo{author}{Liu, N.}, \bibinfo{author}{Xu, Z.}, \bibinfo{author}{Meng, W.}, \bibinfo{author}{Cheng, R.}, \bibinfo{author}{Peng, Y.}, \bibinfo{author}{Shen, C.}, et~al., \bibinfo{year}{2025}.
\newblock \bibinfo{title}{Chatvla: Unified multimodal understanding and robot control with vision-language-action model}.
\newblock \bibinfo{journal}{arXiv preprint arXiv:2502.14420} .
\bibitem[{Zhu et~al.(2025a)Zhu, Zhu, Li, Zhou, Wen, Liu, Shen, Peng and Feng}]{cite:223}
\bibinfo{author}{Zhu, M.}, \bibinfo{author}{Zhu, Y.}, \bibinfo{author}{Li, J.}, \bibinfo{author}{Zhou, Z.}, \bibinfo{author}{Wen, J.}, \bibinfo{author}{Liu, X.}, \bibinfo{author}{Shen, C.}, \bibinfo{author}{Peng, Y.}, \bibinfo{author}{Feng, F.}, \bibinfo{year}{2025}a.
\newblock \bibinfo{title}{Objectvla: End-to-end open-world object manipulation without demonstration}.
\newblock \bibinfo{journal}{arXiv preprint arXiv:2502.19250} .
\bibitem[{Zhu et~al.(2025b)Zhu, Zhang, Zhou, Yu, Wang, Ma and Xu}]{zhu2025opendrivevla}
\bibinfo{author}{Zhu, T.}, \bibinfo{author}{Zhang, H.}, \bibinfo{author}{Zhou, Y.}, \bibinfo{author}{Yu, W.}, \bibinfo{author}{Wang, Y.}, \bibinfo{author}{Ma, L.}, \bibinfo{author}{Xu, H.}, \bibinfo{year}{2025}b.
\newblock \bibinfo{title}{Opendrive-vla: Generalist vision-language-action agent for autonomous driving}.
\newblock \bibinfo{journal}{arXiv preprint arXiv:2505.01871} \URLprefix \url{https://arxiv.org/abs/2505.01871}.
\bibitem[{Zhu et~al.(2020)Zhu, Gupta, Ebert et~al.}]{zhu2020robosuite}
\bibinfo{author}{Zhu, Y.}, \bibinfo{author}{Gupta, A.}, \bibinfo{author}{Ebert, F.}, et~al., \bibinfo{year}{2020}.
\newblock \bibinfo{title}{robosuite: A modular simulation framework and benchmark for robot learning}.
\newblock \bibinfo{journal}{arXiv preprint arXiv:2009.12293} .
\bibitem[{Zitkovich et~al.(2023)Zitkovich, Yu, Xu, Xu, Xiao, Xia, Wu, Wohlhart, Welker, Wahid et~al.}]{cite:224}
\bibinfo{author}{Zitkovich, B.}, \bibinfo{author}{Yu, T.}, \bibinfo{author}{Xu, S.}, \bibinfo{author}{Xu, P.}, \bibinfo{author}{Xiao, T.}, \bibinfo{author}{Xia, F.}, \bibinfo{author}{Wu, J.}, \bibinfo{author}{Wohlhart, P.}, \bibinfo{author}{Welker, S.}, \bibinfo{author}{Wahid, A.}, et~al., \bibinfo{year}{2023}.
\newblock \bibinfo{title}{Rt-2: Vision-language-action models transfer web knowledge to robotic control}.
\newblock \bibinfo{journal}{Conference on Robot Learning, PMLR} , \bibinfo{pages}{2165--2183}.

\end{thebibliography}

\end{document}